\documentclass[10pt,twocolumn,letterpaper]{article}

\usepackage{cvpr}              %

\usepackage{graphicx}
\usepackage{amsmath}
\usepackage{amssymb}
\usepackage{booktabs}

\usepackage[breaklinks,colorlinks,pdfa]{hyperref}
\hypersetup{
    pdftitle = {GenDR: A Generalized Differentiable Renderer},
    pdflang = {en-US},
}

\usepackage[backend=biber,style=ieee,citestyle=ieee]{biblatex}
\AtBeginBibliography{\small}

\makeatletter
\@namedef{ver@everyshi.sty}{}
\makeatother
\usepackage{tikz}
\usetikzlibrary{trees,positioning,shapes,shadows,arrows}

\usepackage{justtrees}

\usepackage{relsize}
\usepackage{afterpage}

\usepackage{pifont}%
\newcommand{\cmark}{\ding{51}}%
\newcommand{\xmark}{\ding{55}}%

\newcommand{\subheading}[1]{{\bf #1.}}

\newcommand{\density}{f}
\newcommand{\cdf}{F}
\newcommand{\cdfsq}{F_\text{sq}}
\newcommand{\cdfrev}{F_\text{Rev.}}

\newcommand{\xcaption}[1]{\vspace{-2mm}\caption{#1}}

\usepackage{amsthm}
\newtheorem{theorem}{Theorem}

\newtheorem{definition}[theorem]{Definition}
\newtheorem{example}[theorem]{Example}

\newtheorem{remark}[theorem]{Remark}

\def\cvprPaperID{12016} %
\def\confName{CVPR}
\def\confYear{2022}

\newcommand{\numrenderers}{1\,242}

\begin{document}

\title{GenDR: A Generalized Differentiable Renderer}

\author{Felix Petersen$^1$~~~~~~Bastian Goldluecke$^1$~~~~~~Christian Borgelt$^2$~~~~~~Oliver Deussen$^1$\\[.3em]
\,$^1$University of Konstanz~~~~~~~$^2$University of Salzburg
}
\maketitle

\begin{abstract}
In this work, we present and study a generalized family of differentiable renderers.
We discuss from scratch which components are necessary for differentiable rendering and formalize the requirements for each component.
We instantiate our general differentiable renderer, which generalizes existing differentiable renderers like SoftRas and DIB-R, with an array of different smoothing distributions to cover a large spectrum of reasonable settings.
We evaluate an array of differentiable renderer instantiations on the popular ShapeNet 3D reconstruction benchmark and analyze the implications of our results.
Surprisingly, the simple uniform distribution yields the best overall results when averaged over 13 classes; in general, however, the optimal choice of distribution heavily depends on the task.
\end{abstract}

\section{Introduction}
\label{sec:intro}

In the past years, many differentiable renderers have been published.
These include the seminal differentiable mesh renderer OpenDR~\cite{Loper2014}, the Neural 3D Mesh Renderer~\cite{Kato2017}, and SoftRas~\cite{Liu2019-SoftRas} among many others.
Using a differentiable renderer enables a multitude of computer vision applications, such as human pose estimation~\cite{bogo2016keep}, camera intrinsics estimation~\cite{Palazzi2019}, 3D shape optimization~\cite{Kato2017}, 3D~reconstruction~\cite{Kato2017,Liu2019-SoftRas,Chen2019DIB}, and 3D style transfer~\cite{Kato2017}.

A fundamental difference between different classes of differentiable renderers is the choice of the underlying 3D representation.
In this work, we focus on differentiable 3D mesh renderers~\cite{Loper2014,Kato2017,Liu2019-SoftRas,Chen2019DIB}; however, the aspects that we investigate could also be applied to other differentiable rendering concepts, such as rendering voxels~\cite{Yan2016}, point clouds~\cite{insafutdinov18pointclouds}, surfels~\cite{Yifan2019-DiffSurfaceSplatting}, signed distance functions~\cite{Jiang2020-SDFDiff}, and other implicit representations~\cite{liu2019learning,Sitzmann2019SceneRepresentations}.

Differentiable mesh renderers can be constructed in different ways:
either using an exact and hard renderer with approximate surrogate gradients or using an approximate renderer with natural gradients.
Loper~\etal~\cite{Loper2014} and Kato~\etal~\cite{Kato2017} produce approximate surrogate gradients for their differentiable renderer, while their forward rendering is hard.
In contrast, other differentiable renderers approximate the forward rendering in such a way that they produce a natural gradient. 
This can be achieved by modeling or approximating a renderer under a probabilistic perturbation, which is continuous and makes the renderer differentiable.
For that, Rhodin~\etal~\cite{rhodin2015versatile} model it with a Gaussian distribution, while Liu~\etal~\cite{Liu2019-SoftRas} model it with the square root of a logistic distribution, Petersen~\etal~\cite{petersen2021learning} use a logistic distribution, and Chen~\etal~\cite{Chen2019DIB} use the exponential distribution.
While this variational interpretation of perturbing by a respective distribution is not stressed in some of these papers~\cite{Liu2019-SoftRas, Chen2019DIB}, we believe it is important because it explicitly allows comparing the characteristics of the differentiable renderers. 
Moreover, the methods that only approximate gradients can also be seen as approximately modelling a perturbation: the gradient computed for the Neural 3D Mesh Renderer~\cite{Kato2017} is approximately a perturbation by a uniform distribution.
Note that, here, the solutions for rendering under perturbations are obtained analytically in closed-form without sampling.

In this work, we introduce a generalized differentiable renderer (GenDR).
By choosing an appropriate probability distribution, we can (at least approximately) recover the above differentiable mesh renderers, which shows that a core distinguishing aspect of differentiable renderers is the type of distributions that they model. 
The choice of probability distribution herein is directly linked to the sigmoid (i.e., S-shaped) function used for the rasterization.
For example, a Heaviside sigmoid function corresponding to the Dirac delta distribution yields a conventional non-differentiable renderer, while a logistic sigmoid function of squared distances corresponds to the square root of a logistic distribution.
Herein, the sigmoid function is the cumulative distribution function (CDF) of the corresponding distribution.
In this work, we select and present an array of distributions and examine their theoretical properties.

Another aspect of approximate differentiable renderers is their aggregation function, i.e., the function that aggregates the occupancy probabilities of all faces for each pixel.
Existing differentiable renderers commonly aggregate the probabilities via the probabilistic sum ($\bot^P(a,b)=a+b-ab$ or $1-\prod_{t\in T} (1-p_t)$), 
which corresponds to the probability that at least one face covers the pixel
assuming that probabilities $p_t$ for each triangle $t$ are stochastically independent 
(cf.~Eq.~4~in~\cite{Liu2019-SoftRas} or Eq.~6~in~\cite{Chen2019DIB}).
In the field of real-valued logics and adjacent fields, this is well-known as a T-conorm, a relaxed form of the logical `or'.
Two examples of other T-conorms are the maximum T-conorm~$\bot^M(a,b)=\max(a,b)$~and
the Einstein sum~$\bot^E(a,b)=(a+b)/(1+ab)$,
which models the relativistic addition of velocities.
We generalize our differentiable renderer to use any continuous T-conorm and present a variety of suitable T-conorms.

In total, the set of resulting concrete instances arising from our generalized differentiable renderer and the proposed choices amounts to $\numrenderers$ concrete differentiable renderers.
We extensively benchmark all of them on a shape optimization task and a camera pose estimation task.
Further, we evaluate the best performing and most interesting instances on the popular ShapeNet~\cite{Chang2015ShapeNet} 13~class single-view 3D~reconstruction experiment~\cite{Kato2017}.
Here, we also include those instances that approximate other existing differentiable renderers.
We note that we do not introduce a new shading technique in this paper, and rely on existing blended shaders instead.

\vspace{0.15em}
We summarize our contributions as follows: 
\vspace{-0.1em}
\begin{itemize}
  \setlength\itemsep{0em}
    \item We propose a generalized differentiable mesh renderer.
    \item We identify existing differentiable renderers (approximately) as instances of our generalized renderer.
    \item We propose a variety of suitable sigmoid functions and T-conorms and group them by their characteristics.
    \item We extensively benchmark $\numrenderers$ concrete differentiable renderers, analyze which characteristics and families of functions lead to a good performance, and find that the best choice heavily depends on the task, class, or characteristics of the data.
\end{itemize}

\section{Related Work}

The related work can be classified into those works that present differentiable renderers and those which apply them, although there is naturally also a significant overlap.
For additional details on differentiable rendering approaches, cf.~the survey by Kato~\etal~\cite{kato2020differentiable}.

\subheading{Analytical Differentiable Renderers}
The first large category of differentiable renderers are those which produce approximate gradients in an analytical and sampling-free way.
This can either happen by surrogate gradients during backpropagation, as in \cite{Kato2017}, or by making the forward computation naturally differentiable by perturbing the distances between pixels and triangles analytically in closed-form \cite{Liu2018, Chen2019DIB, petersen2019pix2vex}.
Our work falls into this category and is of the second case.
Existing works each present their renderer for a specific distribution or sigmoid function.
We formally characterize the necessary functions to a differentiable renderer and present an array of options.

\subheading{Monte-Carlo Differentiable Renderers}
An alternative to analytical differentiable renderers are those which are based on Monte-Carlo sampling techniques.
The first example for this is the ``redner'' path tracer by Li~\etal~\cite{Li2018}, who use edge sampling to approximate the gradients of their renderer.
Loubet~\etal~\cite{Loubet2019ReparameterizingRendering} build on these ideas and reparameterize the involved discontinuous integrands yielding improved gradient estimates.
Zhang~\etal~\cite{zhang2020path} extend these ideas by differentiating the full path integrals which makes the method more efficient and effective.
Lidec~\etal~\cite{lidec2021differentiable} approach Monte-Carlo differentiable rendering by estimating the gradients of a differentiable renderer via the perturbed optimizers method \cite{berthet2020learning}.

\subheading{Applications}
Popular applications for differentiable renderers are pose~\cite{Loper2014, Kato2017, Liu2019-SoftRas, Chen2019DIB, Palazzi2019, lidec2021differentiable, ravi2020accelerating}, shape~\cite{Kato2017, zhang2020path, petersen2019pix2vex, ravi2020accelerating}, material~\cite{liu2017material, shi2020match}, texture~\cite{Liu2019-SoftRas, Chen2019DIB, Loubet2019ReparameterizingRendering}, and lighting~\cite{zhang2020path} estimation.
Here, the parameters of an initial scene are optimized to match the scene in a reference image or a set of reference images.
Another interesting application is single-view 3D shape prediction without 3D supervision. Here, a neural network predicts a 3D representation from a single image, and the rendering of the image is compared to the original input image. This learning process is primarily guided by supervision of the object silhouette.
It is possible to omit this supervision via adversarial style transfer~\cite{petersen2021style}.
Other applications are generating new 3D shapes that match a data set~\cite{Henzler2018, Henderson_2020_CVPR} as well as adversarial examples in the real world~\cite{liu2019beyond}.

In our experiments, we use optimization for pose and shape to benchmark \textit{all} proposed differentiable renderer combinations. 
As the single-view 3D mesh reconstruction is a complex experiment requiring training a neural network, we benchmark our method on this task only for a selected subset of differentiable renderers.

\subheading{T-norms and T-conorms}
T-norms and T-conorms (triangular norms and conorms) are binary functions that generalize the logical conjunction (`and') and disjunction (`or'), respectively, to real-valued logics or probability spaces~\cite{klement2013triangular,van2022analyzing}.
A generalization of `or' is necessary in a differentiable renderer to aggregate the occlusion caused by faces.
The existing analytical differentiable renderers all use the probabilistic T-conorm.

\section{Generalized Differentiable Renderer}
\label{sec:method}

In this section, we present our generalized differentiable mesh renderer.
With a differentiable renderer, we refer to a renderer that is continuous everywhere and differentiable almost everywhere (a.e.). 
Note that, in this context, continuity is a stricter criterion than differentiable a.e.~because formally (i)~conventional renderers are already differentiable a.e.~(which does not mean that they can provide any meaningful gradients), and (ii)~almost all existing ``differentiable'' renderers are not actually differentiable everywhere.

Let us start by introducing how a classic hard rendering algorithm operates.
The first step is to bring all objects into image space, which is typically a sequence of affine transformations followed by the camera projection.
This step is already differentiable.
The second step is the rasterization:
For each pixel, we need to compute the set of faces (typically triangles) which cover it.
If the pixel is covered by at least one face, the face that is closest to the camera is displayed.

\subsection{Differentiable Occlusion Test}
To make the test whether a pixel~$p$ is occluded by a face~$t$ differentiable, we start by
computing the signed Euclidean distance~$d(p, t)$ between pixel and face boundary.
By convention, pixels inside the triangle have a positive distance, pixels outside the triangle a negative distance.
For pixels exactly on the boundary, the distance to the face is~$0$. 

For a hard occlusion test, we would just check whether~$d(p,t)$ is non-negative.
In a differentiable renderer, we instead introduce a perturbation in the form of a probability distribution with density~$\density$
together with a temperature or scale parameter~$\tau>0$.
We then evaluate the probability that the perturbed distance~$d(p,t)-\tau\epsilon$ is non-negative,
where~$\epsilon$ is distributed according to~$\density$.
Thus, we compute the probability that~$t$ occludes $p$ as
\begin{equation}
\begin{aligned}
    \mathbb{P}_{\epsilon \sim \density}( d(p, t) -\tau\epsilon&\geq 0 )    
    = \mathbb{P}_{\epsilon \sim \density}{( \epsilon \leq d(p, t)/\tau ) }\\
    &\kern-1.5em = \int_{-\infty}^{d(p, t)/\tau} \kern-2.25em f(x)\, dx = \cdf\left(\frac{d(p, t)}{\tau}\right),
\end{aligned}
\label{eq:occlusion_test}
\end{equation}
where~$\cdf$ is the CDF of the distribution $\density$ and thus yields a closed-form solution for the desired probability (provided that $\cdf$ has a closed-form solution or can be appropriately approximated). 
In a differentiable renderer, we require~$\cdf$ being continuous.
Typically, $\cdf$ has the S-shape of a sigmoid function, see Table~\ref{tab:vis-sigmoids}.
Therefore, we refer to CDFs as sigmoid functions in this paper.

Most existing differentiable renderers use sigmoid functions or transformations thereof, see Section~\ref{sec:instantiations},
to softly evaluate whether a pixel lies inside a triangle. 
This accords to the probabilistic interpretation in Equation~(\ref{eq:occlusion_test}) where the probability distribution is defined via the sigmoid function used in each case.
Here, the logistic sigmoid function is a popular choice of such a sigmoid function. 
Note that, recently, it has frequently been referred to as ``the'' sigmoid in the literature, which is not to be confused with the original and more
general terminology.
\begin{example}[Logistic Sigmoid]
    $\cdf_L(x)=1 / (1 + \exp(-x))$ is the logistic sigmoid function, which corresponds to the logistic distribution.
\end{example}

\subsection{Aggregation}
The second step to be made differentiable is the aggregation of multiple faces.
While this is conventionally done via a logical `or', the differentiable real-valued counterpart is a T-conorm.
T-conorms are formally defined as follows.
\begin{definition}[T-conorm]\label{def:t-conorm}
    A T-conorm is a binary operation~$\bot : [0,1] \times [0,1] \to [0,1]$, which satisfies
    \begin{itemize}
    \setlength\itemsep{0em}
        \item associativity: $\bot(a, \bot(b, c)) = \bot(\bot(a, b), c)$,
        \item commutativity: $\bot(a, b) = \bot(b, a)$,
        \item monotonicity: $(a\leq c) \land (b\leq d) \Rightarrow \bot(a, b) \leq \bot(c, d)$,
        \item $0$ is a neutral element $\bot(a, 0) = a$.
    \end{itemize}
\end{definition}
\begin{remark}[T-conorms and T-norms]
    While T-conorms~$\bot$ are the real-valued equivalents of the logical `or',
    so-called T-norms~$\top$ are the real-valued equivalents of the logical `and'. %
    Certain T-conorms and T-norms are dual in the sense that
    one can derive one from the other using a complement (typically $1-x$) and De~Morgan's laws ($\top(a, b) = 1-\bot(1-a, 1-b)$).
\end{remark}

Let us proceed by stating the T-conorm which is used in all applicable previous approximate differentiable renderers with natural gradients.
\begin{example}[Probabilistic Sum]
    The \emph{probabilistic sum} is a T-conorm that corresponds to the probability that at least one out of two independent events occurs. It is defined as \vspace{-.5em}\begin{equation}
        \bot^P(a, b) = a + b - ab.
    \end{equation}
\end{example}
An alternative to this is the Einstein sum, which is based on the relativistic addition of velocities.
\begin{example}[Einstein Sum]
    The \emph{Einstein sum} is a T-conorm that corresponds to the velocity addition under special relativity: \vspace{-.75em}\begin{equation}
        \bot^P(a, b) = \frac{a + b}{1 + ab}.
    \end{equation}
\end{example}

Combining the above concepts, we can compute the occupancy or coverage of a pixel $p$ given a set of faces $T$ as 
\begin{equation}
    \mathcal{A}_O(p, T) = \underset{t\in T}{\mathlarger{\mathlarger{\mathlarger{\bot}}}}\; \cdf(\,d(p,t) / \tau\,)\,.
\end{equation}

\begin{figure*}[]
    \centering
\forestset{
  squared just tree/.style={
    for tree={
      edge path={
        \noexpand\path [\forestoption{edge}] (!u.parent anchor) -- +(0,-5pt) -| (.child anchor)\forestoption{edge label};
      },
      align=center,
      if={(isodd(n_children))&&(n_children>2)}
      {
        for children={
          if={equal(n,((n_children("!u"))+1)/2)}
          {
            calign with current
          }{},
        }
      }{},
      if n children=0
      {
        before packing={tier=terminus}
      }{},
    },
  }
}
\begin{minipage}{0.2375\textwidth}
    \caption{
        Taxonomy of probability distributions corresponding to sigmoid functions.
        The subdivisions are chosen wrt.~properties that have a categorically different influence on the behavior of the corresponding renderer. 
        The order of splits when going down in the tree (which could be chosen differently, e.g., symmetric/asymmetric could be the first split) reflects the importance of the properties.
    }
    \label{fig:sigmoid-taxonomy}
\end{minipage}
\hfill
\begin{minipage}{0.725\textwidth}
\resizebox{\linewidth}{!}{
    \begin{justtree}
      {%
        left justifications,
        squared just tree,
      }
      [Taxonomy of Distributions
        [Finite Support
          [Exact
            [Dirac Delta\\ (Heaviside)]
          ]
          [Continuous
            [Uniform\\ Cubic Hermite\\ Wigner Semicircle]
          ]
        ]
        [Infinite Support
          [Symmetrical
            [Exponential Conv.
              [Gaussian\\ Laplace\\ Logistic\\ \kern-4.5em Hyperbolic secant\hbox to0pt{ (Gudermannian)}
              ]
            ]
            [Linear Conv.
              [Cauchy\\ Reciprocal
              ]
            ]
          ]
          [Asymmetrical
            [Two-Sided
              [Gumbel-Max\\ Gumbel-Min
              ]
            ]
            [One-Sided
              [Exponential\\ Gamma\\ Levy
              ]
            ]
          ]
        ]
      ]
    \end{justtree}
}
\end{minipage}
\end{figure*}

\begin{table*}[]
    \centering
    \scriptsize
    \newcommand{\mktriangle}[3]{\kern-3em\includegraphics[width=4em]{img/triangles/t_#1_0_#3_s#2.png}}
    \begin{tabular}{cccccc}
        \toprule
        \includegraphics[]{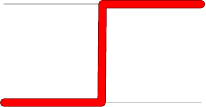}\mktriangle{0}{1.0}{0} &
        \includegraphics[]{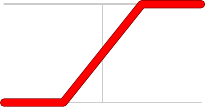}\mktriangle{1}{1.0}{0} &
        \includegraphics[]{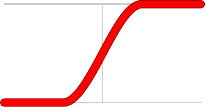}\mktriangle{3}{0.8}{0} &
        \includegraphics[]{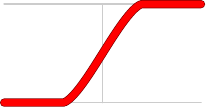}\mktriangle{5}{0.8}{0} &
        \includegraphics[]{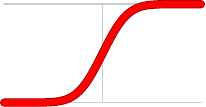}\mktriangle{7}{1.0}{0} &
        \includegraphics[]{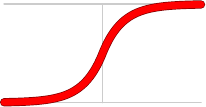}\mktriangle{9}{1.0}{0} 
        \\
        Heaviside &
        Uniform &
        Cubic Hermite &
        Wigner Semicircle &
        Gaussian &
        Laplace
        \\
        \midrule
        \includegraphics[]{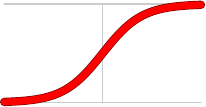}\mktriangle{11}{1.0}{0} &
        \includegraphics[]{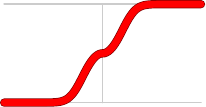}\mktriangle{12}{2.2}{1} &
        \includegraphics[]{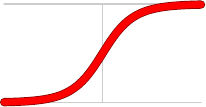}\mktriangle{13}{1.0}{0} &
        \includegraphics[]{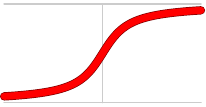}\mktriangle{15}{1.0}{0} &
        \includegraphics[]{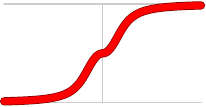}\mktriangle{16}{2.0}{1} &
        \includegraphics[]{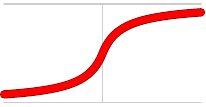}\mktriangle{17}{1.0}{0} 
        \\
        Logistic &
        Logistic (squares) &
        Hyperbolic secant &
        Cauchy &
        Cauchy (squares) &
        Reciprocal 
        \\
        \midrule
        \includegraphics[]{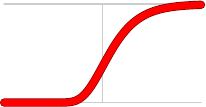}\mktriangle{19}{1.0}{0} &
        \includegraphics[]{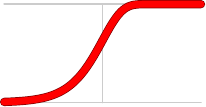}\mktriangle{21}{1.0}{0} &
        \includegraphics[]{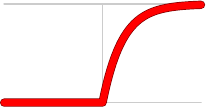}\mktriangle{23}{1.0}{0} &
        \includegraphics[]{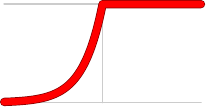}\mktriangle{27}{1.0}{0} &
        \includegraphics[]{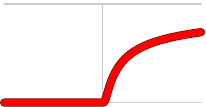}\mktriangle{47}{1.5}{0} &
        \includegraphics[]{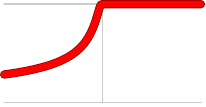}\mktriangle{51}{1.5}{0} 
        \\
        Gumbel-Max &
        Gumbel-Min &
        Exponential &
        Exponential (Rev.) &
        Levy &
        Levy (Rev.) 
        \\
        \midrule
        \includegraphics[]{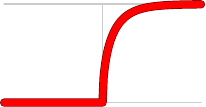}\mktriangle{37}{1.0}{0} &
        \includegraphics[]{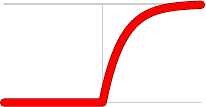}\mktriangle{35}{1.0}{0} &
        \includegraphics[]{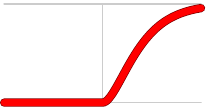}\mktriangle{33}{1.0}{0} &
        \includegraphics[]{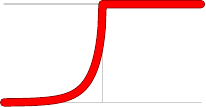}\mktriangle{45}{1.0}{0} &
        \includegraphics[]{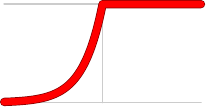}\mktriangle{43}{1.0}{0} &
        \includegraphics[]{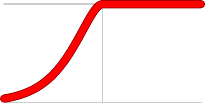}\mktriangle{41}{1.0}{0} 
        \\
        Gamma ($p=0.5$) &
        Gamma ($p=1$) &
        Gamma ($p=2$) &
        Gamma ($p=.5$) (Rev.) &
        Gamma ($p=1$) (Rev.) &
        Gamma ($p=2$) (Rev.) 
        \\
        \midrule
        \includegraphics[]{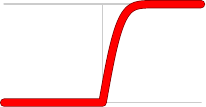}\mktriangle{38}{2.0}{1} &
        \includegraphics[]{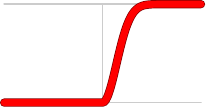}\mktriangle{36}{2.0}{1} &
        \includegraphics[]{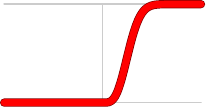}\mktriangle{34}{2.0}{1} &
        \includegraphics[]{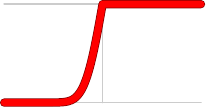}\mktriangle{46}{2.0}{1} &
        \includegraphics[]{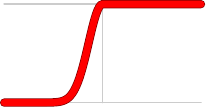}\mktriangle{44}{2.0}{1} &
        \includegraphics[]{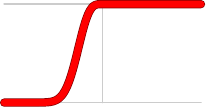}\mktriangle{42}{2.0}{1} 
        \\
        Gamma ($p=0.5$) &
        Gamma ($p=1$) &
        Gamma ($p=2$) &
        Gamma ($p=.5$) (Rev.) &
        Gamma ($p=1$) (Rev.) &
        Gamma ($p=2$) (Rev.) 
        \\
        (squares) &
        (squares) &
        (squares) &
        (squares) &
        (squares) &
        (squares) 
        \\
        \bottomrule
    \end{tabular}
    \xcaption{Visualization of a selection of sigmoid functions, which are the CDFs of probability distributions. For each distribution, we display a single rendered triangle to demonstrate their different effects.}
    \label{tab:vis-sigmoids}
\end{table*}

\subsection{Shading}
The coloring of faces is handled via the Phong model or any other shading model, which is already differentiable. 
In the literature, Chen~\etal~\cite{Chen2019DIB} compare different choices.
Finally, to aggregate the coloring of each pixel depending on the distance of the face to the camera (depth), there are two popular choices in the literature:
no depth perturbations and taking the closest triangle (like \cite{Loper2014, Kato2017, Chen2019DIB}) and Gumbel depth perturbations (like \cite{Liu2019-SoftRas, petersen2019pix2vex}).
Only the latter choice is truly continuous, and the closed-form solution for Gumbel depth perturbations is the well known $\operatorname{softmin}$.
As there are (i) no closed-form solutions for adequate alternatives to Gumbel perturbations in the literature, and (ii) these two options have been extensively studied in the literature~\cite{lidec2021differentiable, Loper2014, Kato2017, Chen2019DIB, Liu2019-SoftRas, petersen2019pix2vex}, in this work we do not modify this component and focus on the differentiable silhouette computation and aggregation. 
While we implement both options in GenDR, in our evaluation, we perform all experiments
agnostic to the choice of shading aggregation as the experiments rely solely on the silhouette.

\section{Instantiations of the GenDR}
\label{sec:instantiations}

Let us proceed by discussing instantiations of the generalized differentiable renderer (GenDR).

\begin{table*}
    \centering
    \resizebox{\linewidth}{!}{\small
    \begin{tabular}{lllccccccc}
        \toprule
        T-conorm &             & equal to / where                       & continuous    & contin.~diff. & strict    & idempotent    & nilpotent & Archimedean   & $\uparrow / \downarrow$~wrt.~$p$\\
        \midrule
        (Logical `or') & $\lor$ &                                       & (\xmark)      & (\xmark)      & ---       & (\cmark)      & ---       & ---           & --- \\
        Maximum & $\bot^M$ &                                            & \cmark        & \xmark        & \xmark    & \cmark        & \xmark    & \xmark        & --- \\
        Probabilistic & $\bot^P$ & $=\bot^H_1 =\bot^A_1$                & \cmark        & \cmark        & \cmark    & \xmark        & \xmark    & \cmark        & --- \\
        Einstein & $\bot^E$      & $=\bot^H_0$                          & \cmark        & \cmark        & \cmark    & \xmark        & \xmark    & \cmark        & --- \\
        \midrule
        Hamacher & $\bot^H_p$ & $p\in(0, \infty)$                       & \cmark        & \cmark        & \cmark    & \xmark        & \xmark    & \cmark        & $\downarrow$ \\
        Frank & $\bot^F_p$ & $p\in(0, \infty)$                          & \cmark        & \cmark        & \cmark    & \xmark        & \xmark    & \cmark        & $\downarrow$ \\
        Yager & $\bot^Y_p$ & $p\in(0, \infty)$                          & \cmark        & \xmark        & \xmark    & \xmark        & \cmark    & \cmark        & $\uparrow$ \\
        Acz\'el-Alsina & $\bot^A_p$ & $p\in(0, \infty)$                 & \cmark        & \cmark        & \cmark    & \xmark        & \xmark    & \cmark        & $\uparrow$ \\
        Dombi & $\bot^D_p$ & $p\in(0, \infty)$                          & \cmark        & \cmark        & \cmark    & \xmark        & \xmark    & \cmark        & $\uparrow$ \\
        Schweizer-Sklar & $\bot^{SS}_p$\kern-.25em & $p\in(-\infty, 0)$ & \cmark        & \cmark        & \cmark    & \xmark        & \xmark    & \cmark        & --- \\
        \bottomrule
    \end{tabular}
    }
    \vspace{-.25em}
    \xcaption{Overview over a selection of suitable T-conorms, which we also benchmark.}
    \label{tab:t-conorms}
\end{table*}
\begin{figure*}
  \centering
  \includegraphics[width=.23\linewidth,trim={0 0 0 1.6cm},clip]{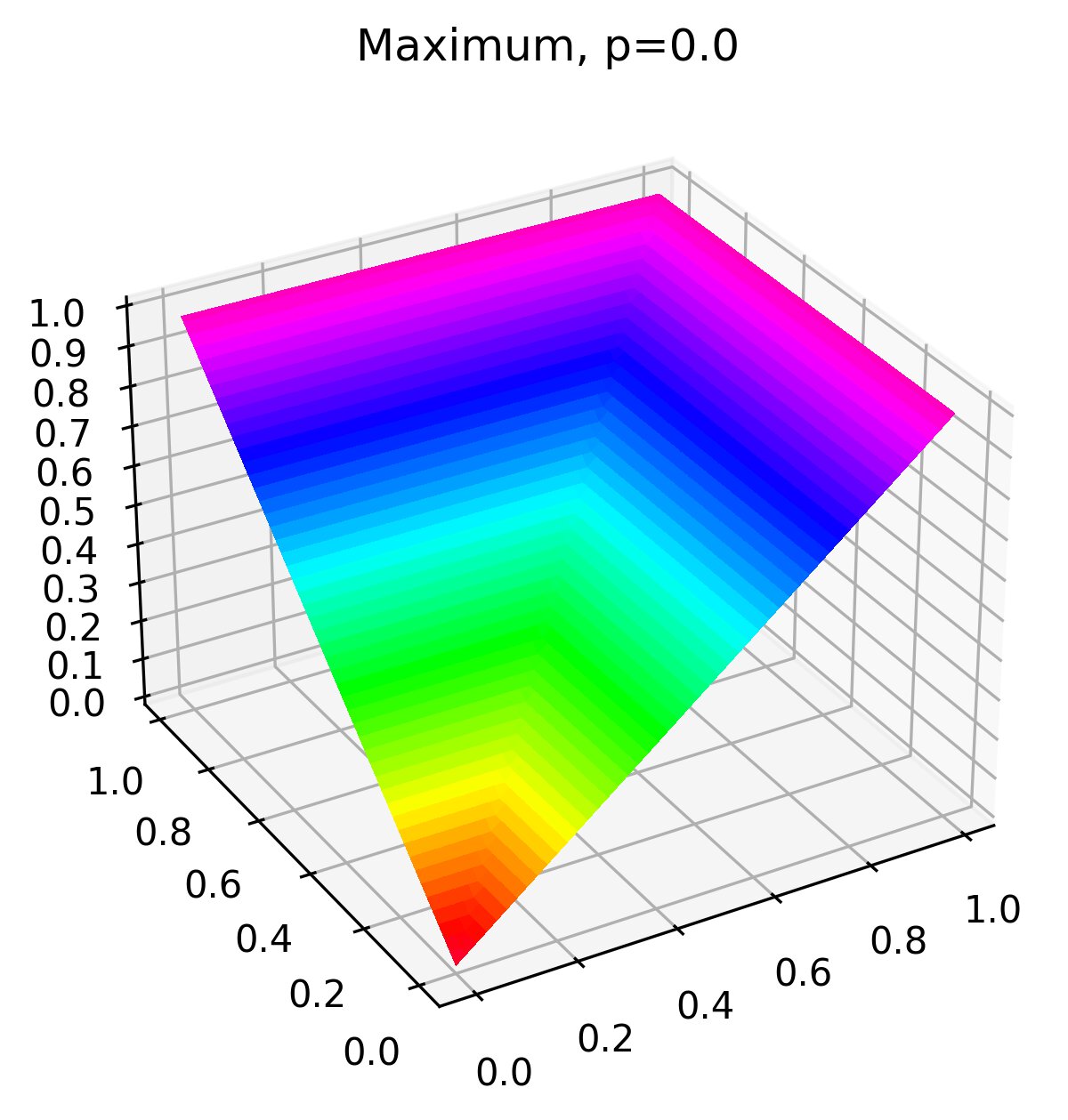}\hfill
  \includegraphics[width=.23\linewidth,trim={0 0 0 1.6cm},clip]{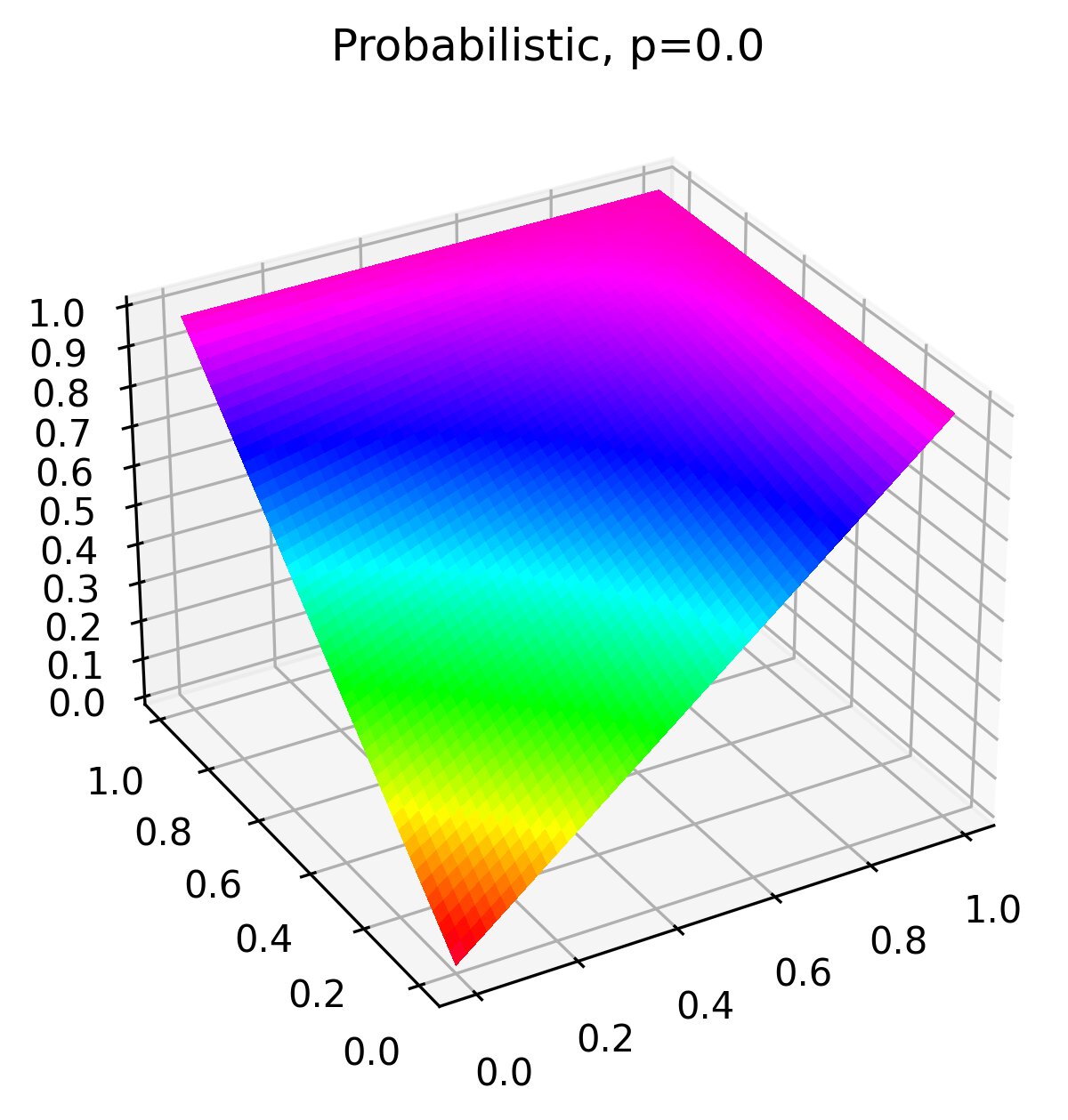}\hfill
  \includegraphics[width=.23\linewidth,trim={0 0 0 1.6cm},clip]{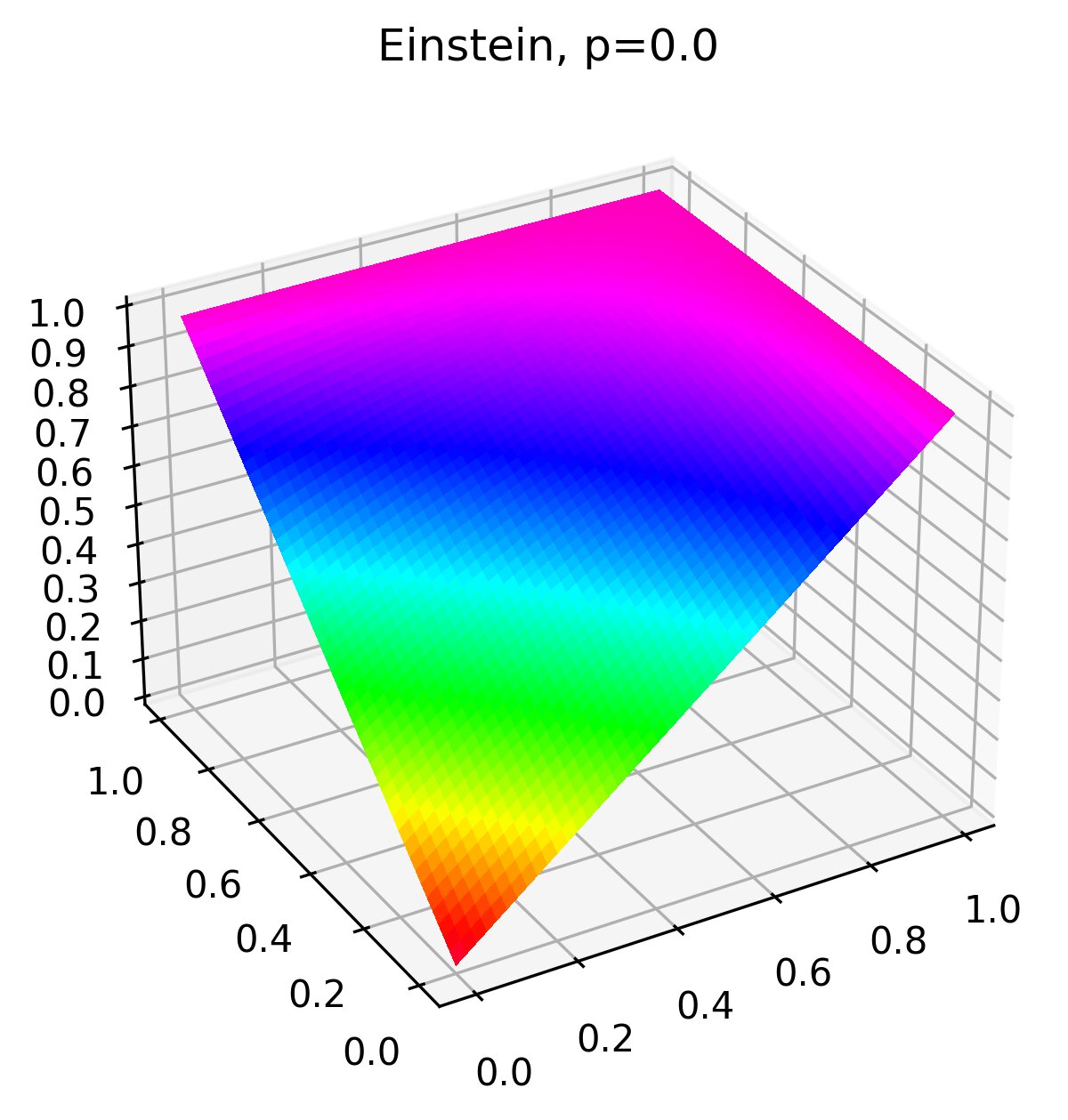}\hfill
  \includegraphics[width=.23\linewidth,trim={0 0 0 1.6cm},clip]{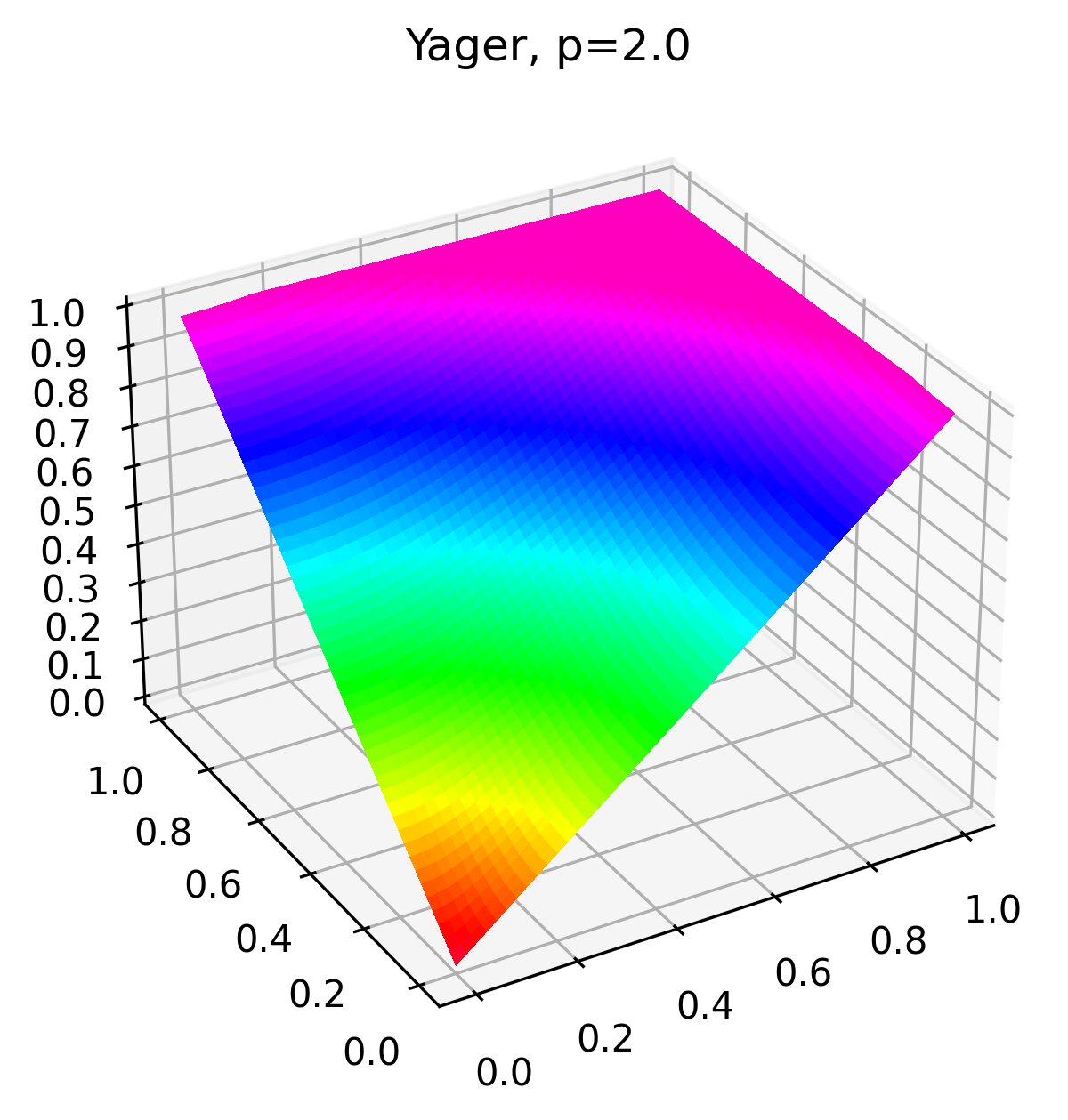}\\[-.25em]
  \noindent
  {\small \hfill (a) \hfill\hspace{.0267\linewidth}\hfill (b) \hfill\hspace{.0267\linewidth}\hfill (c) \hfill\hspace{.0267\linewidth}\hfill (d) \hfill}
  \xcaption{Plot of four selected T-conorms. From left to right: Maximum, Probabilistic, Einstein, and Yager (w/ $p=2$).
    While (b) and (c) are smooth, the Yager T-conorm (d) is non-smooth, it plateaus and the value is constant outside the unit circle.
  }
  \label{fig:t-conorms}
\end{figure*}
\begin{figure}[b]
    \centering
    \includegraphics[width=.49\linewidth]{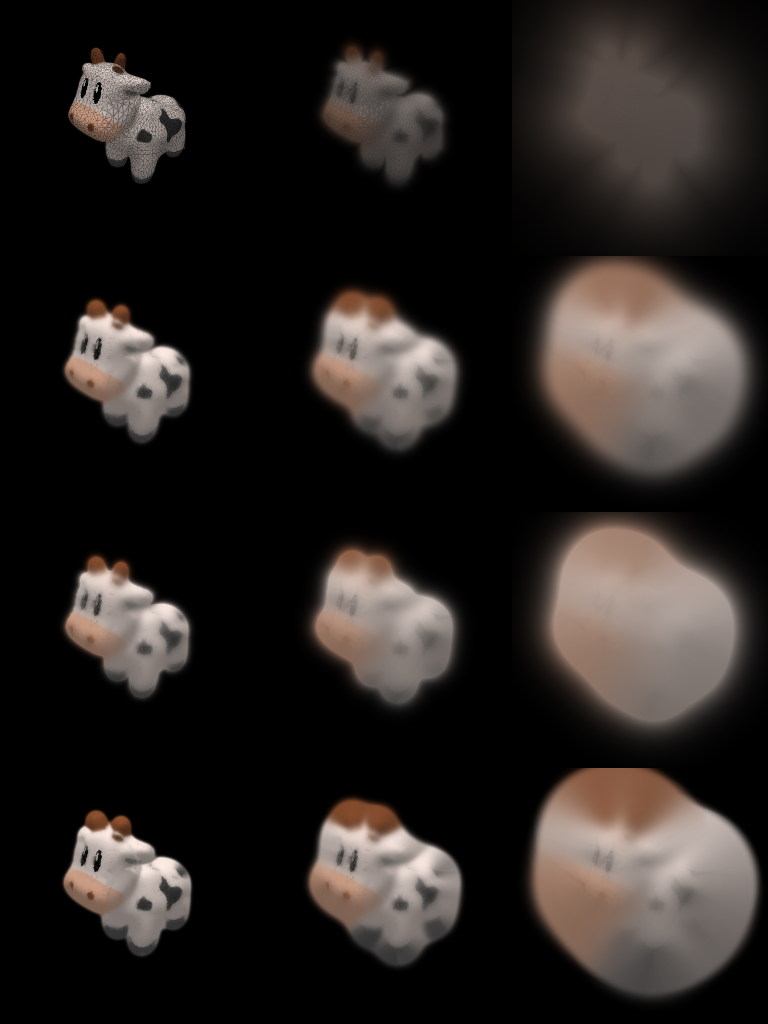}
    \hfill
    \includegraphics[width=.49\linewidth]{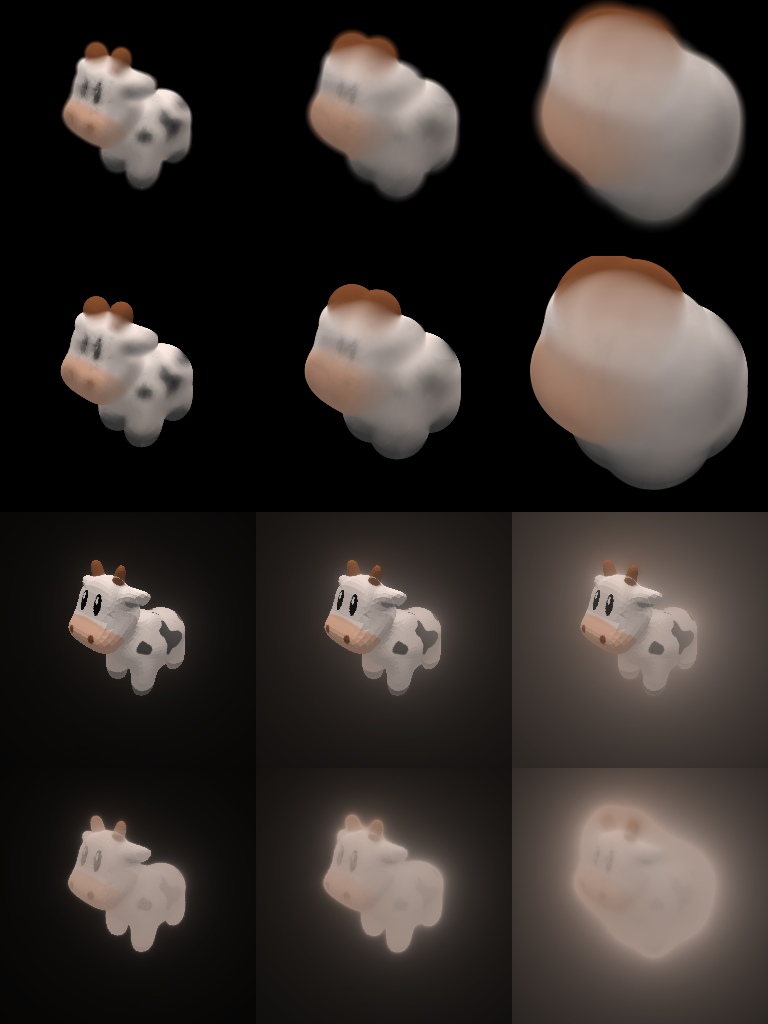}
    \xcaption{
        Visual comparison of different instances of GenDR.
        In each image, moving from left to right increases the temperature or scale~$\tau$ of the distribution.
        Left: we use a logistic distribution to perturb the triangles and use different T-norms for aggregation (top to bottom: $\bot^M, \bot^P, \bot^Y_2, \bot^A_{0.5}$).
        Right: for the two first rows, we use a uniform distribution and use $\bot^Y_2$ and $\bot^A_{0.5}$. For the last two rows, we use a Cauchy distribution and use $\bot^P$ and $\bot^Y_2$.
    }
    \vspace{-.75em}
    \label{fig:example-render}
\end{figure}

\subheading{Distributions}
Figure~\ref{fig:sigmoid-taxonomy} provides a taxonomy of the distributions and sigmoid functions that are visualized in Table~\ref{tab:vis-sigmoids}.
We classify the distributions into those with finite support as well as others with infinite support, where the support is the set of points for which the PDF is greater than zero. 
Note that the CDFs are constant outside the support region. 
Among the distributions with \textit{finite support}, there is the \textit{exact} Dirac delta distribution corresponding to the Heaviside function, which yields a discrete, non-differentiable renderer. 
There are also \textit{continuous} distributions allowing meaningful gradients, but (due to finite support) only in a limited proximity to each face.
Here, we have, among others, the uniform distribution, which corresponds to a piecewise linear step function. 
The derivative of the uniform distribution is equivalent or very similar (due to minor implementation aspects) to the surrogate gradient of the Neural 3D Mesh Renderer~\cite{Kato2017}.
The distributions with \textit{infinite support} can be categorized into symmetrical and asymmetrical.
Among the symmetrical distributions, the Gaussian, the Laplace, the logistic, and the hyperbolic secant have an \textit{exponential convergence} behavior or exponential decay of probability density.
On the other hand, there is also the Cauchy distribution which has a \textit{linear convergence}. This yields a significantly different behavior.
We include the %
algebraic function~$x\mapsto x / (2 + 2|x|) + 1/2$ and call it reciprocal sigmoid.
This also has a \textit{linear convergence}.
Finally, we consider \textit{asymmetrical} distributions with infinite support.
The Gumbel-Max and Gumbel-Min are extreme value distributions~\cite{Coles2001} and \textit{two-sided}, which means that their support covers both positive and negative arguments.
The exponential, Gamma, and Levy distributions are one-sided distributions. Here, it is important to not only consider the original distributions but also their mirrored or reversed variants, as well as shifted variations as can be seen in the last three rows of Table~\ref{tab:vis-sigmoids}.

SoftRas~\cite{Liu2019-SoftRas}
squares the absolute part of the distance
before applying the logistic sigmoid function 
and thus models the square roots of logistic perturbations. %
Instead of modifying the argument of~$\cdf$, we instead interpret it as applying a transformed
counterpart CDF~$\cdfsq$, which is more in line with the probabilistic interpretation
in Equation~(\ref{eq:occlusion_test}).
More precisely, we compute the occlusion probability as 
\begin{equation}
\cdfsq( d(p,t) / \tau ) := \cdf ( |d(p, t)|\cdot d(p,t) / \tau)\,.
\label{eq:cdfsq}
\end{equation}
That means that for each choice of~$\cdf$, we obtain a counterpart~$\cdfsq$. 
A selection of these for different CDFs~$\cdf$ is visualized in Table~\ref{tab:vis-sigmoids} denoted by ``(squares)''.
For a mathematical definition of each sigmoid function, see SM~\ref{sm:dist}.

\subheading{Aggregations}
Table~\ref{tab:t-conorms} provides an overview over selected T-conorms and displays their properties. 
The logical `or' is not a T-conorm but the discrete and discontinuous equivalent, which is why we include it here.
While %
there are also discontinuous T-conorms such as the drastic T-cornom, these are naturally
not suitable for a differentiable renderer, which is why we exlude them.
All except for the Max and Yager T-conorms are continuously differentiable. %

The top four rows in Table~\ref{tab:t-conorms} contain individual T-conorms, and the remainder are families of T-conorms. 
Here, we selected only suitable ranges for the parameter~$p$.
Note that there are some cases in which the T-conorms coincide, e.g, $\bot^P=\bot^H_1 =\bot^A_1$.
A discussion of the remaining properties and a mathematical definition of each T-conorm can be found in SM~\ref{sm:tcn}.
Figure~\ref{fig:t-conorms} displays some of the T-conorms and illustrates different properties.
In Figure~\ref{fig:example-render}, we display example renderings with different settings and provide a visual comparison on how the aggregation function affects rendering.

\begin{table}
    \centering
    \resizebox{\linewidth}{!}{
    \begin{tabular}{llc}
        \toprule
        Renderer & Distribution & T-conorm \\
        \midrule
        OpenDR~\cite{Loper2014} & Uniform (backward) & --- \\
        N3MR~\cite{Kato2017} & Uniform (backward) & --- \\
        Rhodin~\etal~\cite{rhodin2015versatile} & Gaussian & $\bot^P$ \\
        SoftRas~\cite{Kato2017} & Square-root of Logistic & $\bot^P$ \\
        Log.~Relax~\cite{petersen2021learning} & Logistic & $\bot^P$ \\
        DIB-R~\cite{Chen2019DIB} & Exponential & $\bot^P$ \\
        \bottomrule
    \end{tabular}
    }
    \xcaption{Differentiable renderers that are (approximately) special cases of GenDR. OpenDR and N3MR do not use a specific T-conorm as their forward computation is hard.
    }
    \label{tab:existing-renderers-characterization}
\end{table}

\subheading{Existing Special Cases of GenDR}
In Table~\ref{tab:existing-renderers-characterization}, we list which existing differentiable renderers are conceptually instances of GenDR.
These renderers do each have some other differences, but one key difference lies in the type of distribution employed.
Differences regarding shading are also discussed at the end of Section~\ref{sec:method}.

\section{Experiments\protect\footnote{The source code will be available at \href{https://github.com/Felix-Petersen/gendr}{github.com/Felix-Petersen/gendr}.}}

\subsection{Shape Optimization}
\label{sec:shape-opt}
Our first experiment is a shape optimization task.
Here, we use the mesh of an \textit{airplane}, and render it from $24$ azimuths using a hard renderer.
The task is to optimize a mesh (initialized as a sphere) to fit the silhouette of the airplane within $100$ optimization steps.
Limiting the task to $100$ optimization steps is critical for two reasons: 
(i) The task can be considered to be solved perfectly with any differentiable renderer that produces the correct gradient sign within a large number of steps, but we are interested in the quality of the gradients for the optimization task and how efficient each renderer is.
(ii) The total evaluation is computationally expensive because we evaluate a total of $\numrenderers$ renderers and perform a grid search over the distribution parameters for each one to provide a fair and reliable comparison.

\subheading{Setup}
For optimization, we use the Adam optimizer~\cite{Kingma2014AdamOpt} with parameters $\beta_1=0.5, \beta_2=0.95$.
For each setting, we perform a grid search over three learning rates ($\lambda\in\{10^{-1.25}, 10^{-1.5}, 10^{-1.75}\}$) and temperatures $\tau \in \{10^{-0.1\cdot n}\,|\,n\in\mathbb{N}, 0\leq n\leq 80\}$.
Here, $\lambda = 10^{-1.5}\approx0.03$ performs best in almost all cases. %
As for the scale hyperparameter, it is important to use a fine-grained as well as large grid because this behaves differently for each distribution. 
Here, we intentionally chose the grid larger than the range of reasonable values to ensure that the best choice is used for each setting; the extreme values for the scale were never optimal.
We perform this evaluation from five different elevation angles $\{-60^\circ,-30^\circ,0^\circ,30^\circ,60^\circ\}$ as independent runs, and average the final results for each renderer instance.
Additional results for the experiment applied to the model of a chair can be found in SM~\ref{sm:add-plots}.

\begin{figure}[t]
  \centering
  \includegraphics[width=\linewidth,trim={0.5cm 4.1cm 0.25cm 0.5cm},clip]{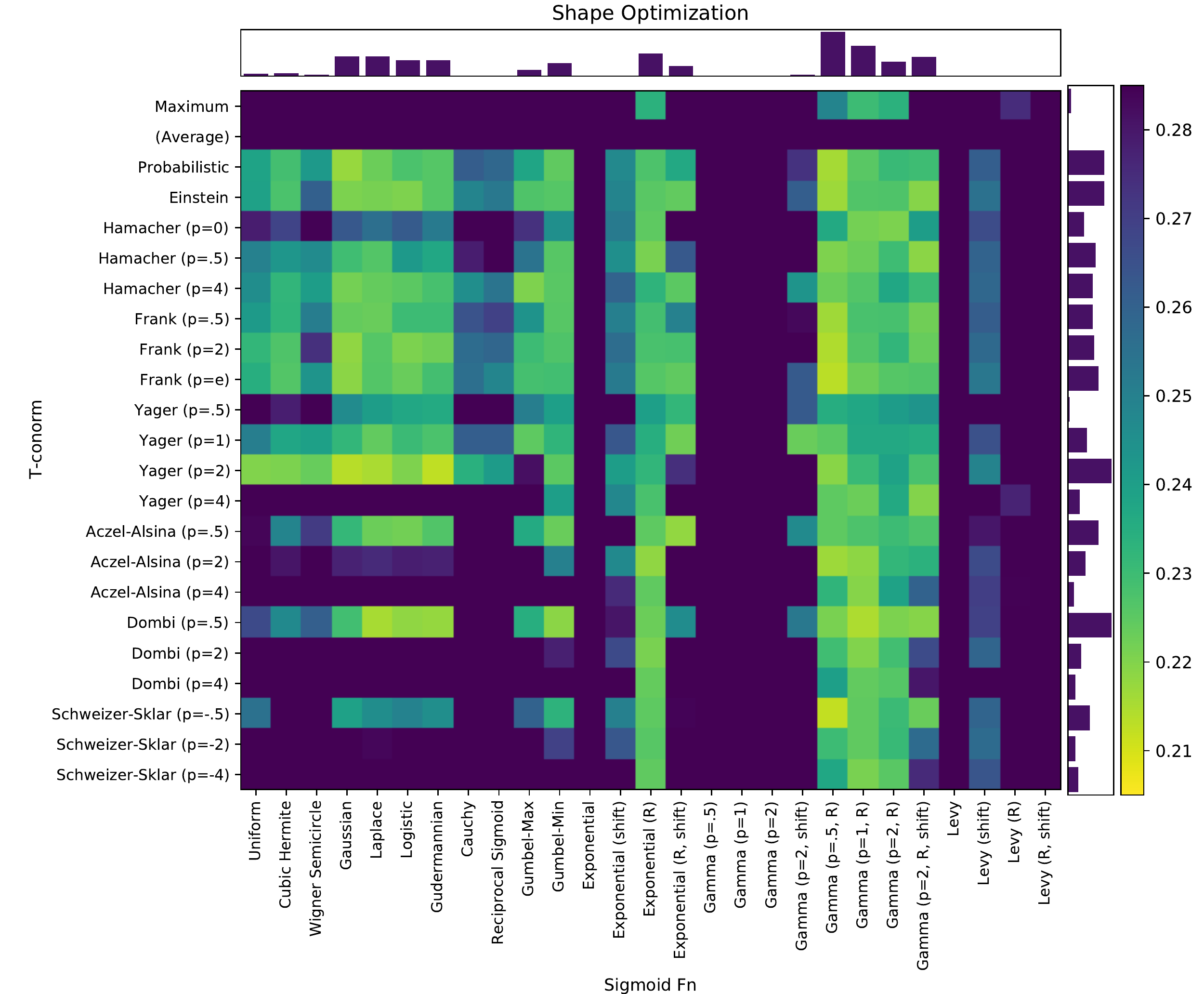}
  \includegraphics[width=\linewidth,trim={0.5cm 0.0cm 0.25cm 0.5cm},clip]{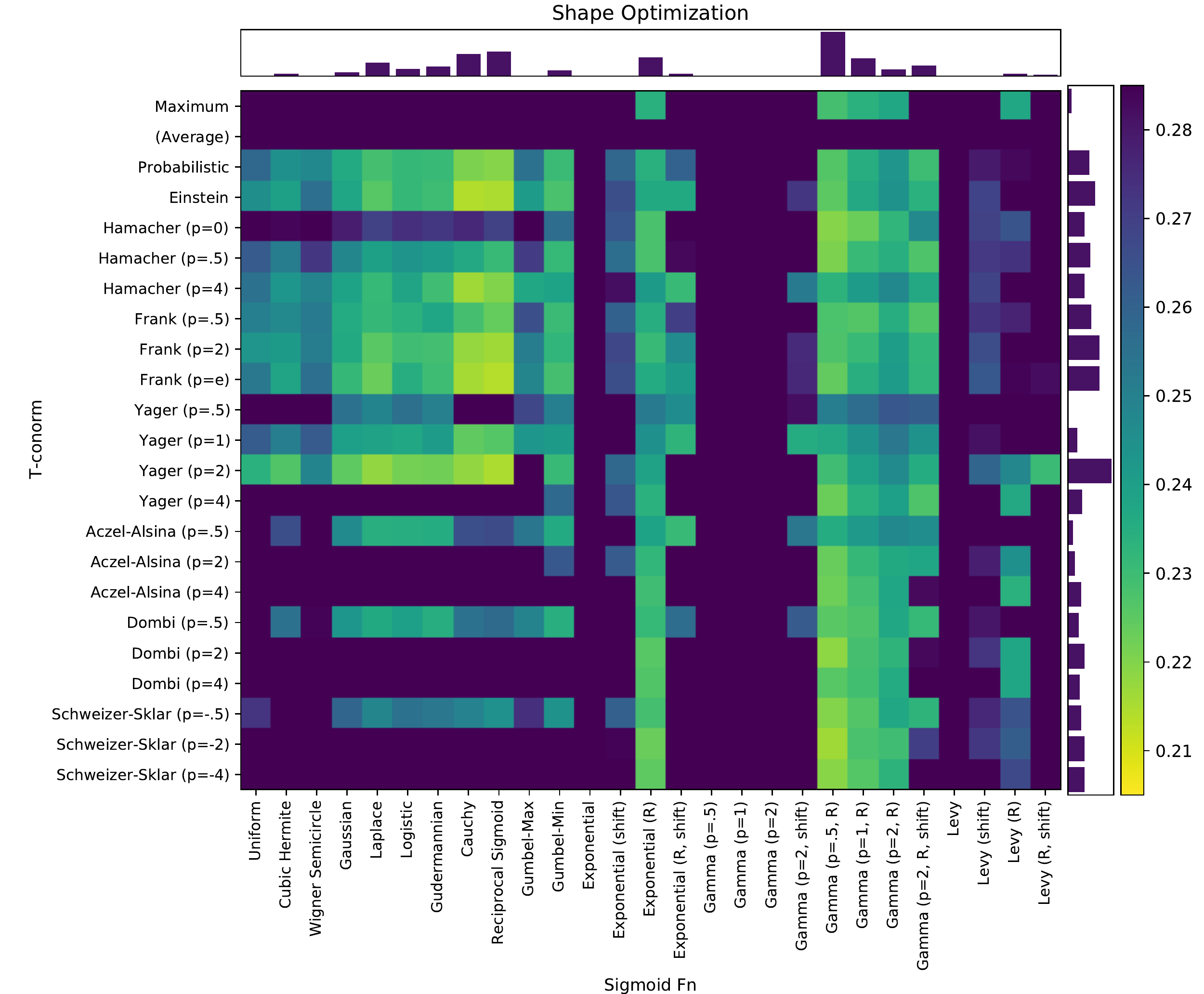}

  \xcaption{
  Results for the 24-view airplane shape optimization task. The optimization is done within a tight budget of 100 steps and the metric is the loss, i.e., lower (=yellow) is better.
  Top: original set of distributions~$\cdf$. Bottom: the respective counter-parts~$\cdfsq$ in the same location.
  The marginal histograms display participation in the top $10\%$ combinations.
  }
  \label{fig:shape-opt-plot}
\end{figure}
\begin{figure}[t]
  \centering
  \includegraphics[width=\linewidth,trim={0.5cm 0.15cm 0.25cm 0.6cm},clip]{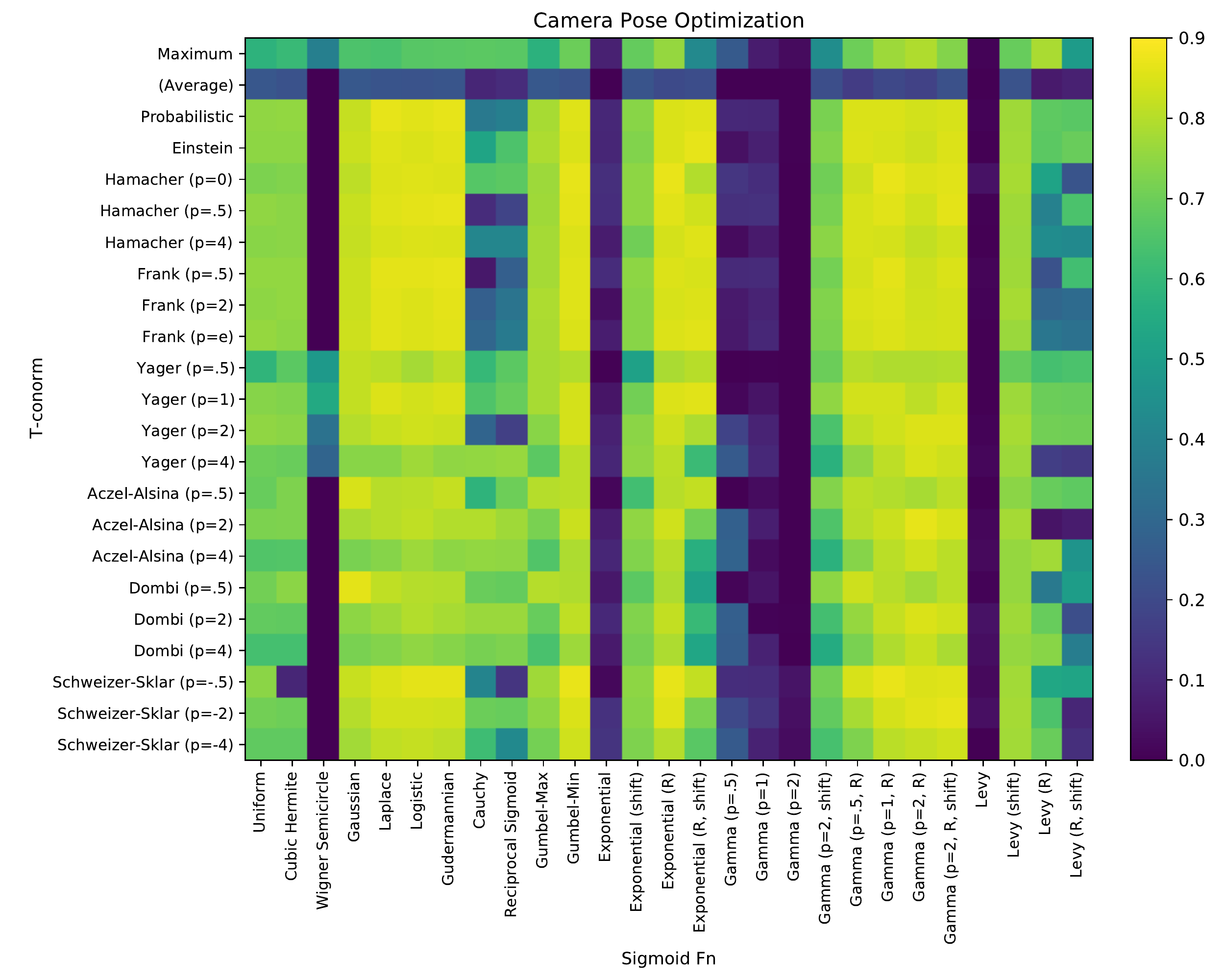}

  \xcaption{
  Results for the tea pot camera pose optimization task. The optimization is done with a temperature $\tau$ that is scheduled to decay.
  The metric is fraction of camera poses recovered, while the initialization angle errors are uniformly sampled from $[15^\circ, 75^\circ]$.
  The figure shows the original set of distributions~$\cdf$, the plot for the respective~$\cdfsq$ can be found in SM~\ref{sm:add-plots}.
  }
  \label{fig:camera-opt-plot}
\end{figure}

\subheading{Results}
In Figure~\ref{fig:shape-opt-plot}, we display the results of our evaluation.
We can observe that the regular distributions~$\cdf$ typically perform better
than the counterpart~$\cdfsq$,
except for the case of Cauchy and reciprocal sigmoid, which are those with a linear convergence rate.
We explain this by the fact that by squaring the distance before applying the sigmoid function,
the function has a quadratic convergence rate instead.
As the linearly converging functions also perform poorly in comparison to the exponentially converging functions
(Gaussian, Laplace, Logistic, Gudermannian), we conclude that linear convergence is inferior to
quadratic and exponential convergence.
Columns~$1-3$ contain the distributions with finite support, and these do not perform very well on this task.
The block of exponentially decaying distributions (columns $4-7$) performs well.
The block of linearly decaying distributions (columns $8-9$) performs badly, as discussed above.
The block of Levy distributions (last $4$ columns) performs even worse because it has an even slower convergence. 
Here, it also becomes slightly better in the squared setting, but it still exhibits worse performance than for linear convergence.

\subheading{Comparison of Distributions}
Gumbel, exponential, and gamma distributions do not all perform equally well, but Gumbel-Min, the reversed exponential, and the reversed gamma are all competitive.
Confer Table~\ref{tab:vis-sigmoids} where it becomes clear that this is because Gumbel-Max, exponential and gamma have all of their mass
inside the triangle, i.e., they yield smaller faces.
This is problematic because in this case, it can cause gaps between neighboring triangles, which hinders optimization.
As the reverse counterparts yield larger faces and do not suffer from this problem, they perform better.
Note that, in this respect, the asymmetrical distributions have an advantage over the symmetrical distributions
because symmetrical distributions always have an accumulated density of~$0.5$ at the edge,
and thus the size of the face stays the same.
We can see that, among the asymmetrical distributions, Gamma performs best.

\subheading{Comparison of T-conorms}
We find that $\bot^M$ and ``average'' (which is not a T-conorm but was used as a baseline in~\cite{Liu2019-SoftRas}) perform poorly.
Also, $\bot^Y_4$, $\bot^A_2$, $\bot^A_4$, $\bot^D_2$, $\bot^D_4$, $\bot^{SS}_{-2}$, and $\bot^{SS}_{-4}$ perform poorly overall.
This can be explained as they are rather extreme members of their respective T-norm families; in all of them, the $p$th power is involved, which can become a problematic component, e.g., $x^4$ is vanishingly small for~$x=0.5$.
Interestingly, the gamma and the exponential distributions still perform well with these,
likely since they are not symmetric and have an accumulated probability of~$1$ on the edge.
Notably, the Yager T-conorm ($p=2$) performs very well, although having a plateau 
and thus no meaningful gradient outside the unit disc,
see Figure~\ref{fig:t-conorms}. %

Finally, we compute histograms of how many times each respective distribution and T-conorm is involved in the best~$10\%$ of overall results. 
This is independent for the top and bottom plots.
We can observe that Gamma ($p=0.5$, Reversed) performs the best overall (because it is more robust to the choice of T-conorm).
Among the T-conorms, we find that $\bot^Y_2$ and $\bot^D_{0.5}$ perform best.
The probabilistic and Einstein sums perform equally, and share the next place. 

\begin{figure*}
    \centering
    \includegraphics[width=\linewidth,trim={0.5cm .0cm 0.5cm 0.cm}]{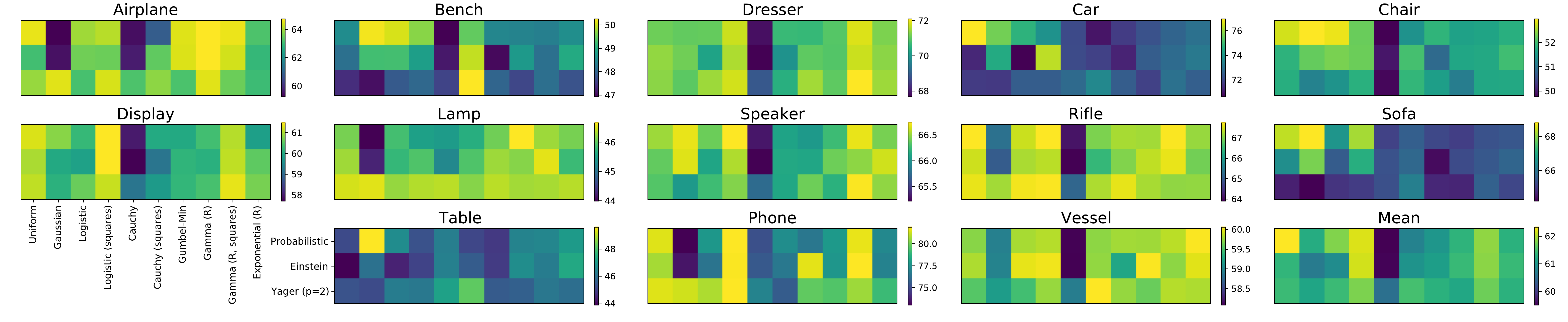}

    \xcaption{
    Single-view reconstruction results for each of the $30$ selected renderers as a 3D IoU (in \%) heatmap for each class.
    While the uniform distribution (w/ $\bot^P$) performs best on average and the square root of logistic (w/ $\bot^P, \bot^E$) performs second-best on average, the optimal setting depends on the characteristics of the respective classes.
    For the `Airplane' class, the Gamma distribution performed best and this is also the distribution that performed best in the airplane shape optimization experiment in Section~\ref{sec:shape-opt}.
    For classes of furniture with legs, such as `Bench', `Chair', and `Table', we find that the Gaussian distribution consistently performs best.
    The pairs of similar classes `Display'+`Phone', `Dresser'+`Speaker', and `Vessel'+`Rifle' also show a similar performance patterns.
    For example, dresser and speakers tend to be cuboid, while rifles and vessels tend to be rather long and slim.
    Considering the Guassian distribution, it is interesting to see that for some classes $\bot^P$ and $\bot^E$ perform better, while for other classes $\bot^Y_2$ performs much better.
    }
    \label{fig:recon-plots}
\end{figure*}
\begin{table*}[]
    \centering
    \resizebox{\textwidth}{!}{
\centering
\setlength{\tabcolsep}{4pt}
\newcommand{\fa}[1]{\bf{#1}}
\begin{tabular}{lllllllllllllll} 
\toprule
Method    & Airplane\kern-1em & Bench  & Dresser\kern-1em & Car    & Chair  & Display\kern-1em & Lamp & Speaker\kern-1em & Rifle & Sofa & Table & Phone & Vessel & \textit{Mean}       \\ 
\midrule
Kato \textit{et al.} 
\cite{Kato2017} N3MR (Uniform Backward)                          & 0.6172   & 0.4998 & 0.7143  & 0.7095 & 0.4990 & 0.5831  & 0.4126 & 0.6536   & 0.6322 & 0.6735  & 0.4829 & 0.7777 & 0.5645  & 0.6015       \\
Liu \textit{et al.} 
\cite{Liu2019-SoftRas} SoftRas (Square-root of Logistic)          & 0.6419   & 0.5080 & 0.7116  & 0.7697 & 0.5270 & 0.6156  & {0.4628} & 0.6654   & {0.6811} & 0.6878  & 0.4487 & 0.7895 & 0.5953  &  0.6234       \\ 
Chen~\etal~\cite{Chen2019DIB} DIB-R (Exponential)                     & 0.570    & 0.498  & 0.763   & 0.788  & 0.527  & 0.588   & 0.403  & 0.726    & 0.561  & 0.677   & 0.508  & 0.743  & 0.609   & 0.612        \\
\midrule
Probabilistic        + Uniform~~($\approx$\cite{Kato2017, Loper2014})              & 0.6456 & 0.4855 & 0.7113 & \fa{0.7696} & 0.5276 & 0.6126 & 0.4611 & 0.6651 & \fa{0.6773} & 0.6835 & 0.4514 & 0.8148 & 0.5971 & \fa{0.6232} \\
Probabilistic        + Logistic~~($=$\cite{petersen2021learning})       & 0.6396 & 0.5005 & 0.7105 & 0.7471 & 0.5288 & 0.6022 & 0.4586 & 0.6639 & 0.6742 & 0.6660 & 0.4666 & 0.7771 & 0.5980 & 0.6179 \\
Probabilistic        + Logistic (squares)~~($=$\cite{Liu2019-SoftRas})  & 0.6416 & 0.4966 & \fa{0.7175} & 0.7386 & 0.5224 & \fa{0.6147} & 0.4550 & \fa{0.6673} & \fa{0.6771} & 0.6818 & 0.4529 & \fa{0.8186} & 0.5984 & 0.6217 \\
Probabilistic        + Exponential (R)~~($=$\cite{Chen2019DIB})         & 0.6321 & 0.4857 & 0.7123 & 0.7298 & 0.5178 & 0.5983 & 0.4611 & 0.6642 & 0.6713 & 0.6546 & 0.4700 & 0.7717 & \fa{0.6005} & 0.6130 \\
Probabilistic        + Gaussian~~($\approx$\cite{rhodin2015versatile})  & 0.5922 & \fa{0.5020} & 0.7104 & 0.7561 & \fa{0.5297} & 0.6080 & 0.4399 & \fa{0.6668} & 0.6533 & \fa{0.6879} & \fa{0.4961} & 0.7301 & 0.5894 & 0.6125 \\
Probabilistic        + Gamma (R)                                        & \fa{0.6473} & 0.4842 & 0.7093 & 0.7220 & 0.5159 & 0.6033 & \fa{0.4665} & 0.6626 & 0.6719 & 0.6505 & 0.4642 & 0.7778 & 0.5978 & 0.6133 \\
Einstein             + Gamma (R, squares)                               & 0.6438 & 0.4816 & \fa{0.7174} & 0.7284 & 0.5170 & 0.6111 & 0.4654 & 0.6647 & 0.6760 & 0.6546 & 0.4626 & \fa{0.8189} & 0.5973 & 0.6184 \\
Yager (p=2)          + Cauchy (squares)                                 & 0.6380 & \fa{0.5026} & 0.7047 & 0.7359 & 0.5188 & 0.5976 & 0.4617 & 0.6612 & 0.6726 & 0.6619 & 0.4819 & 0.7560 & \fa{0.6006} & 0.6149 \\
\bottomrule
\end{tabular}}
    \xcaption{
    Selected single-view reconstruction results measured in 3D IoU. 
    }
\label{tab:recon}
\end{table*}

\subsection{Camera Pose Optimization}
In our second experiment, the goal is to find the camera pose for a model of a \textit{teapot} from a reference image.
The angle is randomly modified by an angle uniformly drawn from $[15^\circ, 75^\circ]$, and the distance and camera view angle are also randomized.
We sample~$600$ pairs of a reference image and an initialization and use this set of settings for each method.
For optimization, we use Adam with a learning rate of either~$0.1$ or~$0.3$ (via grid search)
and optimize for~$1000$ steps.
During the optimization, we transition an initial scale of~$\sigma=10^{-1}$ logarithmically to a final value of~$\sigma=10^{-7}$.
This allows us to avoid a grid search for the optimal scale, and makes sense since an initially large~$\sigma$ is beneficial for pose optimization,
because a smoother model has a higher probability of finding the correct orientation of the object.
This contrasts with the setting of shape estimation,
where this would be fatal because the vertices would collapse to the center.

\subheading{Results}
In Figure~\ref{fig:camera-opt-plot}, we display the results of this experiment.
A corresponding image of the counterpart distributions~$\cdfsq$ as well as results for the experiment applied to the model of a chair can be found in SM~\ref{sm:add-plots}.
The metric is the fraction of settings which achieved matching the ground truth pose up to $3^\circ$.
We find that in this experiment, the results are similar to those in the shape optimization experiment.
Note that there are larger yellow areas because the color map ranges from $0\%$ to $90\%$, while in the shape optimization plot the color map ranges in a rather narrow loss range.

\subsection{Single-View 3D Reconstruction}
\subheading{Setup}
Finally, we reproduce the popular ShapeNet single-view 3D reconstruction experiment from~\cite{Kato2017, Liu2019-SoftRas, Chen2019DIB, petersen2021learning}.
We select three T-conorms ($\bot^P, \bot^E, \bot^Y_2$) and~$10$ distributions (Uniform, Gaussian, Logistic, Logistic (squares), Cauchy, Cauchy (squares), Gumbel-Min, Gamma (R, $p=0.5$), Gamma (R, $p=0.5$, squares), and Exponential (R)).
These have been selected because they have been used in previous works, are notable (Cauchy, Gumbel-Min, Einstein), or have performed especially well in the aircraft shape optimization experiment (Gamma, Yager).
For each setting, we perform a grid search of~$\tau$ at resolution~$10^{0.5}$.
Further experimental details can be found in SM~\ref{sm:implementation-details}.

\subheading{Results}
In Figure~\ref{fig:recon-plots}, we display and discuss the class-wise results for all $30$ selected renderers.
In Table~\ref{tab:recon}, we show the (self-) reported results for existing differentiable renderers in the top block. 
In the bottom block, we display our results for the methods that are equivalent ($=$) or very similar ($\approx$) to the six existing differentiable renderers.
The differences for equivalent methods can be explained with small variations in the setting and minor implementation differences.
Additionally, we include three noteworthy alternative renderers, such as the one that also performed best on the prior airplane shape optimization task.
We conclude that the optimal choice of renderer heavily depends on the characteristics of the 3D models and the task. 
Surprisingly, we find that the simple uniform method achieves consistently good results and the best average score.

\section{Discussion and Conclusion}
In this work, we generalized differentiable mesh renderers and explored a large space of instantiations of our generalized renderer GenDR.
We found that there are significant differences between different distributions for the occlusion test but also between different T-conorms for the aggregation.
In our experiments, we observed that the choice of renderer has a large impact on the kind of models that can be rendered most effectively.
We find that the uniform distribution outperforms the other tested distributions on average, which is surprising considering it simplicity.
Remarkably, the uniform distribution had already been used implicitly for the early surrogate gradient renderers but was later discarded for the approximate differentiable renderers.

\vspace{.7em}
{\small
\textbf{Acknowledgments.} This work was supported by the DFG in the Cluster of Excellence EXC 2117 (Project-ID 390829875) and the SFB Transregio 161 (Project-ID 251654672), and the Land Salzburg within the WISS 2025 project IDA-Lab (20102-F1901166-KZP and 20204-WISS/225/197-2019).\par
}

\clearpage
\printbibliography

\clearpage
\newrefsection
\appendix

\section{Implementation Details}
\label{sm:implementation-details}

For the single-view 3D reconstruction experiment, we closely orient ourselves on the setup by Liu~\etal~\cite{Liu2019-SoftRas}.
We use the same model architecture~\cite{Liu2019-SoftRas} and also train with a batch size of $64$ for $250\,000$ steps using the Adam optimizer~\cite{Kingma2014AdamOpt}
We also schedule the learning rate to $10^{-4}$ for the first $150\,000$ steps and use a learning rate of $3\cdot10^{-5}$ for the remaining training.
At this point (after the first $150\,000$ steps), we also decrease the temperature $\tau$ by a factor of $0.3$.

Using different learning rates (as an ablation) did not improve the results.

\section{Distributions}
\label{sm:dist}

In this section, we define each of the presented distributions / sigmoid functions. 
Figure~\ref{tab:vis-pdf-cdf} displays the respective CDFs and PDFs.

Note that, for each distribution, the PDFs $\density$ is defined as the derivative of the CDF $\cdf$.
Also, note that a reversed (Rev.) CDF is defined as $\cdfrev(x) = 1-\cdf(-x)$, which means that $\cdfrev = \cdf$ for symmetric distributions.
The square-root distribution~$\cdfsq$ is defined in terms of $\cdf$ as in Equation (\ref{eq:cdfsq}).
Therefore, in the following, we will define the distributions via their CDFs~$\cdf$.\\

\newcommand{\disttitle}[1]{\bigskip\smallskip\noindent\textbf{#1}}

\disttitle{Heaviside}
\begin{equation}
    x\mapsto\begin{cases}
    0 & \mathrm{if}~x < 0 \\
    1 & \mathrm{otherwise}
    \end{cases}
\end{equation}

\disttitle{Uniform}
\begin{equation}
    x\mapsto\begin{cases}
    0 & \mathrm{if}~x < -1 \\
    0.5 \cdot (1 + x) & \mathrm{if}~-1 \leq x \leq 1 \\
    1 & \mathrm{otherwise}
    \end{cases}
\end{equation}

\disttitle{Cubic Hermite}
\begin{equation}
    x\mapsto\begin{cases}
    0 & \mathrm{if}~x < -1 \\
    3 y^2 - 2 y^3 & \mathrm{if}~-1 \leq x \leq 1 \\
    1 & \mathrm{otherwise}
    \end{cases}
\end{equation}
where $y:=(x+1)/2$.

\disttitle{Wigner Semicircle}
\begin{equation}
    x\mapsto\begin{cases}
    0 & \mathrm{if}~x < -1 \\
    \frac12+\frac{x\sqrt{1-x^2}}{\pi} + \frac{\arcsin (x)}{\pi} & \mathrm{if}~-1 \leq x \leq 1 \\
    1 & \mathrm{otherwise}
    \end{cases}
\end{equation}

\disttitle{Gaussian}
\begin{equation}
    x\mapsto 
    \frac12\left(
        1 + \operatorname{erf} \left( \frac{x}{\sqrt{2}} \right)
    \right)
\end{equation}

\disttitle{Laplace}
\begin{equation}
    x\mapsto \begin{cases} \frac{1}{2} \exp \left(x\right) & \text{if }x\leq 0 \\ 1-{\frac {1}{2}}\exp \left(-x\right) & \text{if }x\geq 0 \end{cases}
\end{equation}

\disttitle{Logistic}
\begin{equation}
    x\mapsto \frac{1}{1 + \exp (-x)}
\end{equation}

\disttitle{Hyperbolic secant / Gudermannian}
\begin{equation}
    x\mapsto \frac{2}{\pi} \arctan \left( \exp \left( \frac{\pi}{2} x \right)\right)
\end{equation}

\disttitle{Cauchy}
\begin{equation}
    x\mapsto \frac{1}{\pi} \arctan\left( x \right) + \frac{1}{2}
\end{equation}

\disttitle{Reciprocal}
\begin{equation}
    x\mapsto x / (2 + 2|x|) + 1/2
\end{equation}

\disttitle{Gumbel-Max}
\begin{equation}
    x\mapsto e^{-e^{-x}}
\end{equation}

\disttitle{Gumbel-Min}
\begin{equation}
    x\mapsto e^{-e^x}
\end{equation}

\disttitle{Exponential}
\begin{equation}
    x\mapsto 1 -e^{-x}
\end{equation}

\disttitle{Levy}
\begin{equation}
    x\mapsto 2 -2\Phi \left(\sqrt{\frac{1}{x}}\right)
\end{equation}
where $\Phi$ is the CDF of the standard normal distribution.

\disttitle{Gamma}
\begin{equation}
    x\mapsto \frac{1}{\Gamma(p)} \gamma(p,x)
\end{equation}
where $\gamma(p,x)$ is the lower incomplete gamma function and $p>0$ is the shape parameter.

\begin{table*}[]
    \centering
    \resizebox{\linewidth}{!}{
    \scriptsize
    \begin{tabular}{cccccc}
        \toprule
        \includegraphics[]{img/diags/diags-0.pdf} &
        \includegraphics[]{img/diags/diags-1.pdf} &
        \includegraphics[]{img/diags/diags-3.pdf} &
        \includegraphics[]{img/diags/diags-5.pdf} &
        \includegraphics[]{img/diags/diags-7.pdf} &
        \includegraphics[]{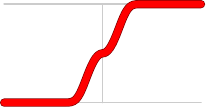}
        \\
        \includegraphics[]{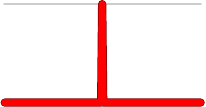} &
        \includegraphics[]{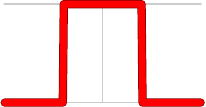} &
        \includegraphics[]{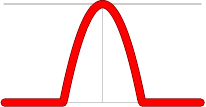} &
        \includegraphics[]{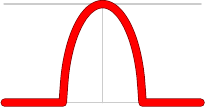} &
        \includegraphics[]{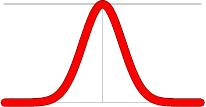} &
        \includegraphics[]{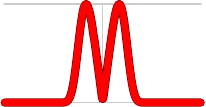}
        \\
        Heaviside &
        Uniform &
        Cubic Hermite &
        Wigner Semicircle &
        Gaussian &
        Gaussian (sq.)
        \\
        \midrule
        \includegraphics[]{img/diags/diags-9.pdf} &
        \includegraphics[]{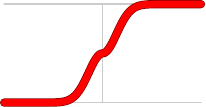} &
        \includegraphics[]{img/diags/diags-11.pdf} &
        \includegraphics[]{img/diags/diags-12.pdf} &
        \includegraphics[]{img/diags/diags-13.pdf} &
        \includegraphics[]{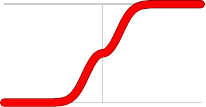} 
        \\
        \includegraphics[]{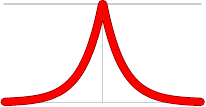} &
        \includegraphics[]{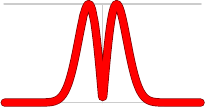} &
        \includegraphics[]{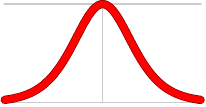} &
        \includegraphics[]{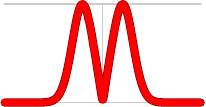} &
        \includegraphics[]{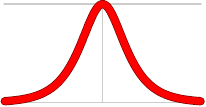} &
        \includegraphics[]{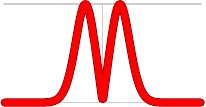} 
        \\
        Laplace &
        Laplace (sq.) &
        Logistic &
        Logistic (sq.) &
        Hyperbolic secant &
        Hyperbolic secant (sq.)
        \\
        \midrule
        \includegraphics[]{img/diags/diags-15.pdf} &
        \includegraphics[]{img/diags/diags-16.pdf} &
        \includegraphics[]{img/diags/diags-17.pdf} &
        \includegraphics[]{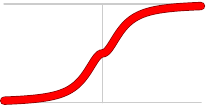} &
        \includegraphics[]{img/diags/diags-19.pdf} &
        \includegraphics[]{img/diags/diags-21.pdf}
        \\
        \includegraphics[]{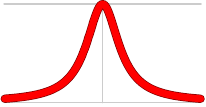} &
        \includegraphics[]{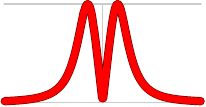} &
        \includegraphics[]{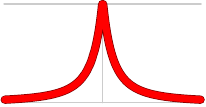} &
        \includegraphics[]{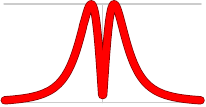} &
        \includegraphics[]{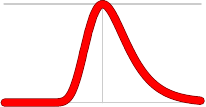} &
        \includegraphics[]{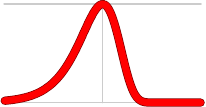}
        \\
        Cauchy &
        Cauchy (sq.) &
        Reciprocal &
        Reciprocal (sq.) &
        Gumbel-Max &
        Gumbel-Min 
        \\
        \midrule
        \includegraphics[]{img/diags/diags-23.pdf} &
        \includegraphics[]{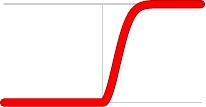} &
        \includegraphics[]{img/diags/diags-27.pdf} &
        \includegraphics[]{img/diags/diags-47.pdf} &
        \includegraphics[]{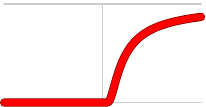} &
        \includegraphics[]{img/diags/diags-51.pdf} 
        \\
        \includegraphics[]{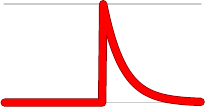} &
        \includegraphics[]{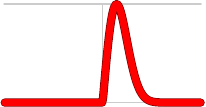} &
        \includegraphics[]{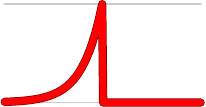} &
        \includegraphics[]{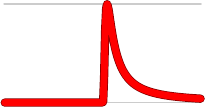} &
        \includegraphics[]{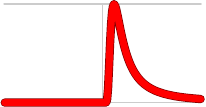} &
        \includegraphics[]{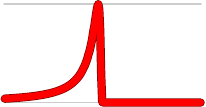} 
        \\
        Exponential &
        Exponential (sq.) &
        Exponential (Rev.) &
        Levy &
        Levy (sq.) &
        Levy (Rev.) 
        \\
        \midrule 
        \includegraphics[]{img/diags/diags-37.pdf} &
        \includegraphics[]{img/diags/diags-35.pdf} &
        \includegraphics[]{img/diags/diags-33.pdf} &
        \includegraphics[]{img/diags/diags-45.pdf} &
        \includegraphics[]{img/diags/diags-43.pdf} &
        \includegraphics[]{img/diags/diags-41.pdf} 
        \\
        \includegraphics[]{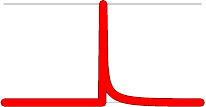} &
        \includegraphics[]{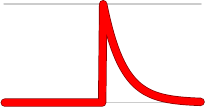} &
        \includegraphics[]{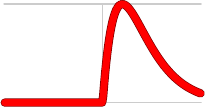} &
        \includegraphics[]{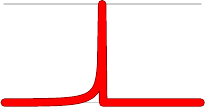} &
        \includegraphics[]{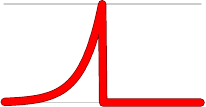} &
        \includegraphics[]{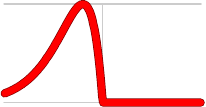} 
        \\
        Gamma ($p{=}0.5$) &
        Gamma ($p{=}1$) &
        Gamma ($p{=}2$) &
        Gamma ($p{=}.5$, R.) &
        Gamma ($p{=}1$, R.) &
        Gamma ($p{=}2$, R.) 
        \\
        \midrule
        \includegraphics[]{img/diags/diags-38.pdf} &
        \includegraphics[]{img/diags/diags-36.pdf} &
        \includegraphics[]{img/diags/diags-34.pdf} &
        \includegraphics[]{img/diags/diags-46.pdf} &
        \includegraphics[]{img/diags/diags-44.pdf} &
        \includegraphics[]{img/diags/diags-42.pdf} 
        \\
        \includegraphics[]{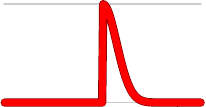} &
        \includegraphics[]{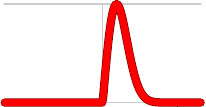} &
        \includegraphics[]{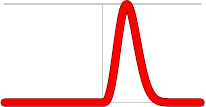} &
        \includegraphics[]{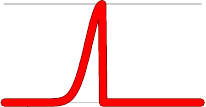} &
        \includegraphics[]{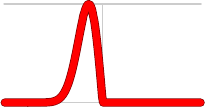} &
        \includegraphics[]{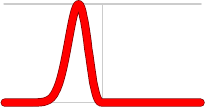} 
        \\
        Gamma ($p{=}0.5$, sq.) &
        Gamma ($p{=}1$, sq.) &
        Gamma ($p{=}2$, sq.) &
        Gamma ($p{=}.5$, R., sq.) &
        Gamma ($p{=}1$, R., sq.) &
        Gamma ($p{=}2$, R., sq.) 
        \\
        \bottomrule
    \end{tabular}
    }
    \xcaption{Visualization of CDFs (top) and PDFs (bottom) for different distributions.}
    \label{tab:vis-pdf-cdf}
\end{table*}

\newpage~\newpage~\newpage

\section{T-Norms and T-Conorms}
\label{sm:tcn}

The axiomatic approach to multi-valued logics (which we need to combine
the occlusions by different faces in a ``soft'' manner) is based on
defining reasonable properties for truth functions. We stated the axioms
for multi-valued generalizations of the disjunction (logical
``or''), called T-conorms, in Definition~\ref{def:t-conorm}. Here we
complement this with the axioms for multi-valued generalizations of the
conjunction (logical ``and''), which are called T-norms.
\begin{definition}[T-norm]
A T-norm (triangular norm) is a binary
operation~$\top: [0,1] \times [0,1] \to [0,1]$,
which satisfies
\begin{itemize}\itemsep0pt
\item associativity:
      $\top(a, \top(b,c)) = \top(\top(a,b), c)$,
\item commutativity:
      $\top(a,b) = \top(b,a)$,
\item monotonicity:
      $(a \leq c) \land (b \leq d)
       \Rightarrow \top(a,b) \leq \top(c,d)$,
\item $1$ is a neutral element:
      $\top(a,1) = a$.
\end{itemize}
\end{definition}
Clearly these axioms ensure that the corners of the unit square, that
is, the value pairs considered in classical logic, are processed as
with a standard conjunction: neutral element and commutativity imply
that $(1,1) \mapsto 1$, $(0,1) \mapsto 0$, $(1,0) \mapsto 0$. From
one of the latter two and monotonicity it follows $(0,0) \mapsto 0$.
Analogously, the axioms of T-conorms ensure that the corners of the
unit square are processed as with a standard disjunction. Actually,
the axioms already fix the values not only at the corners, but on
the boundaries of the unit square. Only inside the unit square (that
is, for $(0,1)^2$) T-norms (as well as T-conorms) can differ.

\begin{table}[h]
\centering
\begin{tabular}{|l|c@{~$=$~}l|}\hline\rule{0pt}{2.6ex}%
Minimum         & $\top^M(a,b)$
  & $\min(a,b)$                  \\[1.2ex]
Probabilistic   & $\top^P(a,b)$
  & $ab$                         \\[1.2ex]
Einstein        & $\top^E(a,b)$
  & $\frac{ab}{2-a-b+ab}$        \\[1.2ex]
Hamacher        & $\top^H_p(a,b)$
  & $\frac{ab}{p+(1-p)(a+b-ab)}$ \\[1.2ex]
Frank           & $\top^F_p(a,b)$
  & $\log_p\left(1+\frac{(p^a-1)(p^b-1)}{p-1}\right)$ \\[1.2ex]
Yager           & $\top^Y_p(a,b)$
  & $\max\left(0, 1-\left(\left(1-a\right)^p+\left(1-b\right)^p\right)^{\frac{1}{p}}\right)$ \\[1.2ex]
Acz\'el-Alsina  & $\top^A_p(a,b)$
  & $\exp\big(-\left(|\log(a)|^p+|\log(b)|^p
               \right)^{\frac{1}{p}}\big)$ \\[0.3ex]
Dombi           & $\top^D_p(a,b)$
  & $\Big(1+\left( \left(\frac{1-a}{a}\right)^p
                  +\left(\frac{1-b}{b}\right)^p
           \right)^{\frac{1}{p}}\Big)^{\!\!-1}$ \\[1.2ex]
Schweizer-Sklar & $\top^S_p(a,b)$
  & $(a^p+b^p-1)^{\frac{1}{p}}$ \\[1ex] \hline
\end{tabular}
\caption{\label{tab.t-norms}(Families of) T-norms.\kern-5.4em}
\end{table}

\newpage
In the theory of multi-valued logics, and especially in fuzzy logic \cite{Klir_and_Yuan_1995},
it was established that the largest possible T-norm is the minimum
and the smallest possible T-conorm is the maximum: for any
T-norm~$\top$ it is $\top(a,b) \le \min(a,b)$ and for any
T-conorm~$\bot$ it is $\bot(a,b) \ge \max(a,b)$. The other extremes,
that is, the smallest possible T-norm and the largest possible T-conorm
are the so-called drastic T-norm, defined as $\top^\circ(a,b) = 0$
for $(a,b) \in (0,1)^2$, and
the drastic T-conorm, defined as
$\bot^\circ(a,b) = 1$ for $(a,b) \in (0,1)^2$. Hence it is
$\top(a,b) \ge \top^\circ(a,b)$ for any T-norm~$\top$ and
$\bot(a,b) \le \bot^\circ(a,b)$ for any T-conorm~$\bot$.
We do not consider the drastic T-conorm for an occlusion test, because it clearly does not
yield useful gradients.

As already mentioned in the paper, it is common to combine a
T-norm~$\top$, a T-conorm~$\bot$ and a negation~$N$ (or complement,
most commonly $N(a) = 1-a$) so that DeMorgan's laws hold. Such a
triplet is often called a {\em dual triplet}.
In Tables~\ref{tab.t-norms} and~\ref{tab.t-conorms} we show the
formulas for the families of T-norms and T-conorms, respectively,
where matching lines together with the standard negation $N(a) = 1-a$
form dual triplets. Note that, for some families, we limited the range
of values for the parameter~$p$ (see Table~\ref{tab:t-conorms})
compared to more general definitions~\cite{Klir_and_Yuan_1995}.

\subsection{T-conorm Plots}

\begin{figure*}[h]
  \centering
  \includegraphics[width=.24\linewidth,trim={.25cm .0cm .25cm .0cm},clip]{img/t-conorms/t_conorm_0} \hfill
  \includegraphics[width=.24\linewidth,trim={.25cm .0cm .25cm .0cm},clip]{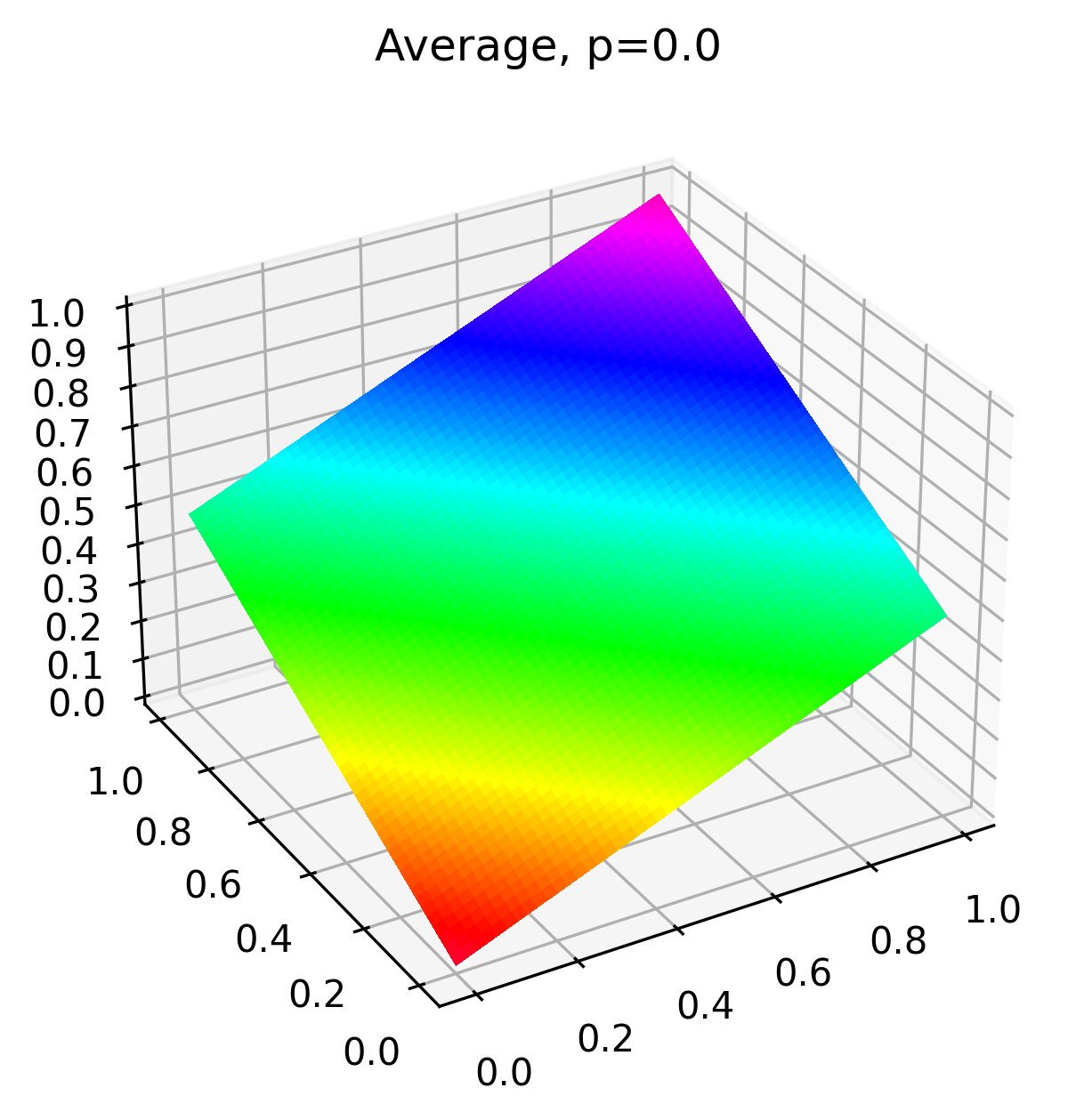} \hfill
  \includegraphics[width=.24\linewidth,trim={.25cm .0cm .25cm .0cm},clip]{img/t-conorms/t_conorm_2} \hfill
  \includegraphics[width=.24\linewidth,trim={.25cm .0cm .25cm .0cm},clip]{img/t-conorms/t_conorm_3} \hfill
  \includegraphics[width=.24\linewidth,trim={.25cm .0cm .25cm .0cm},clip]{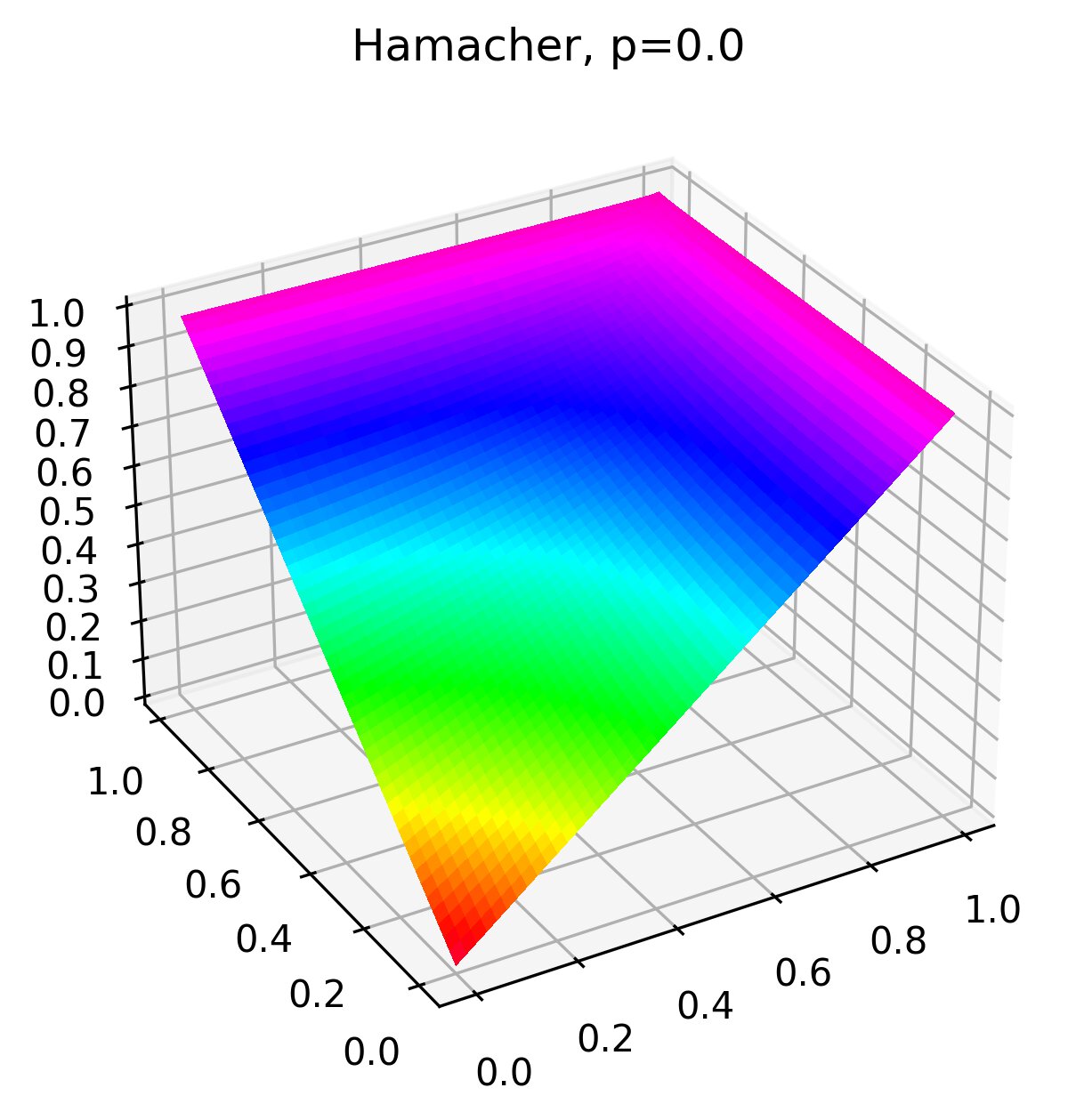} \hfill
  \includegraphics[width=.24\linewidth,trim={.25cm .0cm .25cm .0cm},clip]{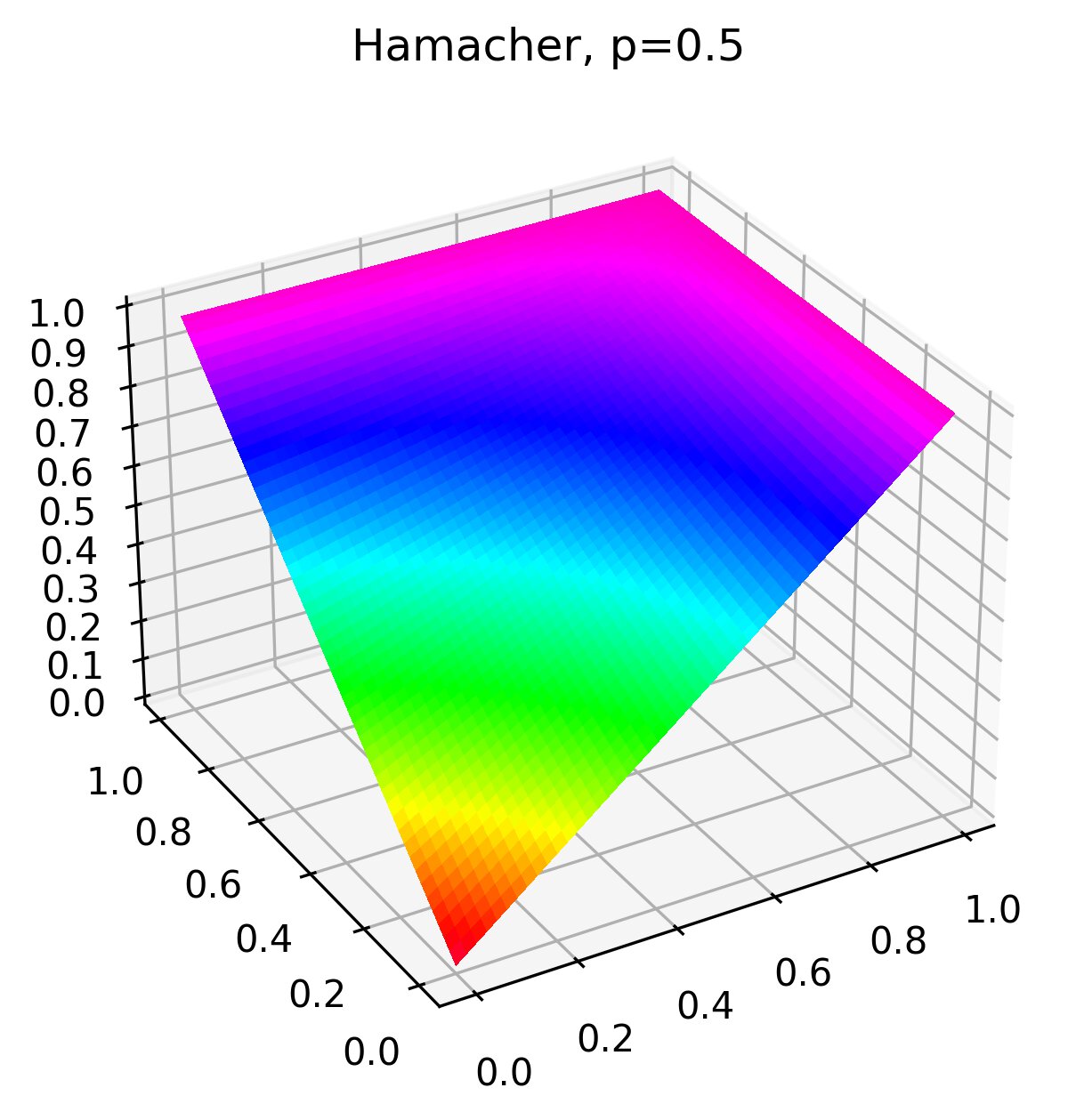} \hfill
  \includegraphics[width=.24\linewidth,trim={.25cm .0cm .25cm .0cm},clip]{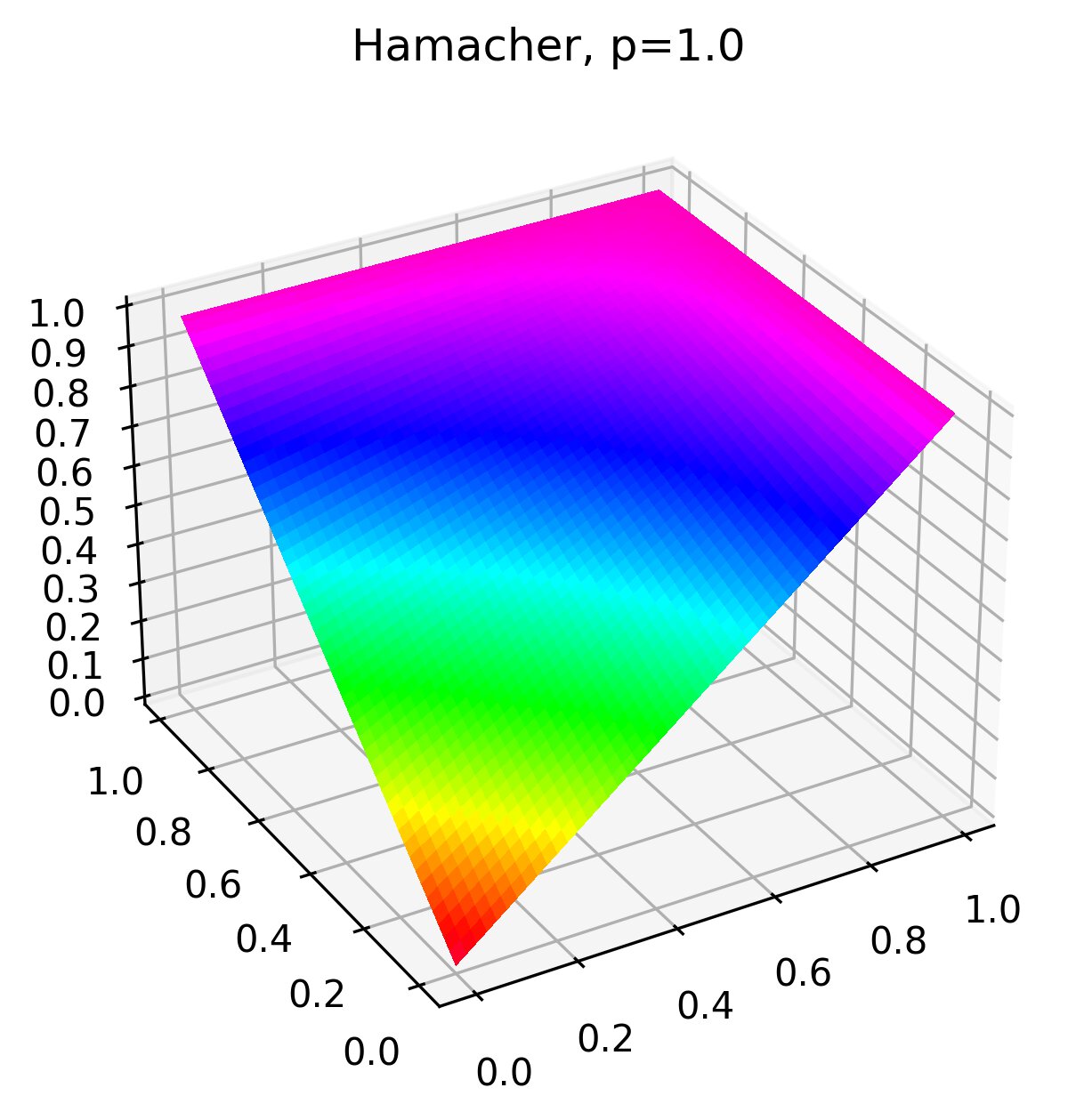} \hfill
  \includegraphics[width=.24\linewidth,trim={.25cm .0cm .25cm .0cm},clip]{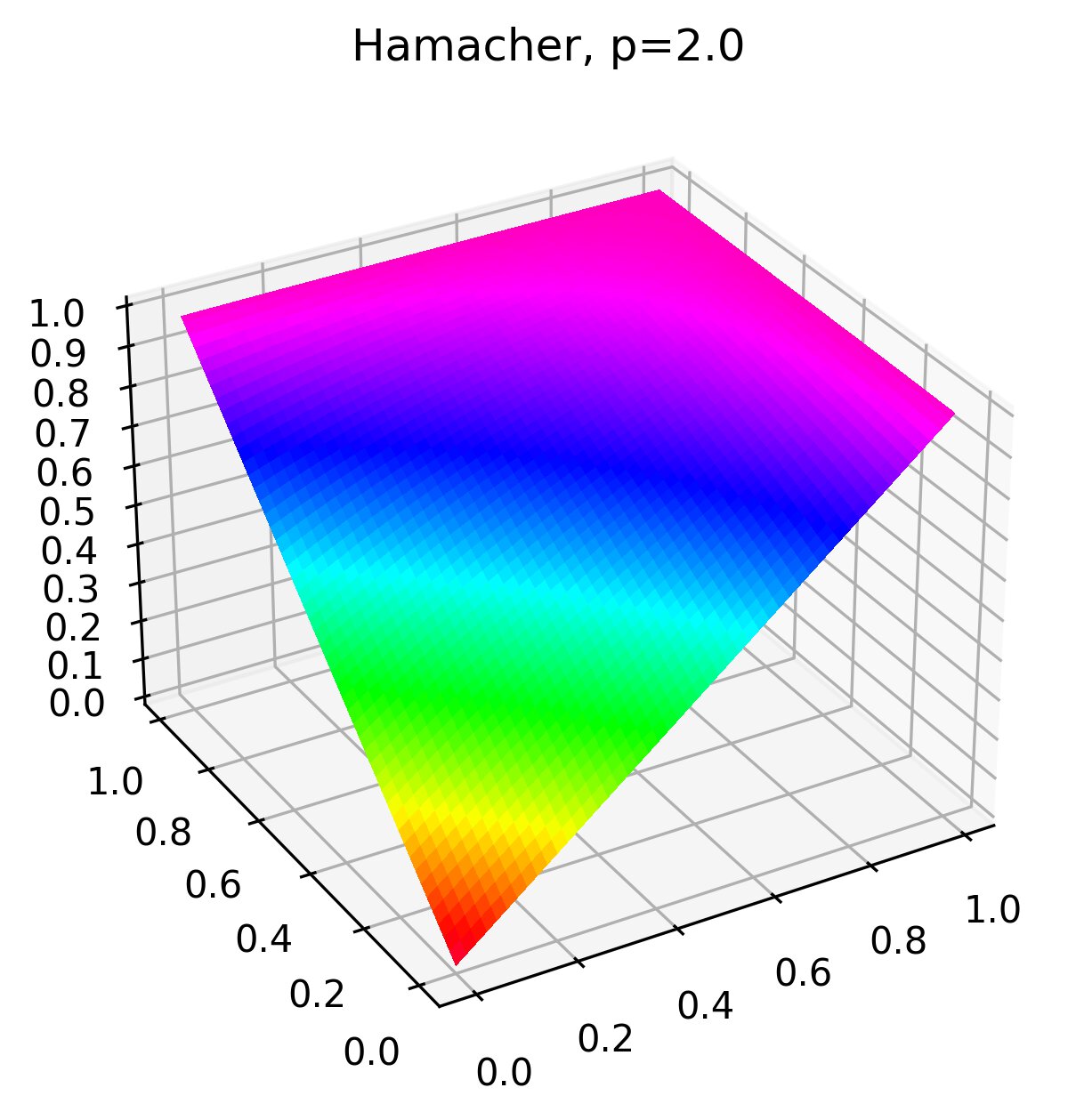} \hfill
  \includegraphics[width=.24\linewidth,trim={.25cm .0cm .25cm .0cm},clip]{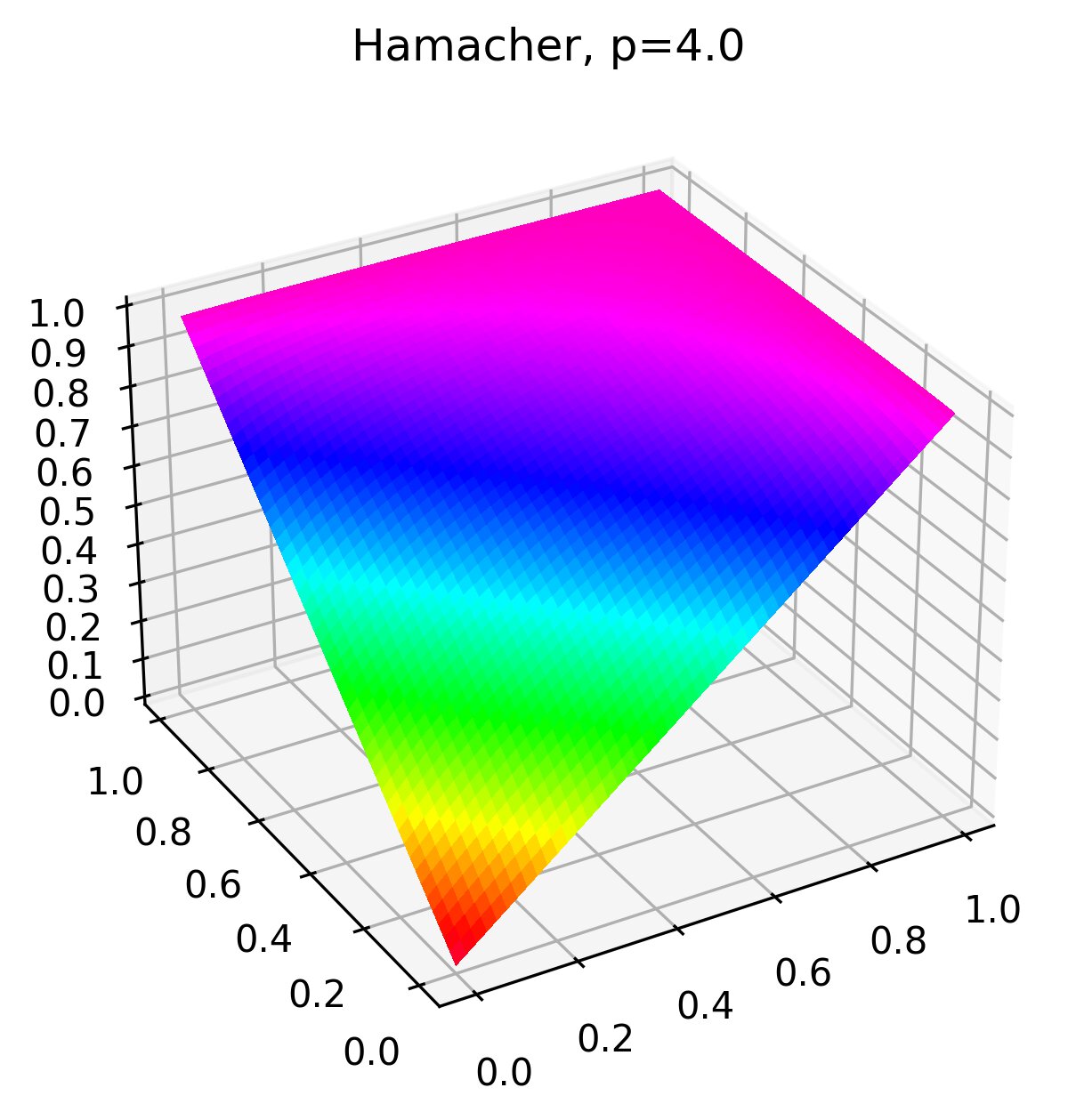} \hfill
  \includegraphics[width=.24\linewidth,trim={.25cm .0cm .25cm .0cm},clip]{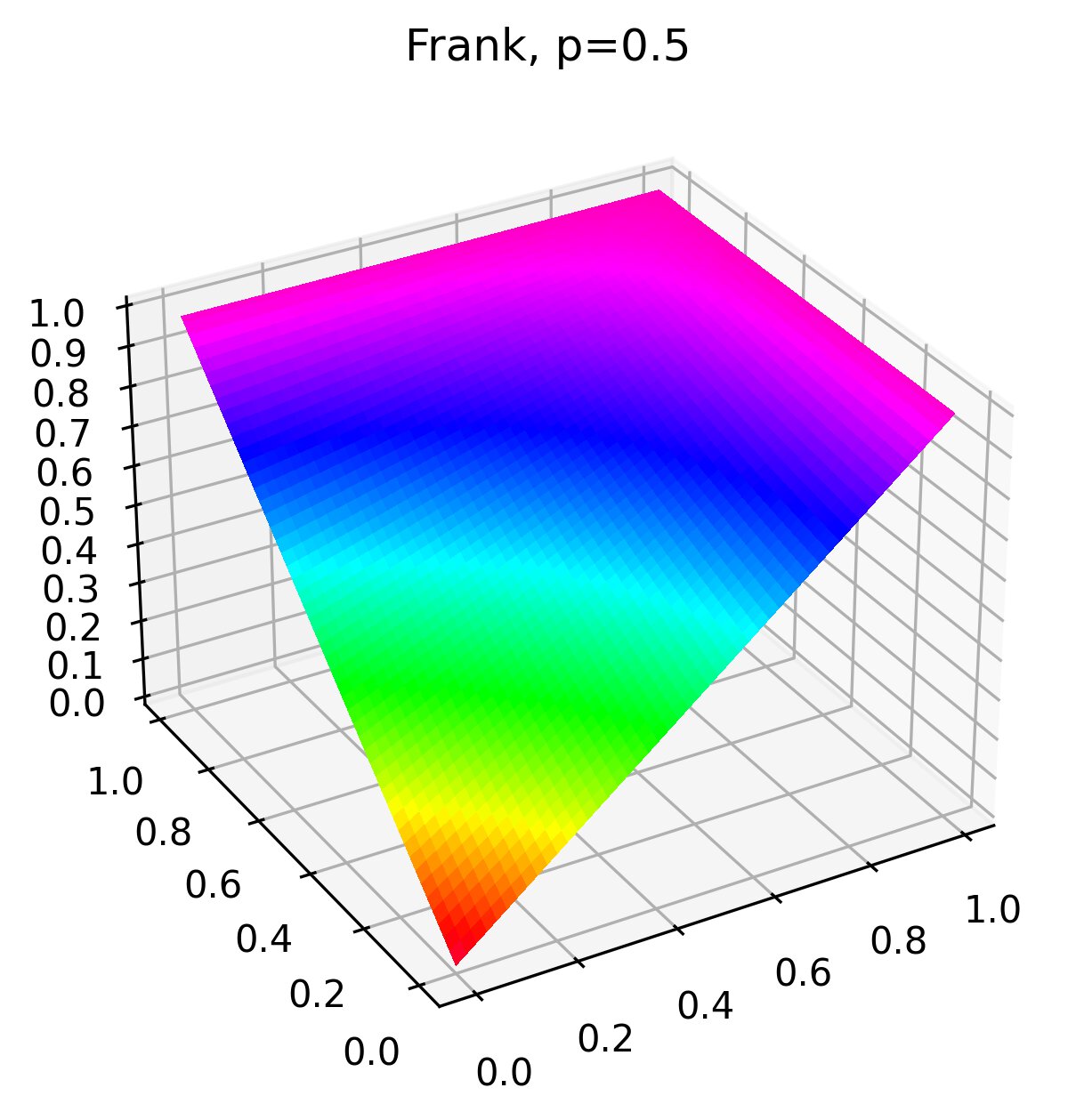} \hfill
  \includegraphics[width=.24\linewidth,trim={.25cm .0cm .25cm .0cm},clip]{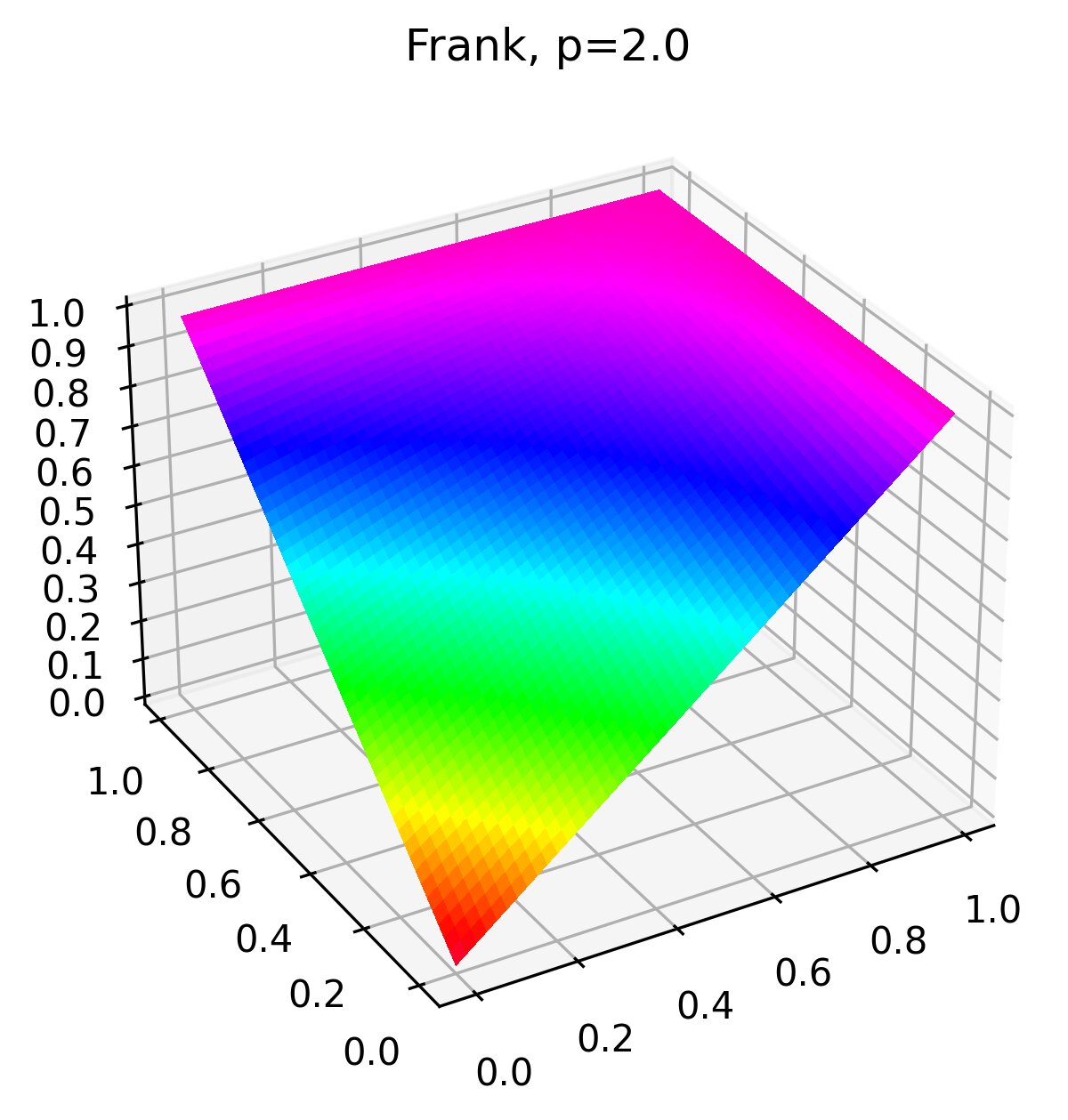} \hfill
  \includegraphics[width=.24\linewidth,trim={.25cm .0cm .25cm .0cm},clip]{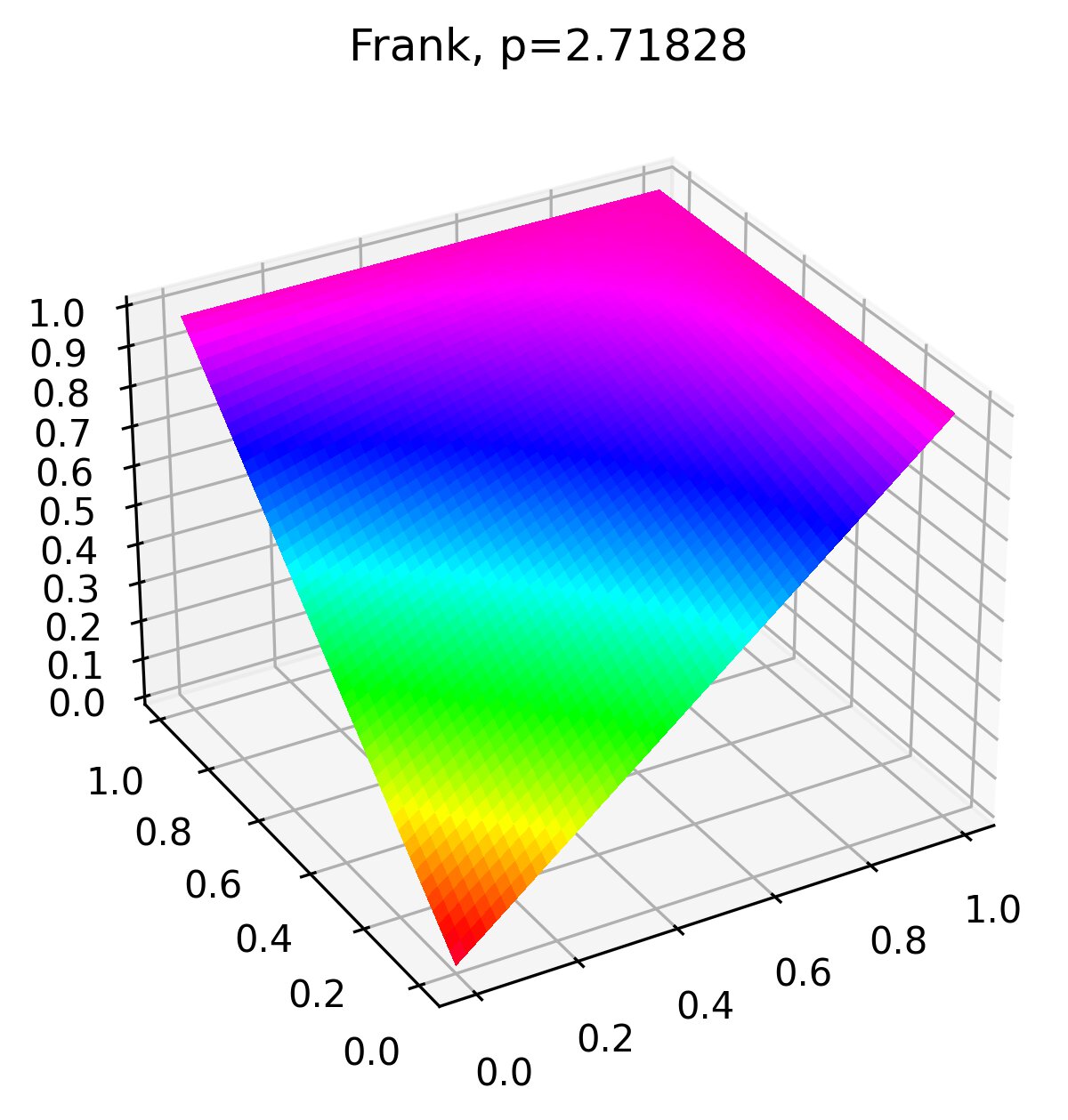} \hfill
  \caption{T-conorm plots (1/2). Note that `Average' is not a T-cornom and just included for reference. Also, Note how `Probabilistic' is equal to `Hamacher $p=1$' and `Einstein' is equal to `Hamacher $p=2$'.}
  \label{fig:t-plot-1}
\end{figure*}

\begin{figure*}
  \centering
  \includegraphics[width=.24\linewidth,trim={.25cm .0cm .25cm .0cm},clip]{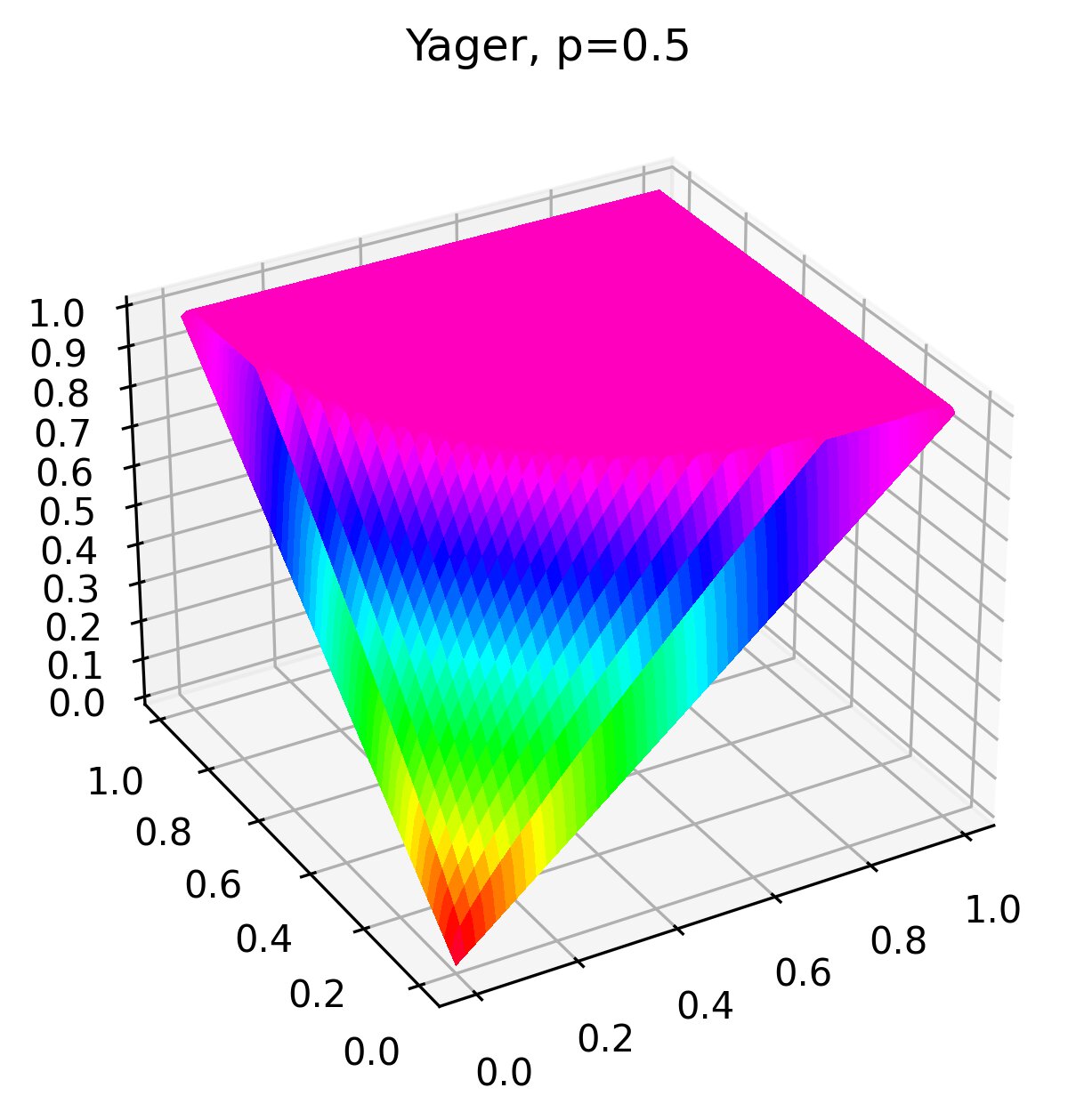} \hfill
  \includegraphics[width=.24\linewidth,trim={.25cm .0cm .25cm .0cm},clip]{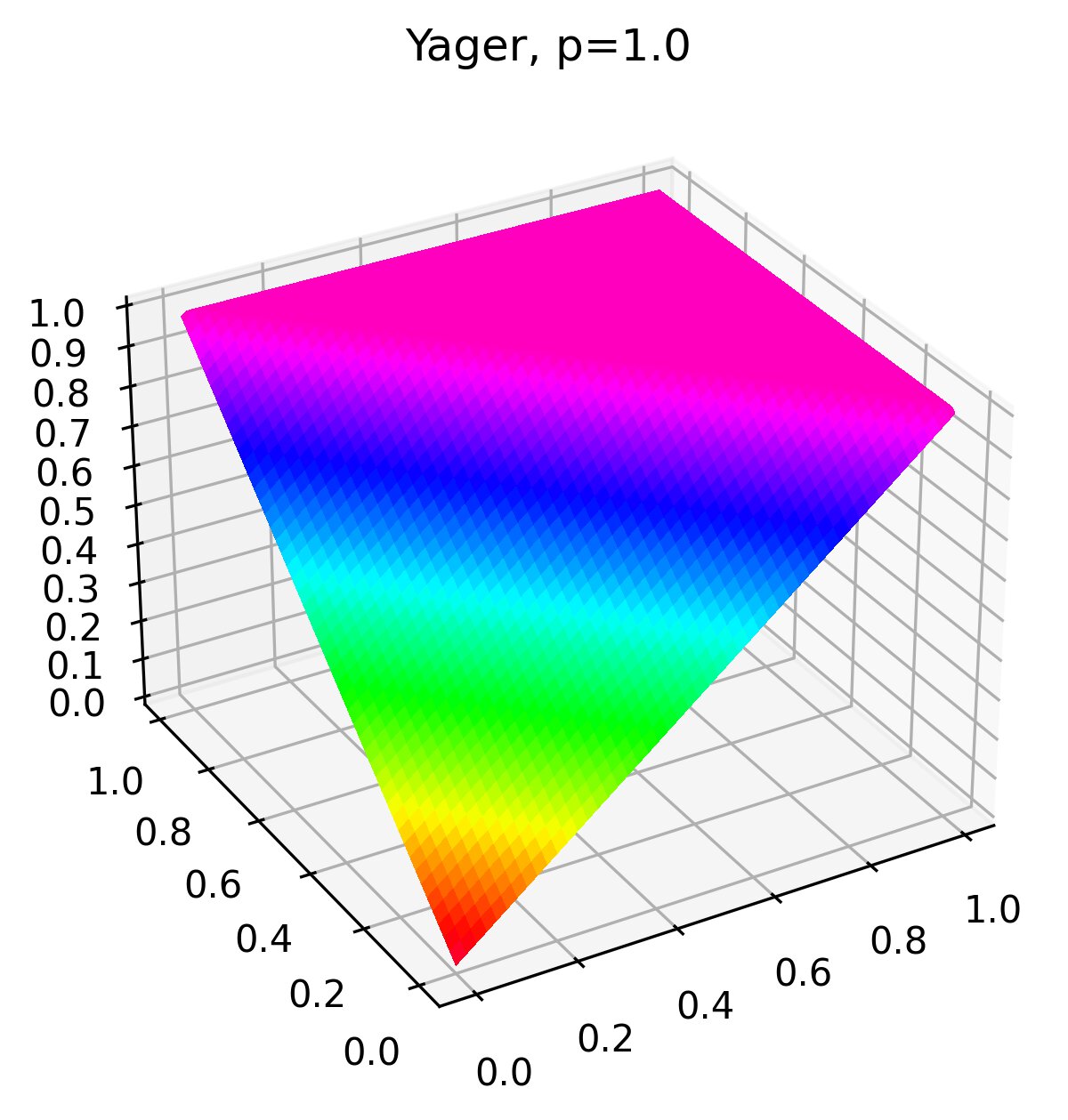} \hfill
  \includegraphics[width=.24\linewidth,trim={.25cm .0cm .25cm .0cm},clip]{img/t-conorms/t_conorm_14} \hfill
  \includegraphics[width=.24\linewidth,trim={.25cm .0cm .25cm .0cm},clip]{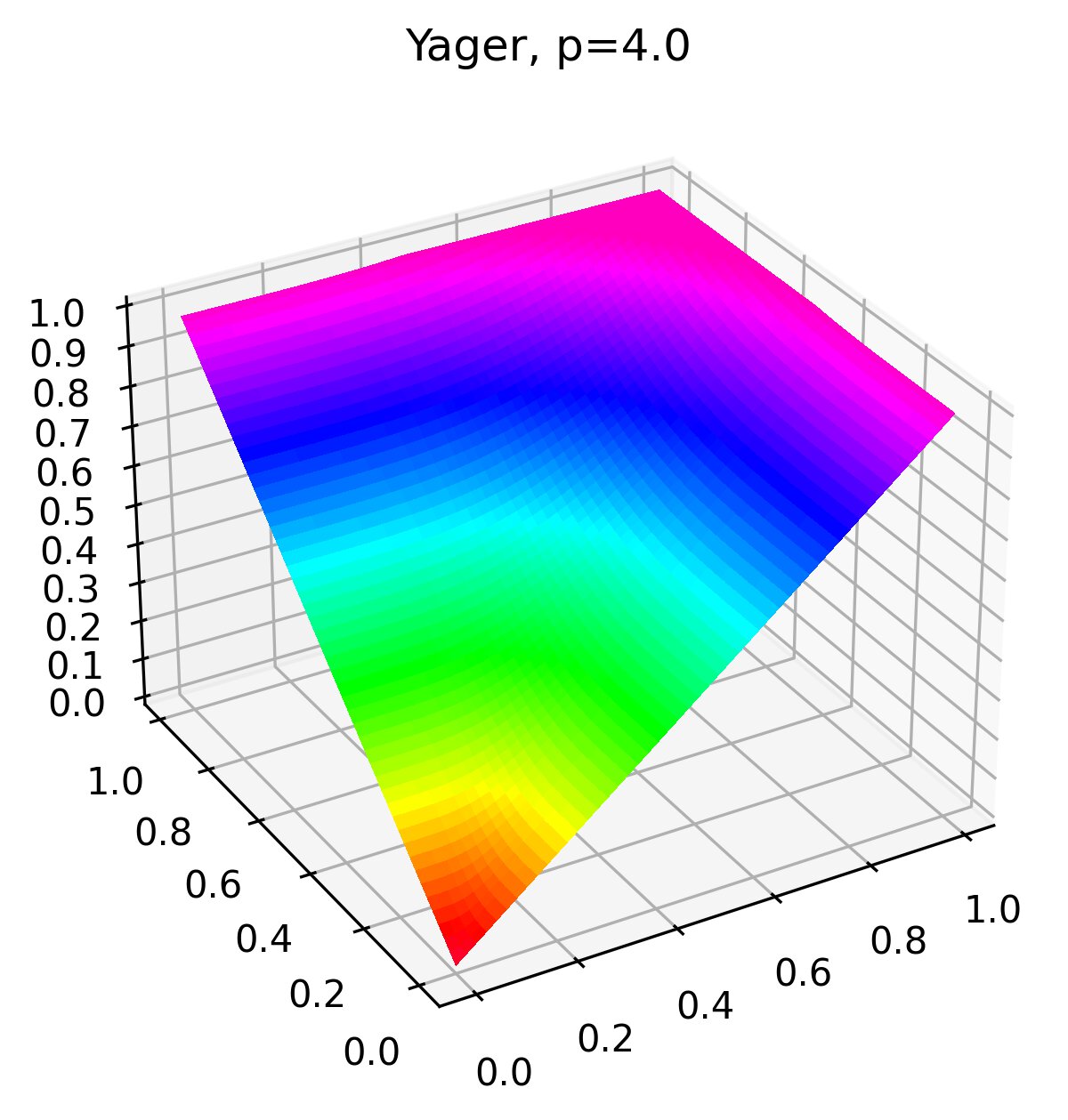} \hfill
  \includegraphics[width=.24\linewidth,trim={.25cm .0cm .25cm .0cm},clip]{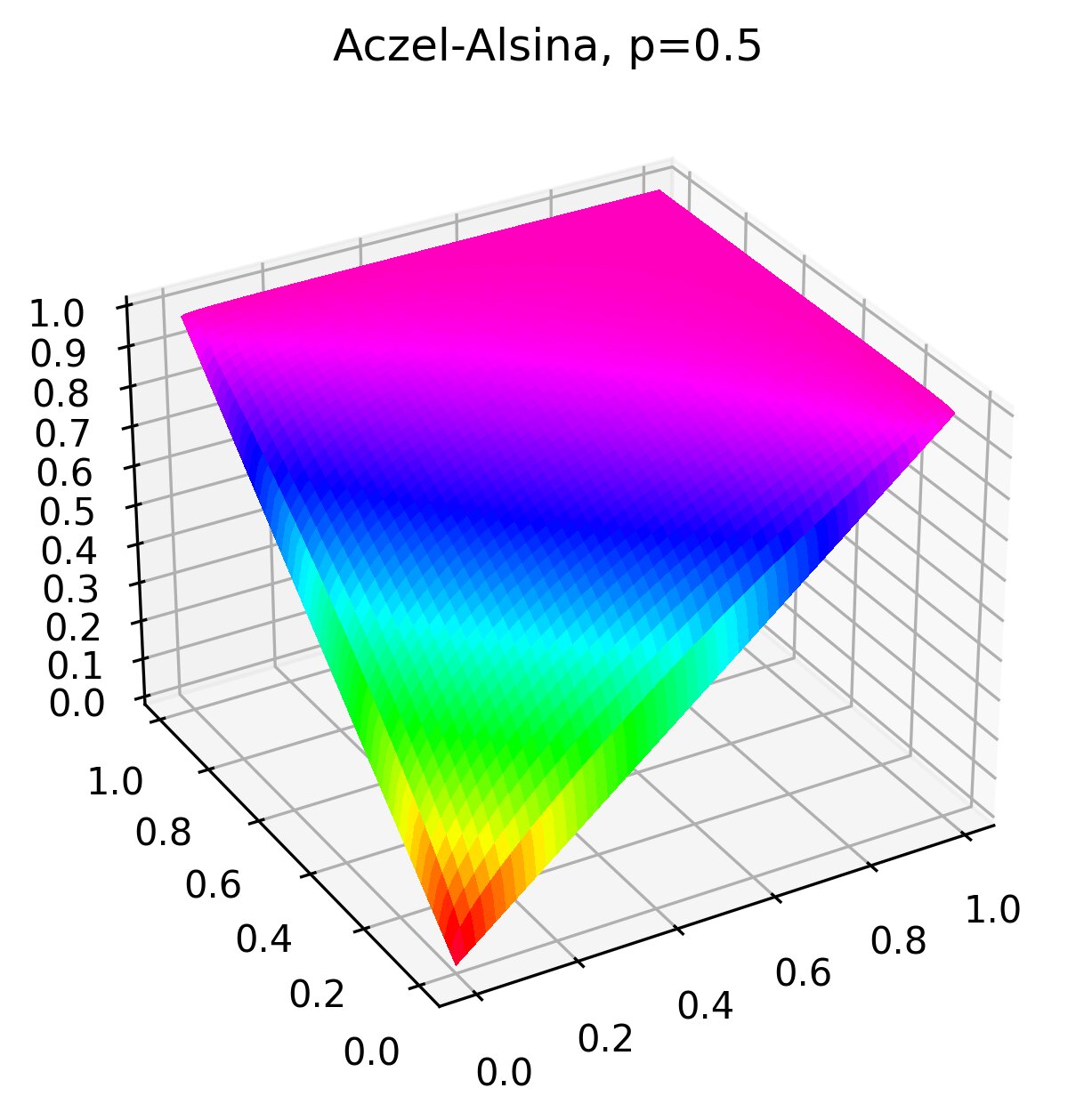} \hfill
  \includegraphics[width=.24\linewidth,trim={.25cm .0cm .25cm .0cm},clip]{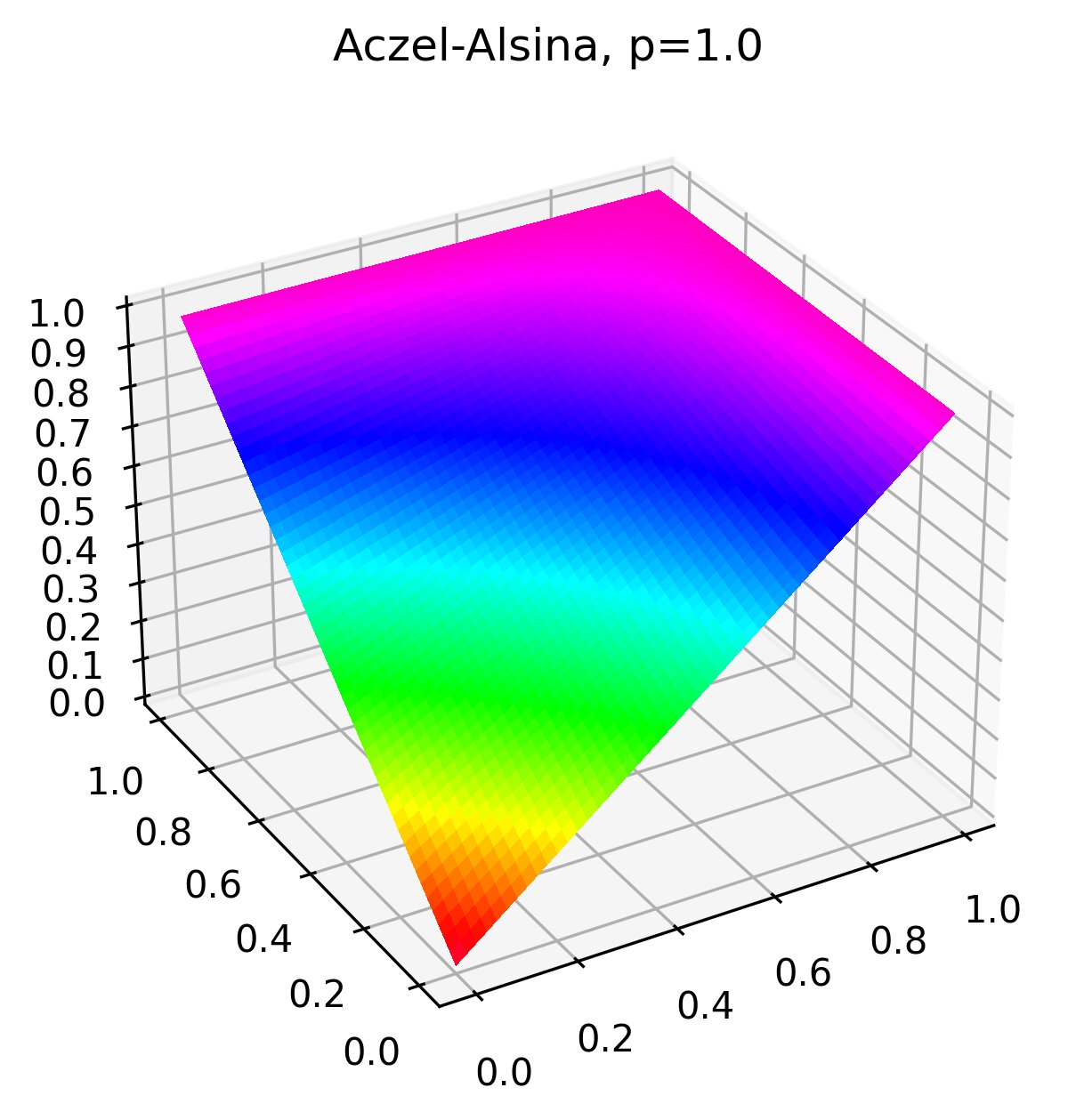} \hfill
  \includegraphics[width=.24\linewidth,trim={.25cm .0cm .25cm .0cm},clip]{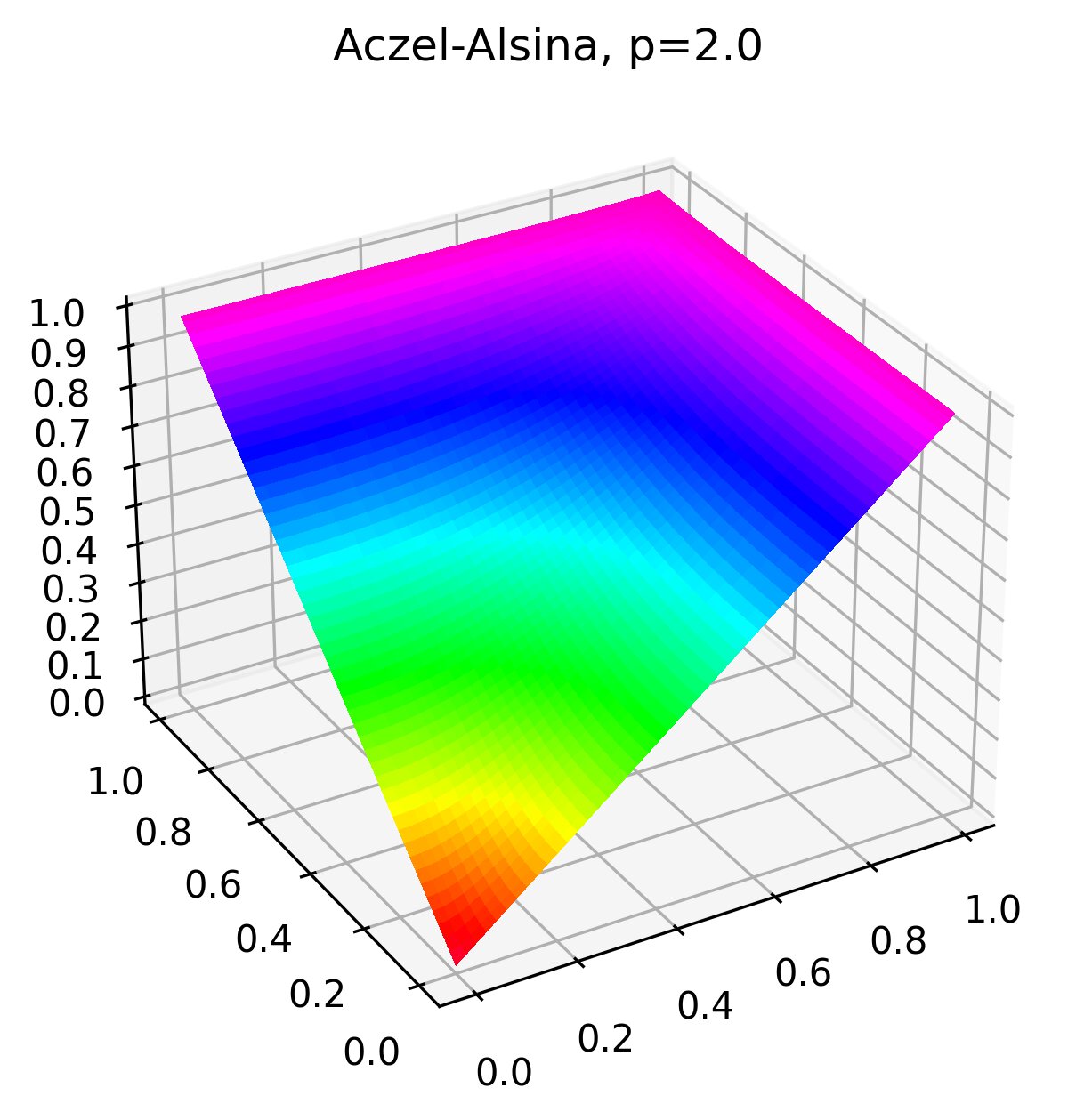} \hfill
  \includegraphics[width=.24\linewidth,trim={.25cm .0cm .25cm .0cm},clip]{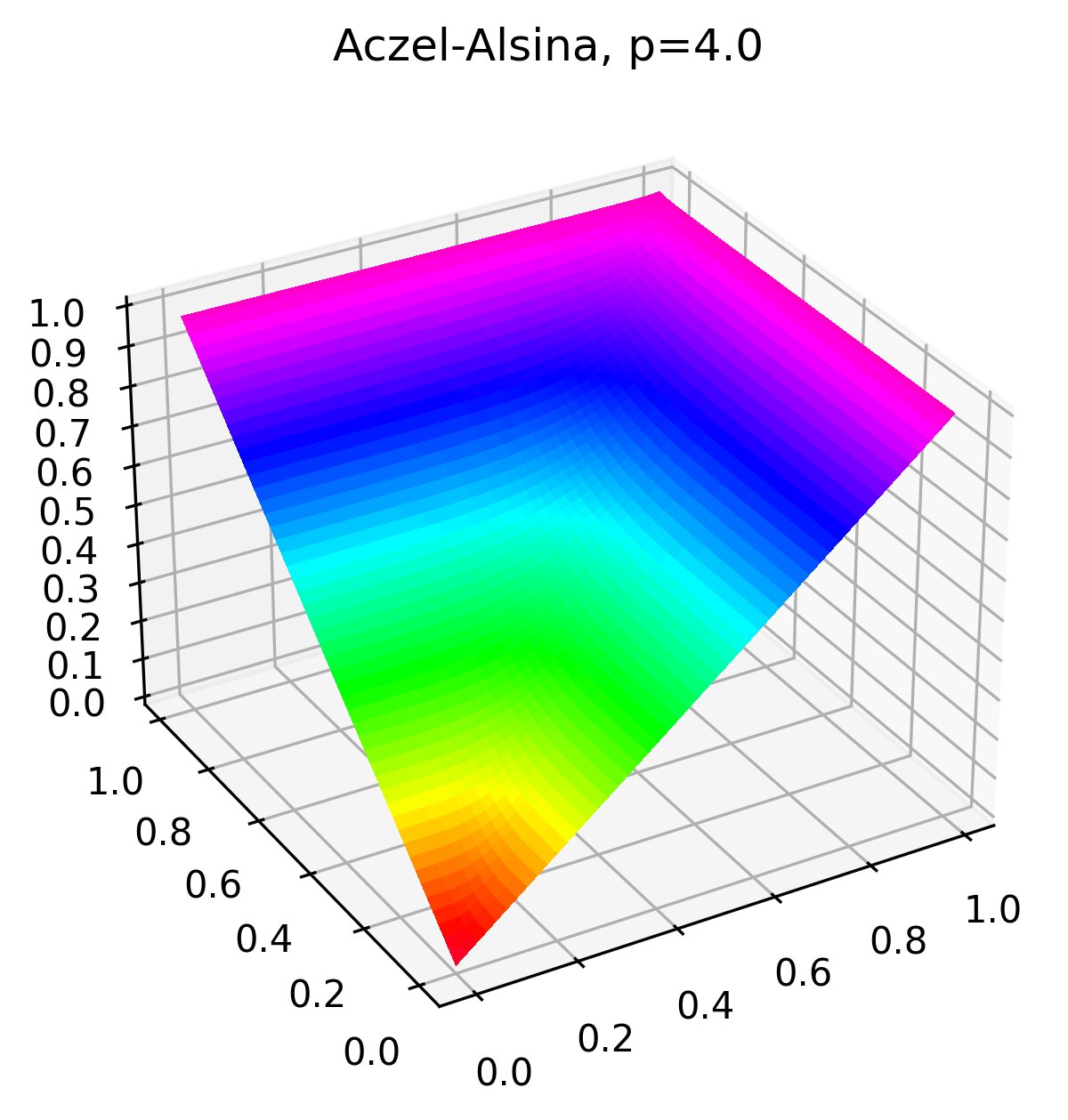} \hfill
  \includegraphics[width=.24\linewidth,trim={.25cm .0cm .25cm .0cm},clip]{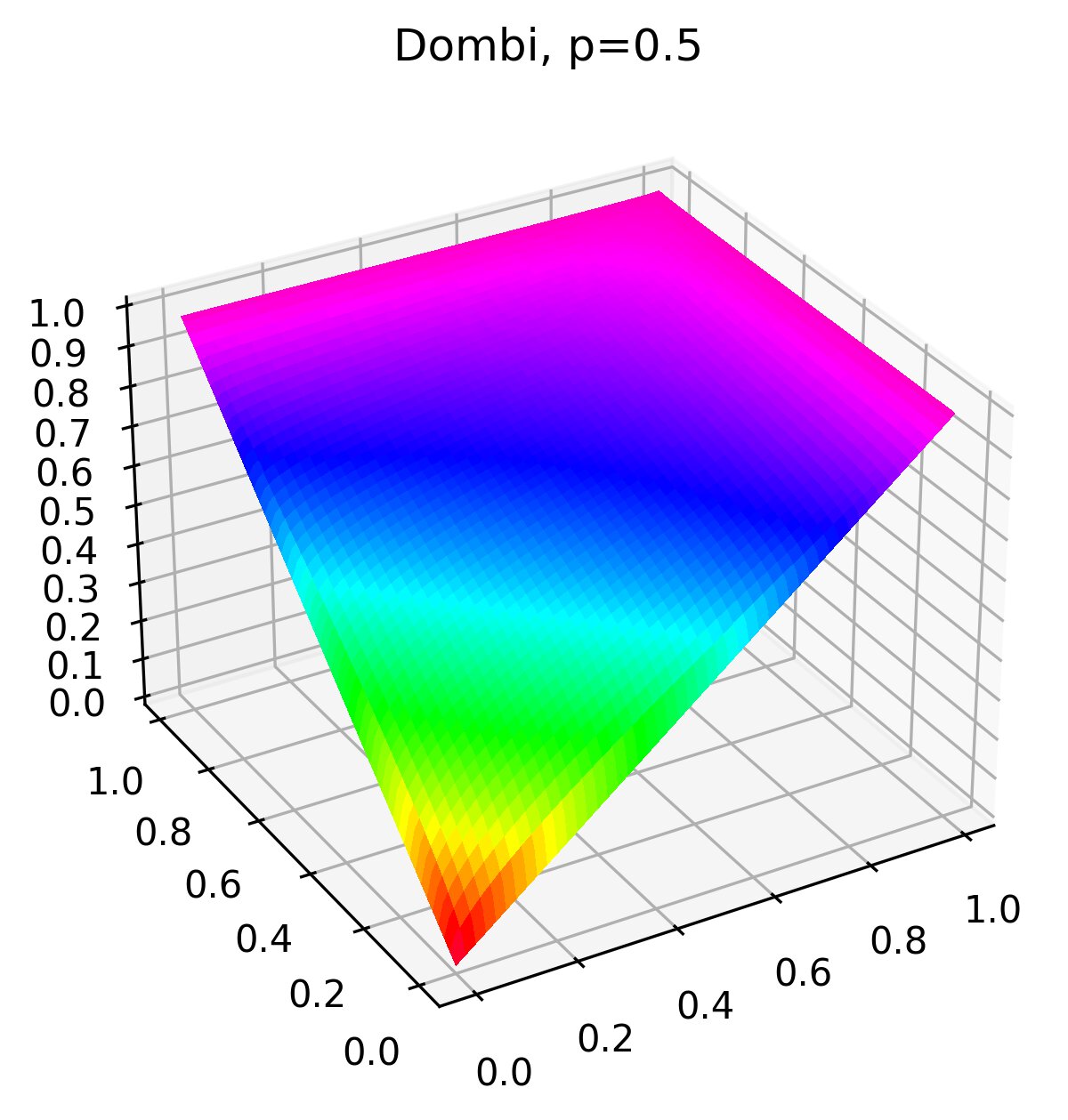} \hfill
  \includegraphics[width=.24\linewidth,trim={.25cm .0cm .25cm .0cm},clip]{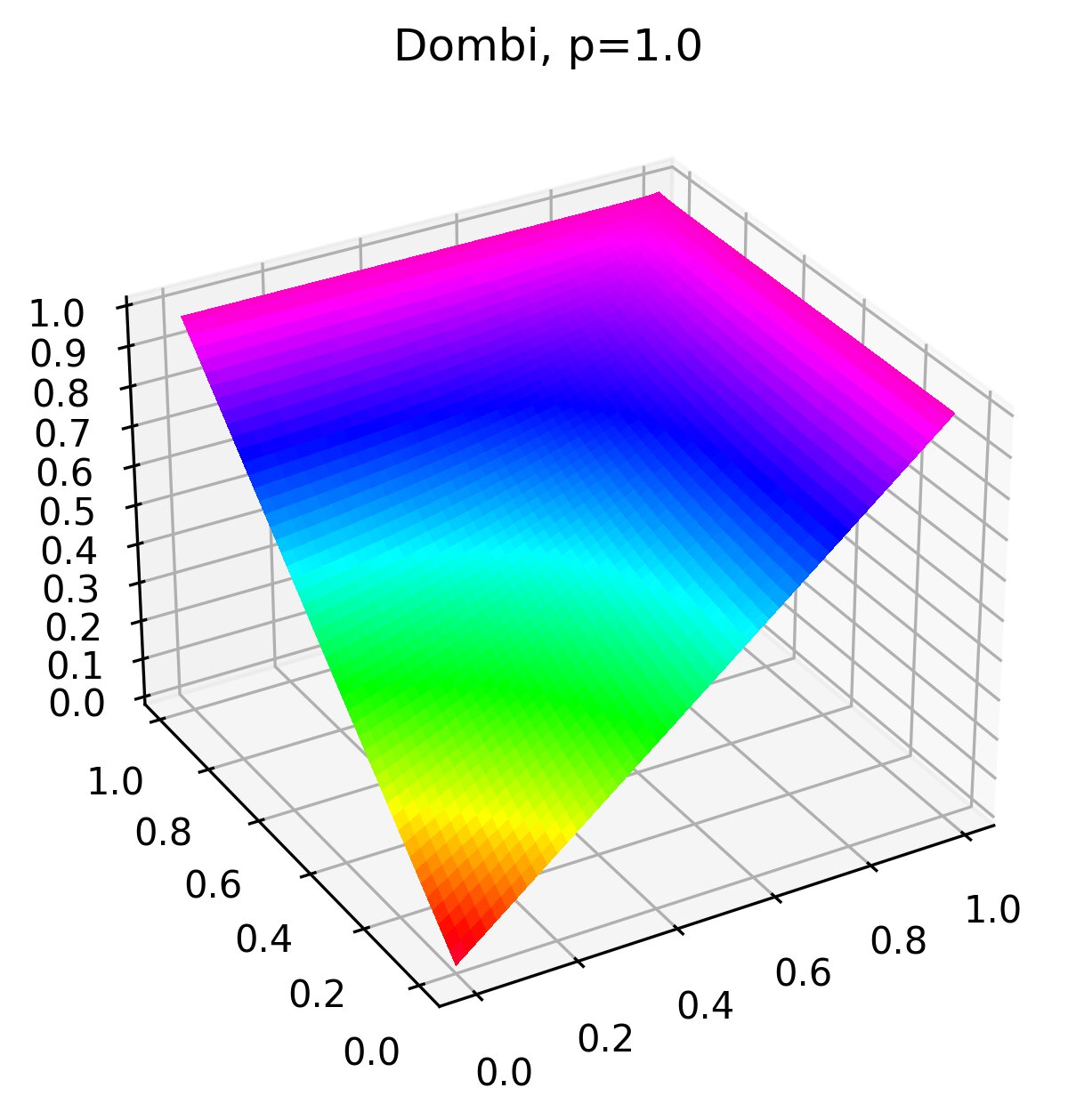} \hfill
  \includegraphics[width=.24\linewidth,trim={.25cm .0cm .25cm .0cm},clip]{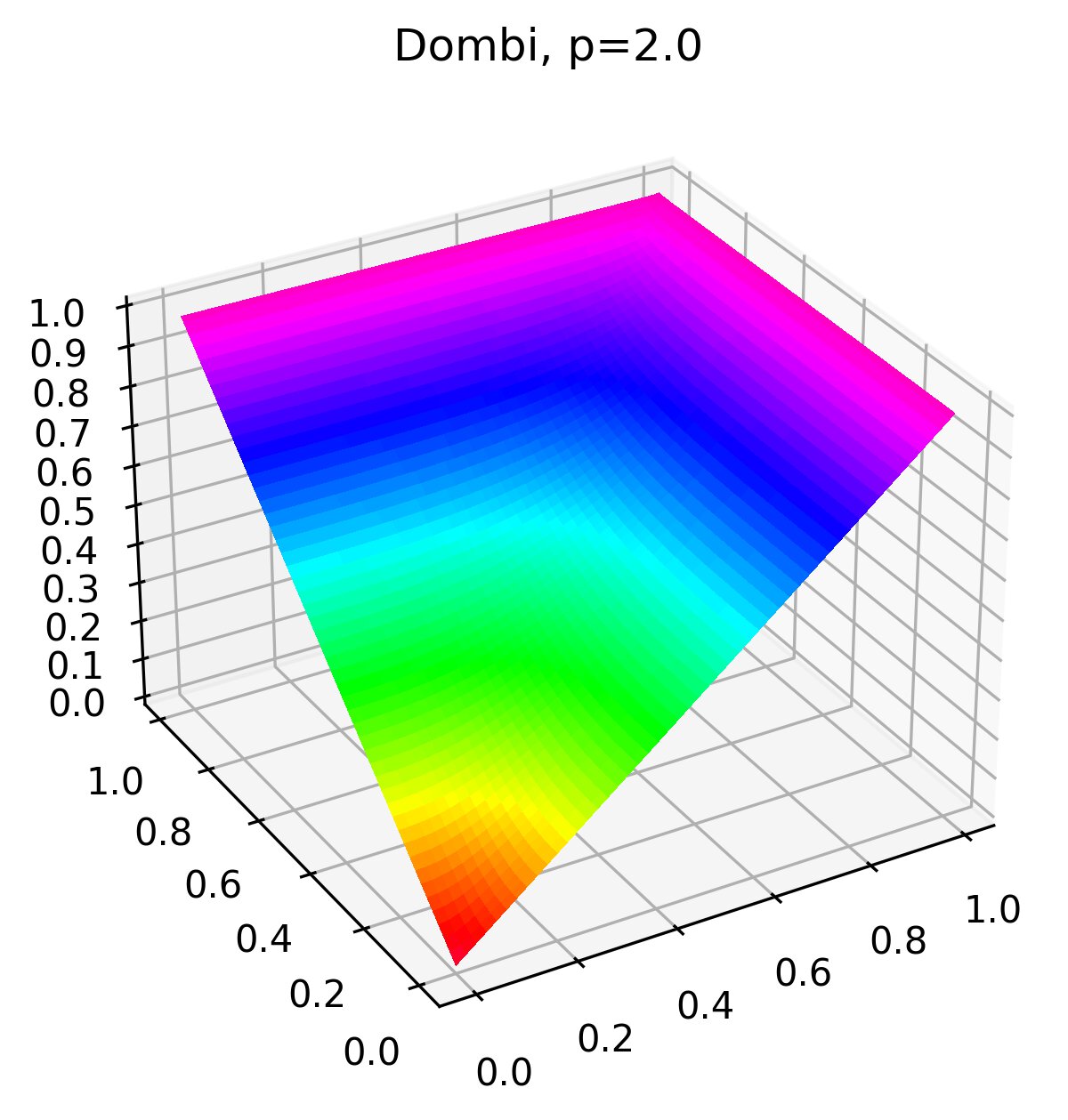} \hfill
  \includegraphics[width=.24\linewidth,trim={.25cm .0cm .25cm .0cm},clip]{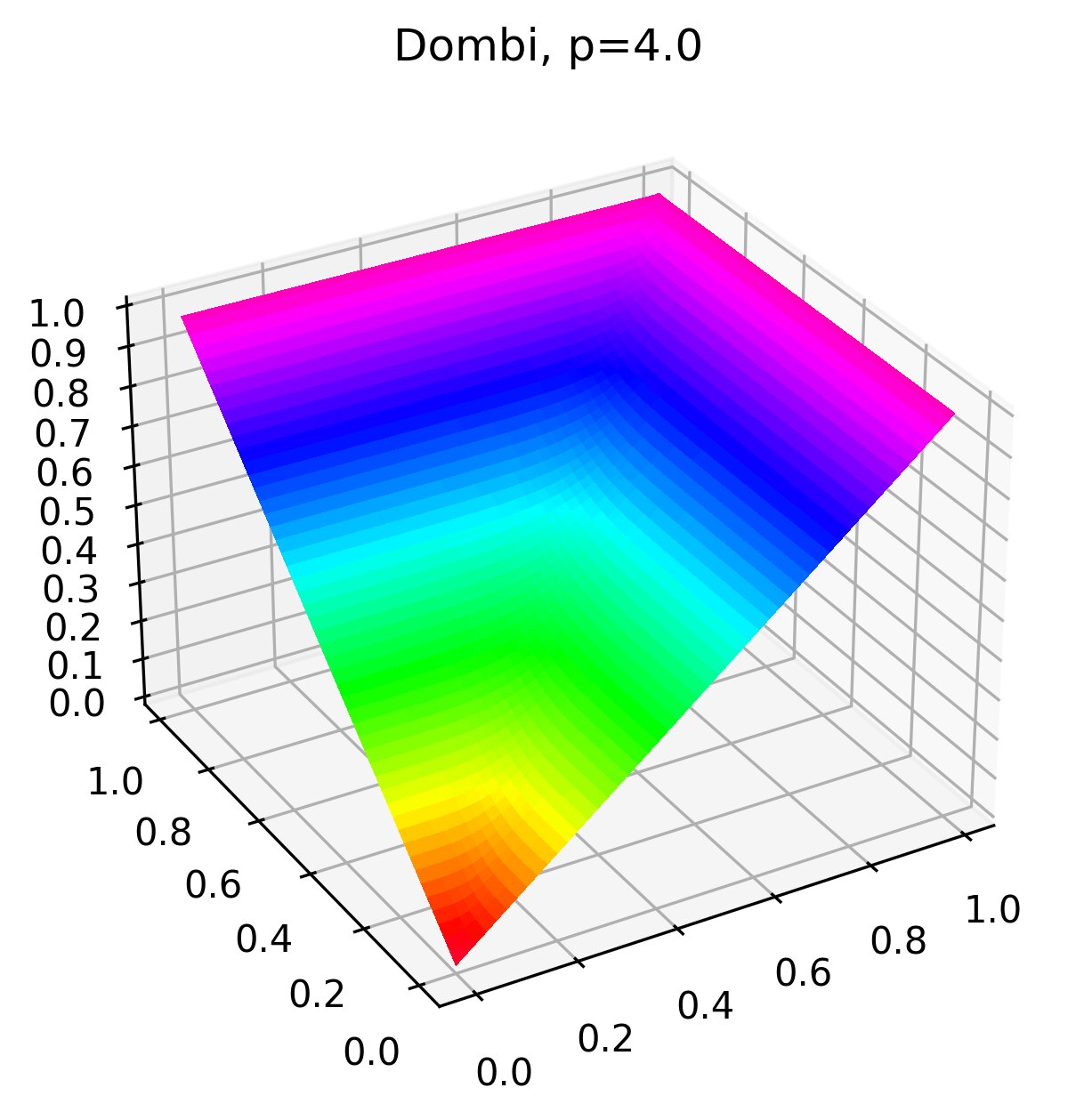} \hfill
  \includegraphics[width=.24\linewidth,trim={.25cm .0cm .25cm .0cm},clip]{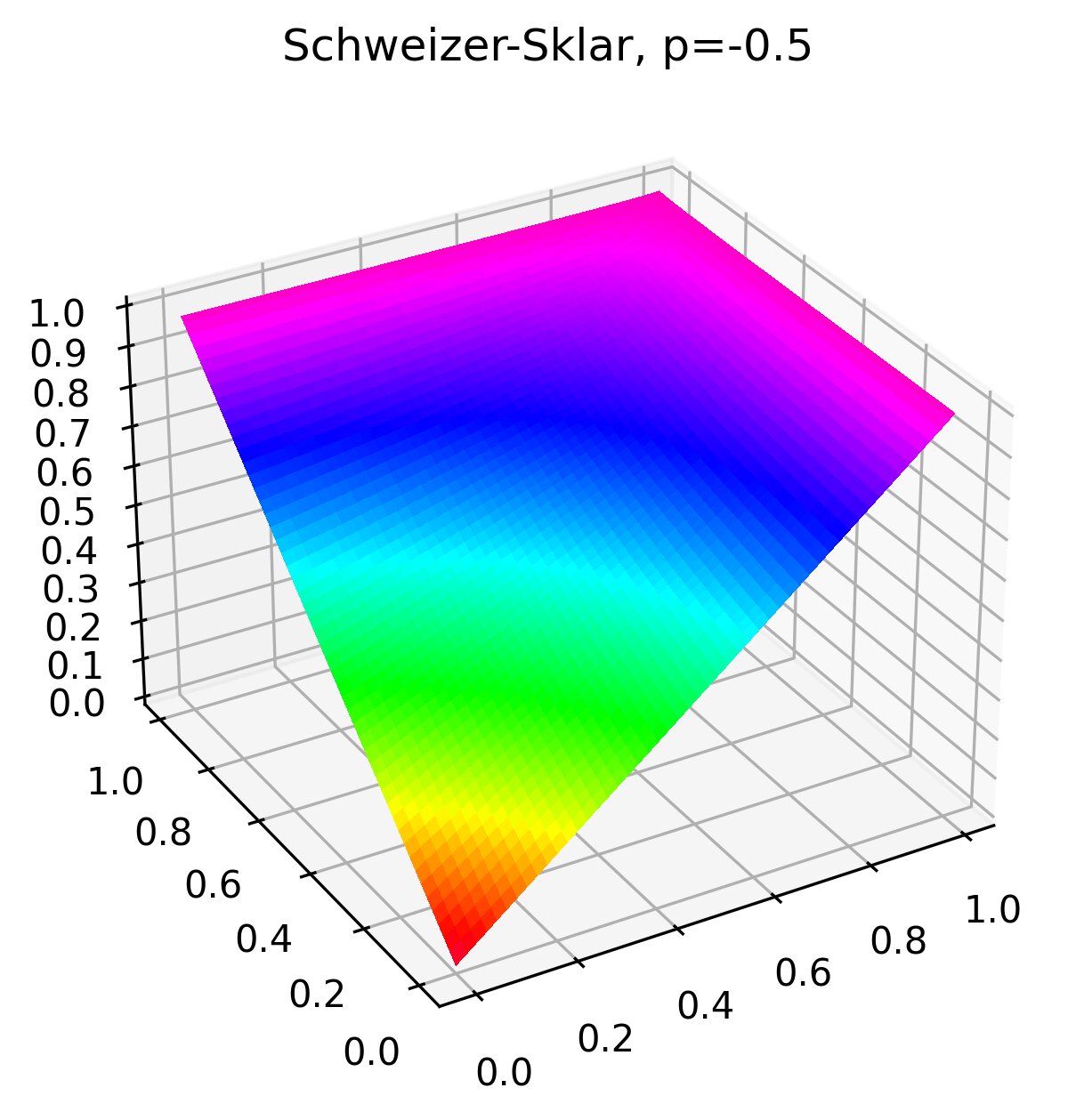} \hfill
  \includegraphics[width=.24\linewidth,trim={.25cm .0cm .25cm .0cm},clip]{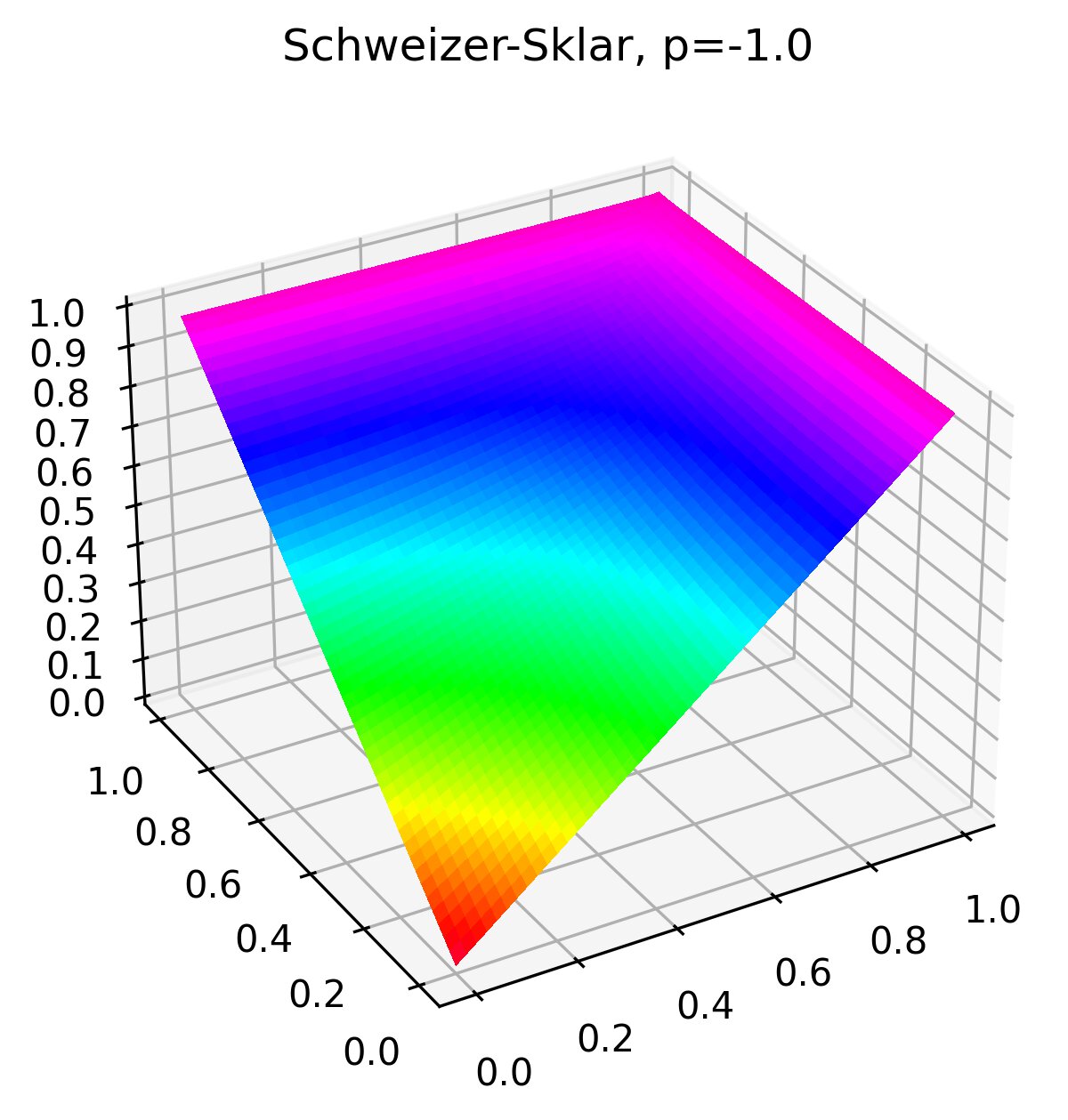} \hfill
  \includegraphics[width=.24\linewidth,trim={.25cm .0cm .25cm .0cm},clip]{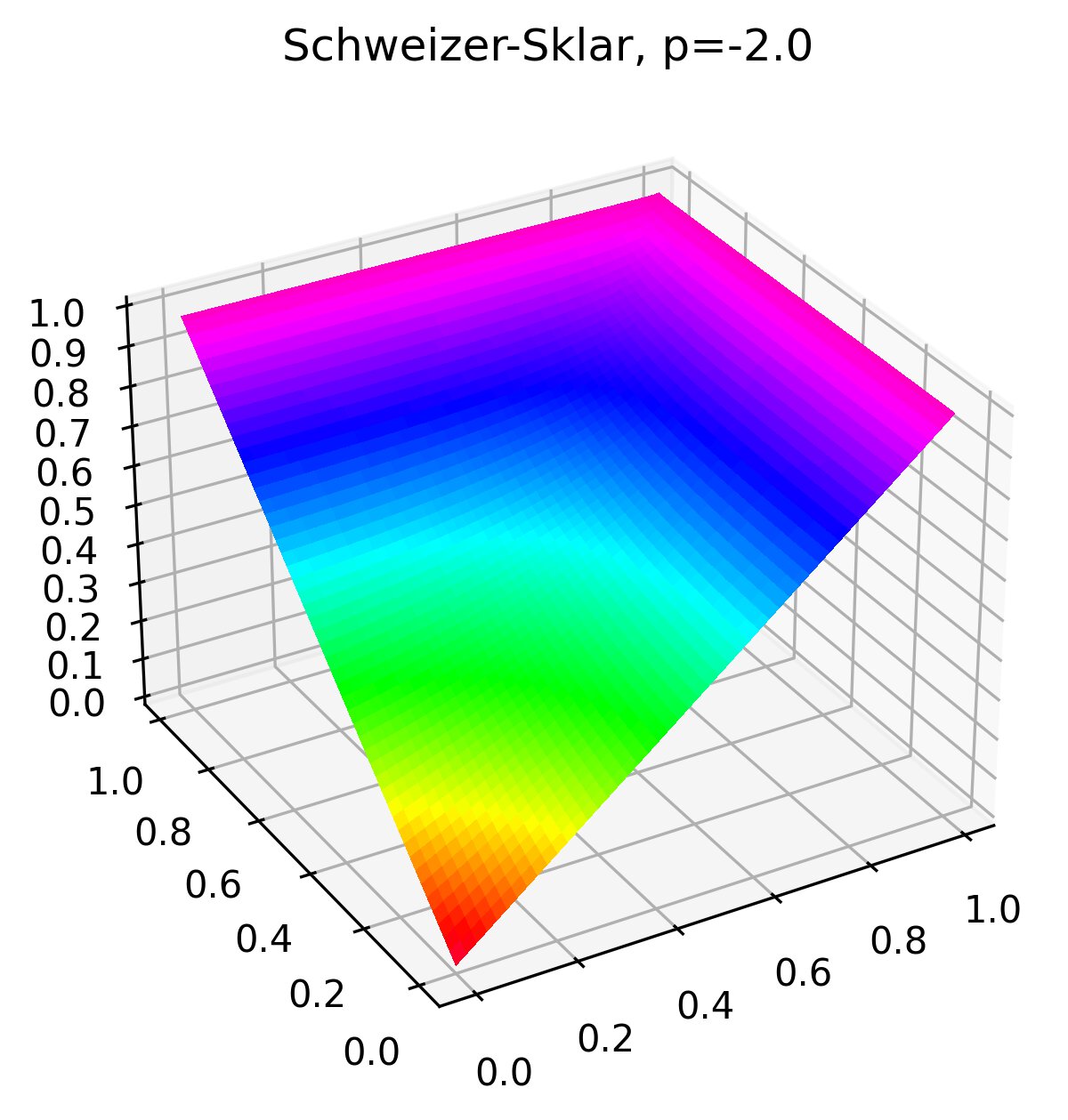} \hfill
  \includegraphics[width=.24\linewidth,trim={.25cm .0cm .25cm .0cm},clip]{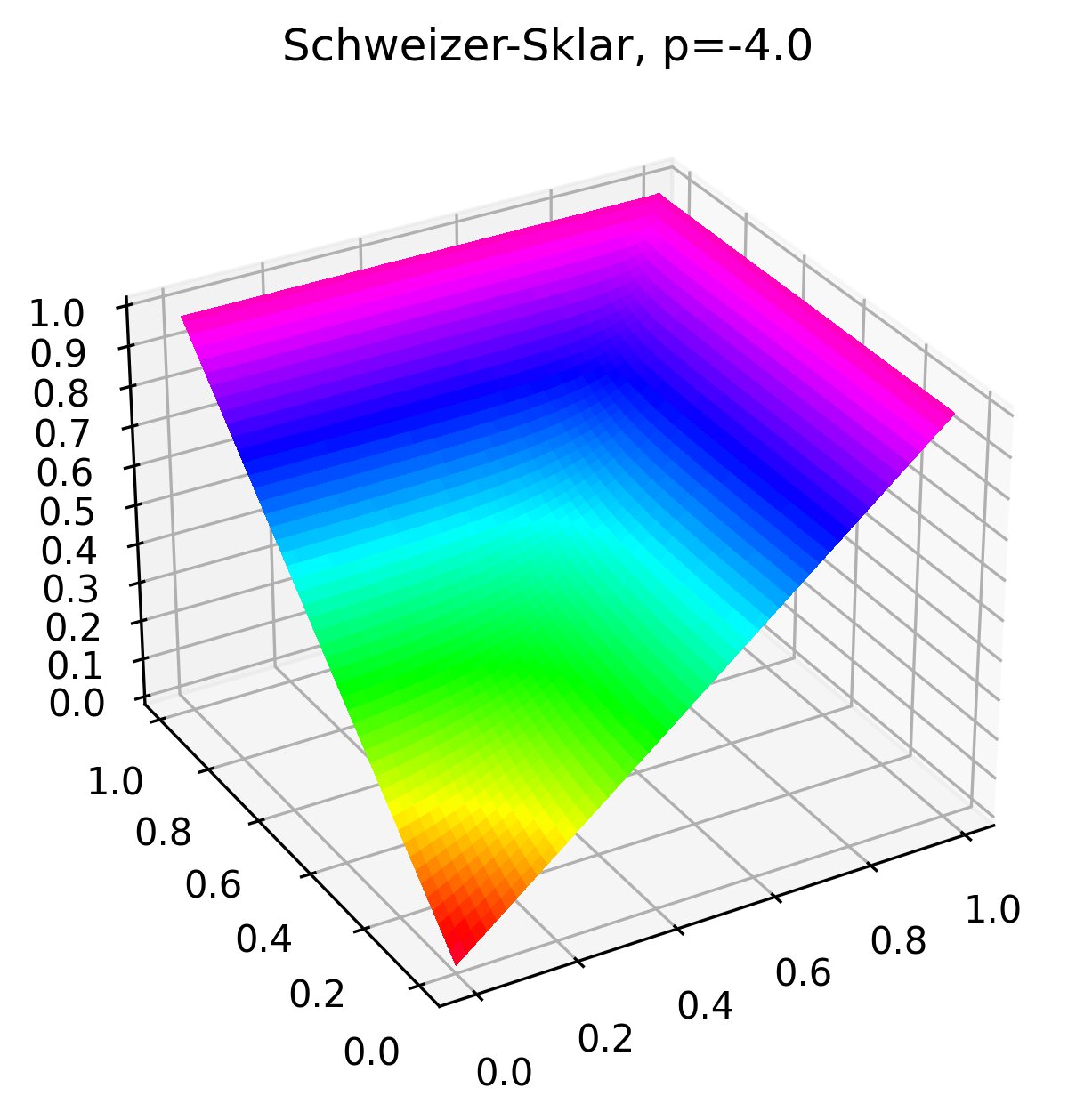} \hfill
  \caption{T-conorm plots (2/2).}
  \label{fig:t-plot-2}
\end{figure*}

Figures~\ref{fig:t-plot-1} and~\ref{fig:t-plot-2} display the considered set of T-conorms.

\clearpage

\begin{table*}[t]
\centering
\begin{tabular}{|l|c@{~$=$~}l|}\hline\rule{0pt}{2.6ex}%
Maximum         & $\bot^M(a,b)$
  & $\max(a,b)$                  \\[1.2ex]
Probabilistic   & $\bot^P(a,b)$
  & $a+b-ab$                     \\[1.2ex]
Einstein        & $\bot^E(a,b)$
  & $\bot^H_2(a,b)=\frac{a+b}{1+ab}$           \\[1.2ex]
Hamacher        & $\bot^H_p(a,b)$
  & $\frac{a+b+(p-2)ab}{1+(p-1)ab}$ \\[1.2ex]
Frank           & $\bot^F_p(a,b)$
  & $1-\log_p\left(1+\frac{(p^{1-a}-1)(p^{1-b}-1)}{p-1}\right)$ \\[1.2ex]
Yager           & $\bot^Y_p(a,b)$
  & $\min\left(1, (a^p+b\kern0.1pt^p)^{\frac{1}{p}}\right)$ \\[1.2ex]
Acz\'el-Alsina  & $\bot^A_p(a,b)$
  & $1 -\exp\big(-\left(|\log(1-a)|^p+|\log(1-b)|^p
                 \right)^{\frac{1}{p}}\big)$ \\[0.3ex]
Dombi           & $\bot^D_p(a,b)$
  & $\Big(1+\left( \left(\frac{1-a}{a}\right)^p
                  +\left(\frac{1-b}{b}\right)^p
           \right)^{-\frac{1}{p}}\Big)^{\!\!-1}$ \\[1.2ex]
Schweizer-Sklar & $\bot^S_p(a,b)$
  & $1-((1-a)^p+(1-b)^p-1)^{\frac{1}{p}}$ \\[1ex] \hline
\end{tabular}
\caption{\label{tab.t-conorms}(Families of) T-conorms.}
\end{table*}

\clearpage

\section{Additional Plots}
\label{sm:add-plots}
\begin{figure}[h!]
  \centering
  \includegraphics[width=\linewidth,trim={0.5cm 0.15cm 0.25cm 0.6cm},clip]{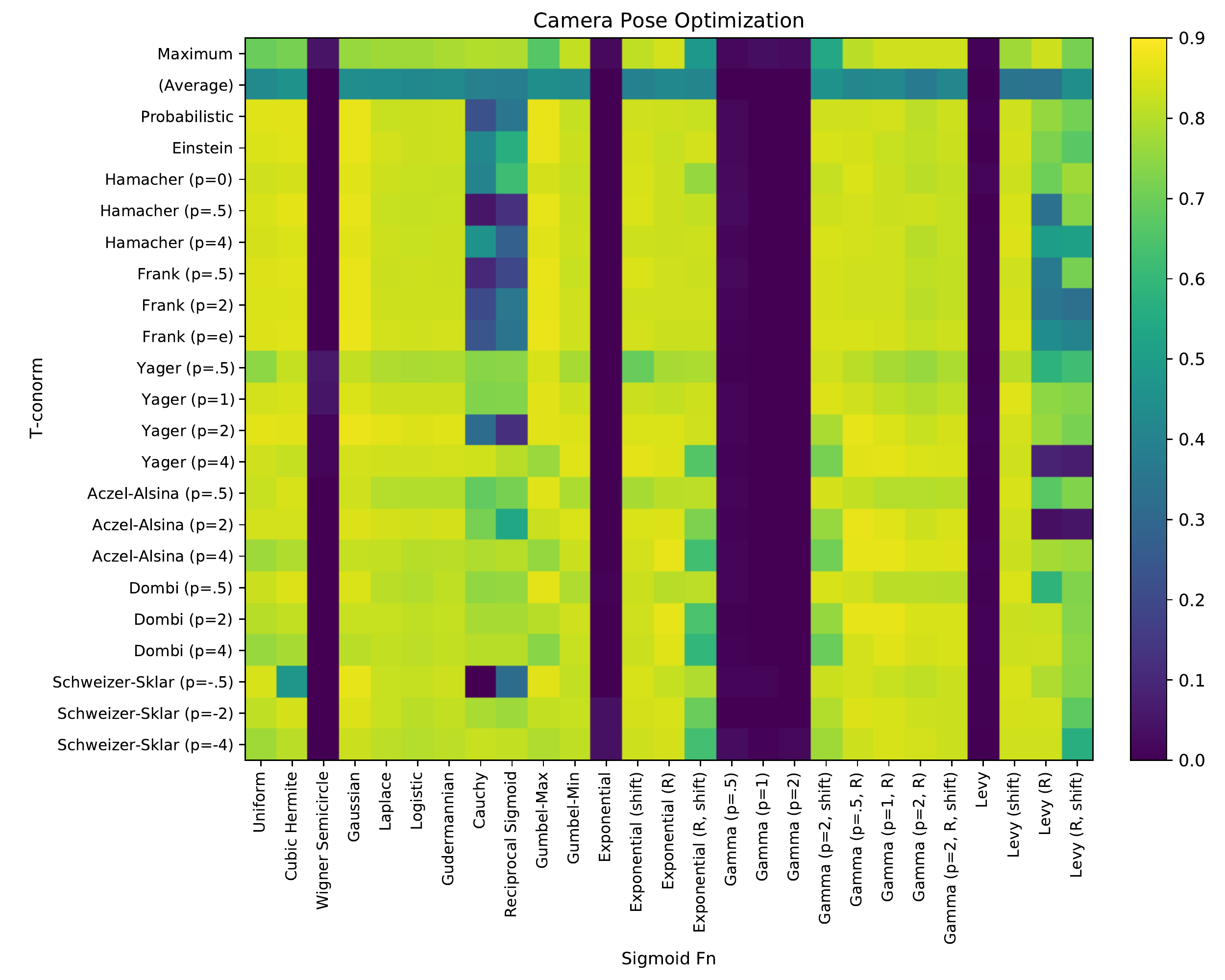}

  \xcaption{
  Results for the tea pot camera pose optimization task for the respective square-root distribution~$\cdfsq$.
  }
  \label{fig:camera-opt-plot-squares}
\end{figure}

\begin{figure*}[h]
    \centering
    \includegraphics[width=.5\linewidth,trim={0.5cm 0.0cm 0.25cm 0.5cm},clip]{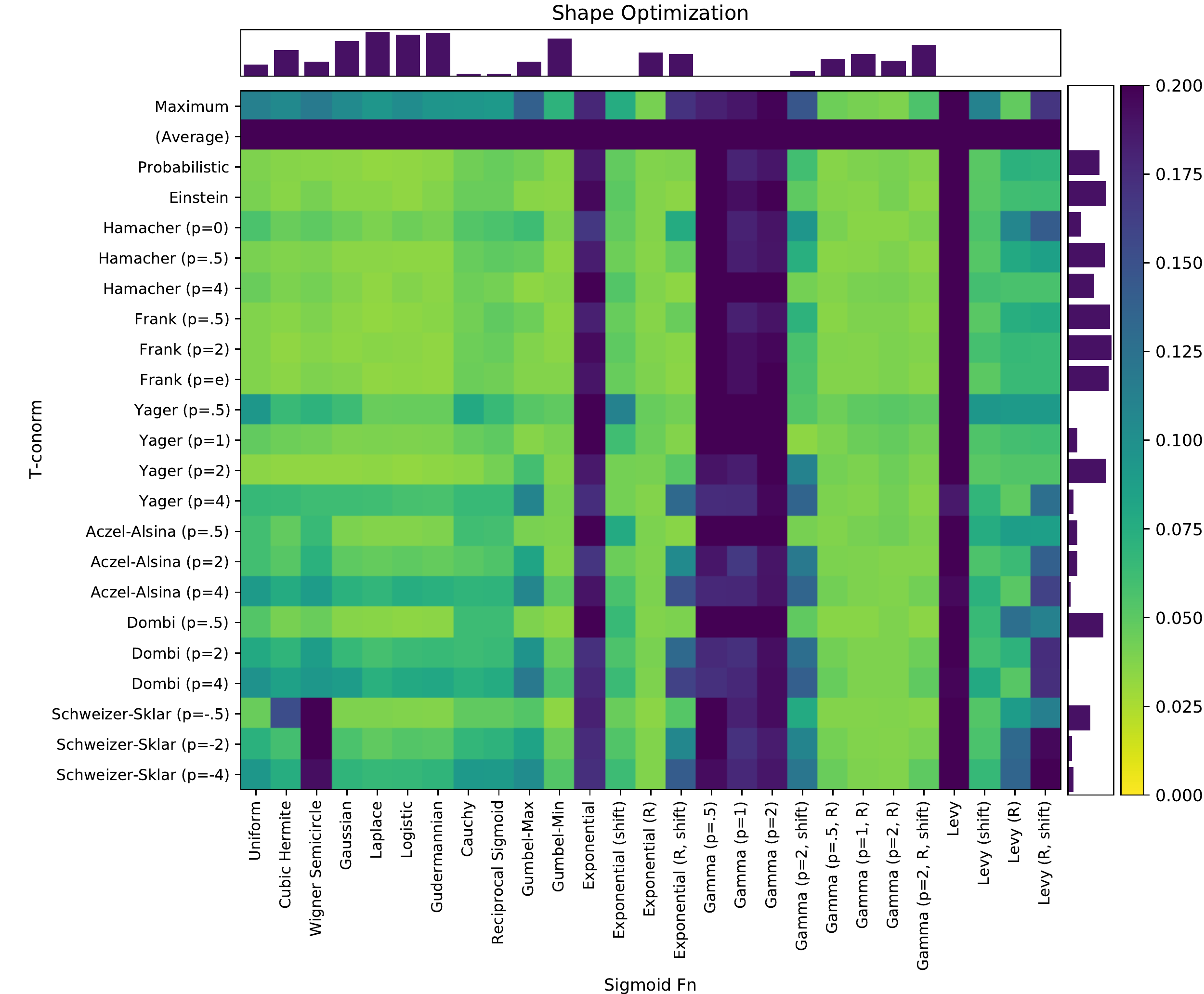}%
    \includegraphics[width=.5\linewidth,trim={0.5cm 0.0cm 0.25cm 0.75cm},clip]{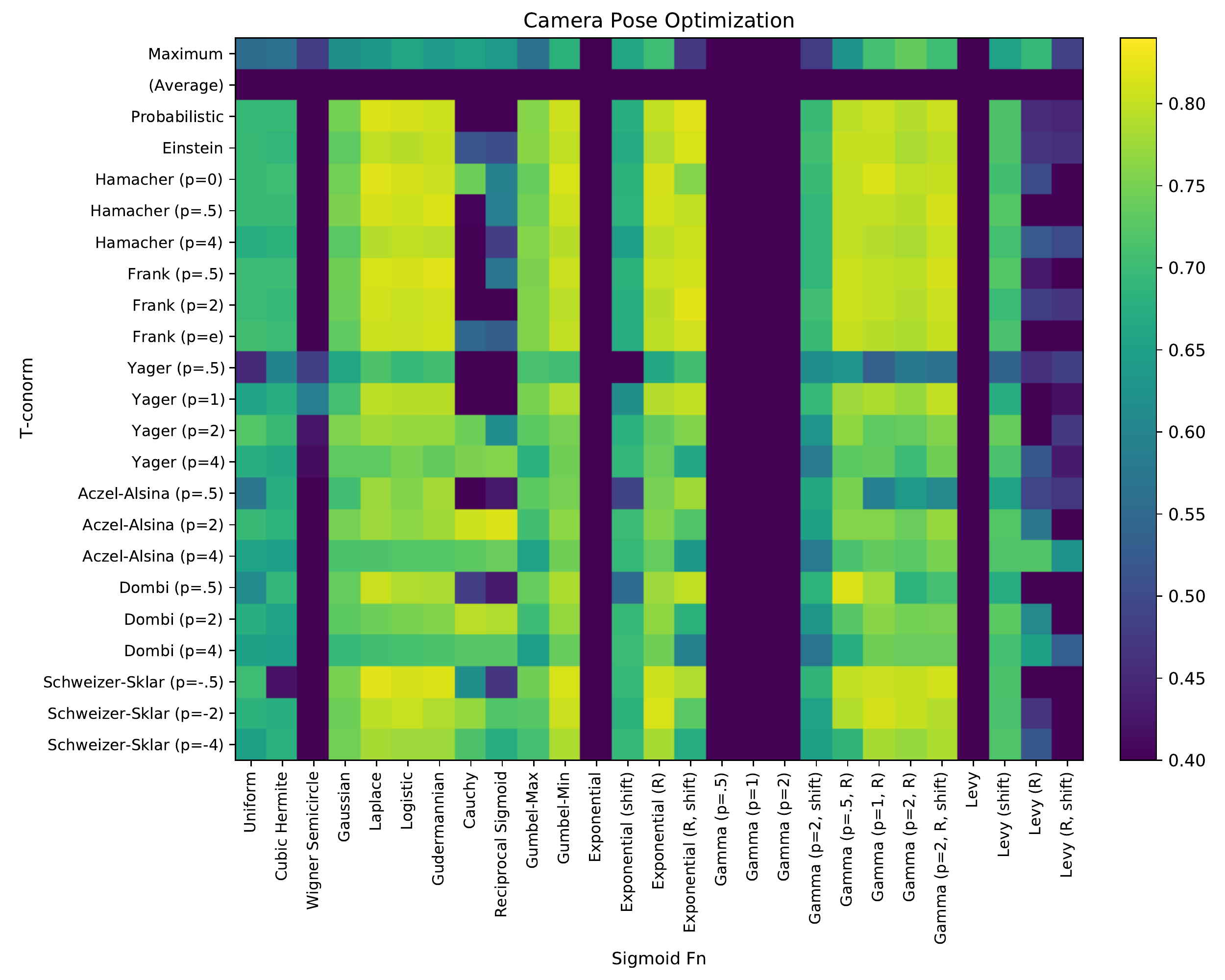}\\%
    \includegraphics[width=.5\linewidth,trim={0.5cm 0.0cm 0.25cm 0.5cm},clip]{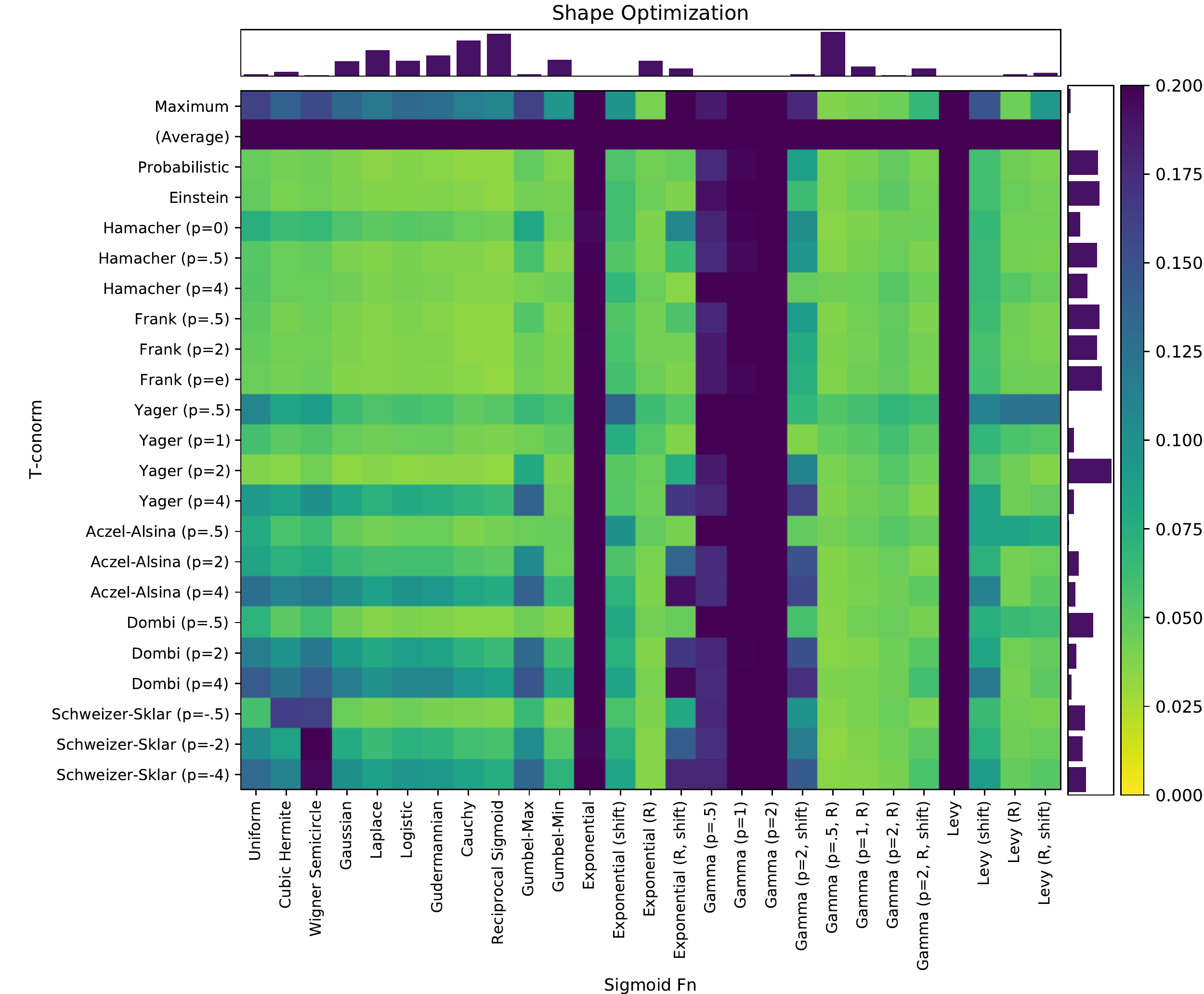}%
    \includegraphics[width=.5\linewidth,trim={0.5cm 0.0cm 0.25cm 0.75cm},clip]{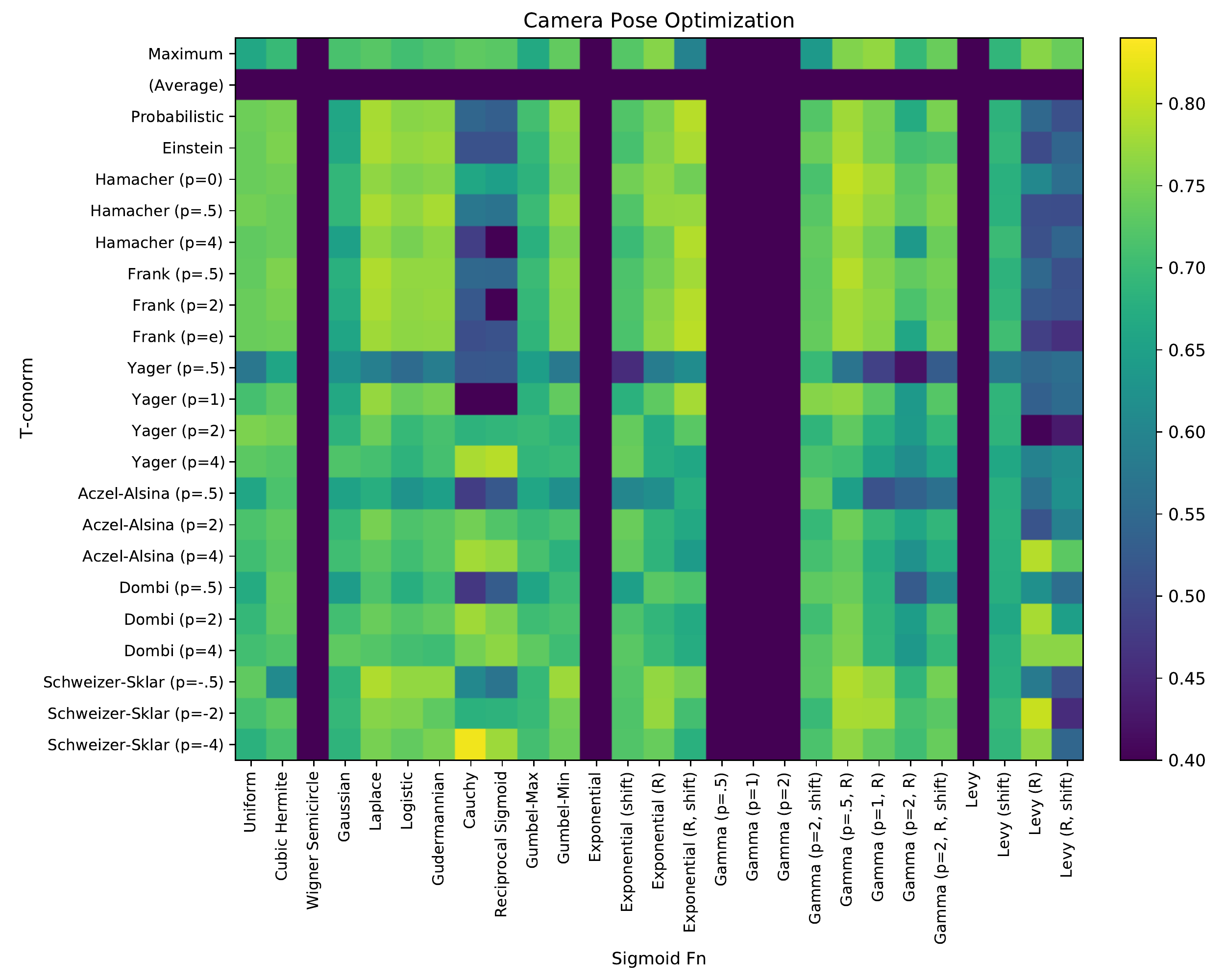}%
    \caption{
    Shape optimization (left) and camera pose optimization (right) applied to a model of a chair. Top: set of original distributions $F$. Bottom: set of the respective square-root distributions $\cdfsq$
    }
    \label{fig:chair-plots}
\end{figure*}

See Figures \ref{fig:camera-opt-plot-squares} and \ref{fig:chair-plots}.

\printbibliography

\end{document}